%% file: minimax.tex
\documentclass[twoside]{article}

\usepackage[preprint]{a2027}

\usepackage[numbers, sort&compress]{natbib}
\usepackage{float}

\usepackage{multirow}
\usepackage{enumitem}
\usepackage[utf8]{inputenc} % allow utf-8 input
\usepackage[T1]{fontenc}    % use 8-bit T1 fonts
\usepackage{hyperref}       % hyperlinks
\usepackage{nicefrac}       % compact symbols for 1/2, etc.
\usepackage{xcolor}         % colors

\usepackage{amsmath,amsfonts}
\usepackage{array}
\usepackage{textcomp}
\usepackage{stfloats}
\usepackage{url}
\usepackage{bm}
\usepackage{verbatim}
\usepackage{graphicx}
\usepackage{natbib}
\usepackage{amsmath}
\usepackage{amssymb}
\usepackage{mathtools}
\usepackage{amsthm}
\usepackage{microtype}
\usepackage{graphicx}
\usepackage{subcaption}
\usepackage{booktabs} % for professional tables
\usepackage[capitalize,noabbrev]{cleveref}
\usepackage{wrapfig}
\usepackage{algorithm}
\usepackage{algorithmic}
\theoremstyle{plain}
\newtheorem{theorem}{Theorem}[section]

\newtheorem{lemma}[theorem]{Lemma}

\theoremstyle{definition}
\newtheorem{definition}[theorem]{Definition}
\newtheorem{assumption}[theorem]{Assumption}

\theoremstyle{remark}
\newtheorem{remark}[theorem]{Remark}

\newcommand{\Y}{\mathcal{Y}}
\DeclareMathOperator*{\argmax}{arg\,max}
\begin{document}
\addtocontents{toc}{\protect\setcounter{tocdepth}{0}}


\runningauthor{M. Yashwanth, S. Ghosh, S. Bulusu, and A. Chakraborty}

\twocolumn[

\aistatstitle{Personalized Projected Ascent for Federated Stackelberg Optimization}

\aistatsauthor{ 
    M Yashwanth \\ \normalsize Indian Institute of Science, Bangalore \\ \normalsize \texttt{yashwanthm@iisc.ac.in} \And 
    Sharannya Ghosh \\ \normalsize CISPA, Germany \\ \normalsize \texttt{sharannyaghosh31@gmail.com} \AND 
    SaiKiran Bulusu \\ \normalsize IIIT-Hyderabad \\ \normalsize \texttt{saikiran.bulusu@iiit.ac.in} \And 
    Anirban Chakraborty \\ \normalsize Indian Institute of Science, Bangalore \\ \normalsize \texttt{anirban@iisc.ac.in} 
}

\aistatsaddress{ ~ } ]
\vspace{-0.35in}
\vspace{-0.35in}

\begin{abstract}
\vspace{-0.15in}
Federated minimax problems, such as distributionally robust optimization and
adversarial training, include an inner maximization variable built from each
client's local data. These variables differ in meaning and dimension from one client to the next, so averaging them across clients, as standard federated minimax methods do, is ill-posed. 
We therefore study a federated minimax problem where the minimization variable $\mathbf{x}$ is shared and aggregated, while each client retains the maximization variable $\mathbf{y}_i$ (personalized). Because these applications constrain $\mathbf{y}_i$ to a compact convex set, we analyze projected ascent directly. We show that, in the nonconvex-concave regime, this projected personalized scheme attains the standard sublinear rate. In the nonconvex-PL regime, however, it hits a stepsize-invariant noise floor due to stochastic $\mathbf{y}_i$ updates. To mitigate this, we propose VR-SGDA, incorporating a variance-reduced ascent step where each client maintains a private variable snapshot. By evaluating current and snapshot gradients on the same sample, the constant variance is replaced by a vanishing drift under a two-timescale regime. VR-SGDA eliminates the noise floor, attaining sublinear leading rates while preserving personalization and projection. Experiments on subpopulation-robustness  and NLP adversarial training confirm both the worst-group robustness benefits and the two-timescale tracking mechanism.
\end{abstract}
\vspace{-0.15in}
\section{Introduction}
\vspace{-0.05in}
\label{sec:introduction}
Federated Learning (FL)~\cite{mcmahan2017communication} is a distributed learning framework to train a shared model across clients whose data cannot be centralized. In many of the settings where this
matters most, the goal is not merely good \emph{average} performance but
\emph{robust} performance: the learner must account for worst-case distribution shifts in the data. Two canonical instances are distributionally robust optimization
(DRO), where one optimizes against the worst reweighting of the data within an
ambiguity set \citep{bental2013,namkoong2016,levy2020}, and adversarial training,
where one optimizes against worst-case input perturbations
\cite{madry2018,sinha2018}. Both are naturally minimax problems: an outer
minimization over the model and an inner maximization that constructs the worst
case. Bringing either into the federated setting therefore requires solving a
federated minimax problem over heterogeneous clients.

The key feature of these problems, and the one that has been overlooked,
is that the inner maximization variable is \emph{associated with the
client's own data}: in federated DRO it is a reweighting
$p_i \in \Delta_{m_i}$ of client $i$'s local samples; in federated
adversarial training it is a perturbation of client $i$'s local inputs.
These variables live in client-specific spaces, $\Delta_{m_i}$ vs.\
$\Delta_{m_j}$ when clients hold different sample counts, with
client-specific feasible sets, so aggregating them at the server is not a meaningful operation. Existing federated minimax methods nonetheless maintain a single global inner variable and average it at the server
\citep{deng2021localsgda,deng2020drfa}; while one could define
aggregation rules, the resulting shared variable would not correspond to
the per-client robust objective that DRO and adversarial training are
designed to optimize.

This motivates maintaining the maximization variable locally: the model
is what is ultimately deployed and remains shared, but the inner
variable must be \emph{personalized}, maintained locally, per client,
and never communicated. We are therefore led, not by design preference
but by the structure of the applications, to the objective
$\min_{\mathbf{x}}\; \frac{1}{n}\sum_{i=1}^n
  \max_{\mathbf{y}_i \in \mathcal{Y}_i} f_i(\mathbf{x}, \mathbf{y}_i),$
in which a shared model $\mathbf{x}$ is trained against a personalized
worst-case variable $\mathbf{y}_i$ on each client, with
$\mathcal{Y}_i$ a compact convex set. This is strictly different from
the shared-variable problem
$\min_{\mathbf{x}}\max_{\mathbf{y}}\frac{1}{n}\sum_i f_i(\mathbf{x},\mathbf{y})$
studied previously.
\begin{table*}[t]
\centering
\caption{Comparison of related federated robust optimization frameworks. Our
method is the only one that maintains personalized maximization variables
$\mathbf{y}_i$ updated by \emph{projected} ascent on constrained
set, under a general adversary coupling.}
\label{tab:literature_comparison}
\resizebox{0.9\textwidth}{!}{%
\begin{tabular}{lllcccc}
\toprule
\textbf{Method} &
\textbf{$\mathbf{x}$} &
\textbf{$\mathbf{y}$} &
\textbf{\begin{tabular}[c]{@{}c@{}}Application\\motivated\end{tabular}} &
\textbf{\begin{tabular}[c]{@{}c@{}}Per-client\\$\mathbf{y}$\end{tabular}} &
\textbf{\begin{tabular}[c]{@{}c@{}}Inner\\coupling\end{tabular}} &
\textbf{\begin{tabular}[c]{@{}c@{}}Projected\\ascent on $\mathbf{y}$\end{tabular}} \\
\midrule
Local SGDA~\cite{deng2021localsgda}         & shared  & shared        & $\times$   & $\times$   & general             & $\times$ \\
DRFA~\cite{deng2020drfa}                    & shared  & shared        & partial    & $\times$   & client-weight       & $\times$ \\
Minimax Convergence Analysis~\cite{pmlr-v162-sharma22c} & shared  & shared        & $\times$   & $\times$   & general             & $\times$ \\
FLRA~\cite{reisizadeh2020robust}       & shared  & personalized  & \checkmark & \checkmark & affine input shift  & $\times$ \\
Partial model~\cite{pillutla2022federated}  & partial & ---           & n/a        & n/a        & n/a                 & n/a \\
\midrule
\textbf{Ours} & \textbf{shared} & \textbf{personalized} & \checkmark & \checkmark & \textbf{general} & \textbf{\checkmark} \\
\bottomrule
\end{tabular}
}
\end{table*}
Personalizing the inner variable places the problem outside existing
federated minimax analyses, which rest on controlling the consensus
error of an averaged inner variable, a quantity that no longer
exists when $\mathbf{y}_i$ is per-client and never shared. What remains is
a different coupling: each client's personalized ascent runs against a
gradient evaluated at its own drifting local model, tracking a
client-specific moving target $\mathbf{y}_i^\star(\mathbf{x}_i)$, where $\mathbf{x}_i$ is the locally updated parameters of global model $\mathbf{x}$ at client $i$. In the nonconvex-concave regime, the bounded domain controls the inner tracking via a telescoping argument, and the two-timescale separation~\cite{lin2025twotimescale} transfers the centralized guarantee additively, with bounded client-drift terms and no snapshot device
required. Here, the diameter $D$ (maximum of the diameters of $\mathcal{Y}_i$) enters the bound through the problem constraints, not an assumption imposed on the iterates.
The nonconvex-PL regime reveals a deeper issue. Under the gradient-mapping PL condition, the vanilla projected SGDA
(Algorithm~\ref{alg:merged_sgda}) produces a \emph{stepsize-invariant
noise floor} $O(\sigma_y^2/\mu^2)$ in the inner tracking error: the
stochastic noise in the $\mathbf{y}$-ascent accumulates in a way that PL
contraction cannot fully absorb, so the iterates converge to a
\emph{neighborhood} of stationarity whose radius is controlled by the
noise variance, not the step sizes. This floor is invisible in
unconstrained analyses. To eliminate it, we propose
\textit{Personalized VR-SGDA} (Algorithm~\ref{alg:merged_sgda}), which
applies per-client SVRG-style variance reduction to the inner ascent:
each client maintains a snapshot of the shared and personalized parameters  and uses it as a control variate, reducing the effective
inner-gradient variance, which vanishes as the learning rate $\eta_x \to 0$. This
removes the noise floor and yields convergence to a true stationary
point. The snapshot here is taken at
synchronization boundaries, and is
used for variance reduction in the \emph{personalized} inner ascent,
not for the cross-client reconciliation that motivated snapshot devices
in prior work \citep{deng2021localsgda}. Our contributions are summarized as 

%\begin{itemize}[noitemsep]
\begin{itemize}[itemsep=5pt]
\item We formalize federated minimax optimization with a
shared minimization variable $\mathbf{x}$ and \emph{personalized}, maximization variables $\mathbf{y}_i$, each constrained to a
compact convex set $\mathcal{Y}_i$ fixed by client $i$'s own data. 

%We argue that this personalization is forced rather than chosen: under heterogeneity the clients' inner variables need not share a common space, and where they do, averaging best responses to distinct local objectives corresponds to no well-posed problem.

\item First, in the nonconvex-concave (NC-C) regime, we show that personalized projected Local SGDA attains standard sublinear federated rates without a noise floor. 
Second, in the nonconvex-PL (NC-PL) regime, we show the vanilla scheme incurs a stepsize-invariant noise floor $O(\sigma_y^2/\mu^2)$ under the gradient-mapping PL condition. To eliminate this floor, we propose Personalized VR-SGDA, which uses model snapshots at synchronization boundaries as per-client control variates to remove inner-gradient variance and restore convergence to a true stationary point.

\item  We instantiate both use cases concretely, the  DRO and adversarial training, and
confirm the predicted behavior empirically.
\end{itemize}
\vspace{-0.15in}
\section{Related Work}
\vspace{-0.1in}
\label{sec:related}
Convergence theory for minimax problems is well understood in the convex-concave regime, first-order primal-dual and mirror-prox methods achieve sub-linear rates
\citep{Nemirovski2004ProxMethodWR,nedic2009subgradient,juditsky2011solving}, and
in the strongly-convex--strongly-concave regime, gradient descent-ascent and its
variants converge linearly \citep{mokhtari2020unified}. Our setting sits outside
this spectrum on every axis: the outer problem is nonconvex, the inner problem is
only concave or PL, the feasible set is constrained, and the whole is solved in a
federated, personalized configuration. The federated minimax literature that does
address nonconvex outer problems inherits further assumptions we discard. We
organize it along four axes: \emph{what the maximization variable is} (a single
global object or per-client); \emph{whether the analysis is tied to a specific
adversary or application structure}, or holds for a general coupling between the
inner and outer variables; \emph{whether the inner ascent is projected} onto a
constraint set, or left unconstrained; and \emph{whether the analysis requires a
snapshot iterate} to control the inner update. As
Table~\ref{tab:literature_comparison} summarizes our configuration, shared
model, personalized maximization variable, general coupling, projected ascent, no
snapshot for nonconvex-concave is covered by no prior federated minimax analysis, and the resulting simplification stems from the modeling choice rather than a new proof technique.

\citet{deng2021localsgda} establish convergence of Local SGDA in the federated
nonconvex-concave setting, with a single global maximization variable $y$
averaged at the server. Their analysis controls an inner-variable consensus
error from averaging $y$ across clients, a quantity with no analogue in our
setting, where $\mathbf{y}_i$ is per-client and never communicated. To handle the federated nonconvex regime, they introduce Local SGDA$^+$,
which uses a snapshot of the outer variable inside the inner update to
decouple the inner-maximization analysis from the moving local iterates.
\citet{pmlr-v162-sharma22c, sharmafederated} extend this line to broader
nonconvex-nonconcave regimes, again with an unconstrained shared maximization variable and a snapshot-style algorithm.

The closest application precedent is (DRFA) \citet{deng2020drfa}, also about federated DRO but with a fundamentally different configuration. DRFA's maximization
variable is a single global $\lambda\in\Delta_N$ over clients, optimizing
$\min_w\max_{\lambda\in\Delta_N}\sum_i\lambda_i f_i(w)$ and updated only at the
server via a snapshot iterate. Our $p_i\in\Delta_{m_i}$ has dimension equal to
client $i$'s local sample count, is updated locally, and expresses the worst
case \emph{within} client $i$'s data. DRFA asks ``which client to protect
against''; ours asks ``which examples within each client are its worst case.''
Prior federated minimax analyses rely on a snapshot iterate to decouple the
maximization update from the moving outer trajectory, reconciling a single
shared maximization variable with $n$ clients' drifting iterates
\citep{deng2021localsgda, pmlr-v162-sharma22c}. Personalization removes this
difficulty at its source: each $\mathbf{y}_i$ tracks its own $\mathbf{x}_i$ via
the two-timescale separation, with no cross-client object to reconcile. The
resulting simplification (no snapshot, additive drift only) is a structural
consequence of our configuration rather than a new proof technique.

Closest to our configuration is \citet{reisizadeh2020robust}, which also
maintains per-client maximization variables, but assumes PL on \emph{both}
variables (excluding our NC-C/DRO regime, in their work
$\mathbf{y}_i$, being the affine transform on $\mathbf{x}_i$ which is very restrictive) states PL on the raw gradient (inappropriate for projected
ascent), and restricts to affine-shift adversaries.
FedNest~\citep{tarzanagh2022fednest} requires a strongly-convex inner problem
with second-order hypergradient computation, excluding both the merely-concave
and constrained-PL regimes we treat.
\citet{wu2023solving} address federated minimax with a shared maximization
variable, unconstrained inner update, and raw-gradient PL in the cross-device
regime. \citet{pillutla2022federated} personalize part of the
\emph{minimization} variable under a standard (non-minimax) objective.

%\paragraph{Summary.}
%Prior federated minimax work has analyzed a configuration (shared maximizationvariable) inherited from centralized minimax---a configuration that does not match the per-client structure of federated DRO and adversarial training. The maximization variable in these applications is derived from local data and inherits the same federated principles: it should not be communicated, averaged, or maintained globally. We make this correction explicit and provide the convergence analysis that the corrected configuration requires.
\vspace{-0.1in}
\section{Method}
\vspace{-0.1in}

\textit{Motivation for Stackelberg:} The minimax problem admits two solution concepts. The symmetric Nash equilibrium guarantees existence and order-invariance in convex-concave settings, but fails under non-convexity: pure equilibria often do not exist, and computing approximate local min-max points is PPAD-complete \citep{daskalakis2021complexity}. Therefore, the asymmetric Stackelberg equilibrium, where the leader commits to $\mathbf{x}$ and followers best-respond with $\mathbf{Y}$ is adopted~\cite{lin2025twotimescale}. This matches our sequential objective, guarantees well-posedness via our compact $\Y_i$ assumption, and serves as the proper local optimality notion for nonconvex problems by respecting the inherent $\min-\max$ asymmetry.

\begin{algorithm}[H]
\caption{Personalized Local SGDA \textcolor{blue}{\&  VR-SGDA}}
\label{alg:merged_sgda}
\begin{flushleft}
\resizebox{0.9\columnwidth}{!}{
\begin{minipage}{\columnwidth}
\begin{algorithmic}[1]
\STATE \textbf{Input:} synchronization interval $\tau$, iterations $T$, initial $\mathbf{x}_i^{(0)} = \mathbf{x}^{(0)}$, $\mathbf{y}_i^{(0)} \in \mathcal{Y}_i$ for $i \in [n]$; stepsizes $\eta_x, \eta_y$.
\STATE \textcolor{blue}{Initialize snapshots $\bar{\mathbf{x}} = \mathbf{x}^{(0)}$, $\bar{\mathbf{y}}_i = \mathbf{y}_i^{(0)}$ and compute snapshot full gradient $\hat v_i^{(0)} = \nabla_{\mathbf{y}} f_i(\bar{\mathbf{x}}, \bar{\mathbf{y}}_i)$.}
\FOR{$t = 0, \ldots, T-1$ \textbf{in parallel for clients} $i = 1, \ldots, n$}
    \STATE Sample batch $\xi_i^t \sim \mathcal{D}_i$
    \STATE $\mathbf{x}_i^{(t+1)} = \mathbf{x}_i^{(t)} - \eta_x \nabla_{\mathbf{x}} f_i(\mathbf{x}_i^{(t)}, \mathbf{y}_i^{(t)}; \xi_i^t)$
    \STATE Compute base ascent direction $v_i^t = \nabla_{\mathbf{y}} f_i(\mathbf{x}_i^{(t)}, \mathbf{y}_i^{(t)}; \xi_i^t)$
    \STATE \textcolor{blue}{ Apply Variance-Reduced correction using the \emph{same} $\xi_i^t$:}
    \vspace{-3mm}
    \begin{equation*}
        \textcolor{blue}{v_i^t \leftarrow v_i^t - \nabla_{\mathbf{y}} f_i(\bar{\mathbf{x}}, \bar{\mathbf{y}}_i; \xi_i^t) + \hat v_i^{(s)}, \quad \text{where } s = \lfloor t/\tau\rfloor \tau}
    \end{equation*}
    \vspace{-3mm}
    \STATE $\mathbf{y}_i^{(t+1)} = \mathcal{P}_{\mathcal{Y}_i}\!\left(\mathbf{y}_i^{(t)} + \eta_y v_i^t\right)$
    \IF{$\tau \mid (t+1)$}
        \STATE Clients send $\mathbf{x}_i^{(t+1)}$ to the server
        \STATE Server computes $\mathbf{x}^{(t+1)} = \frac{1}{n}\sum_{i=1}^n \mathbf{x}_i^{(t+1)}$ and broadcasts it
        \STATE Each client sets $\mathbf{x}_i^{(t+1)} = \mathbf{x}^{(t+1)}$
        \STATE \textcolor{blue}{ Refresh snapshot $\bar{\mathbf{x}} = \mathbf{x}^{(t+1)}$, $\bar{\mathbf{y}}_i = \mathbf{y}_i^{(t+1)}$}
        \STATE \textcolor{blue}{Recompute full gradient $\hat v_i^{(t+1)} = \nabla_{\mathbf{y}} f_i(\bar{\mathbf{x}}, \bar{\mathbf{y}}_i)$}
    \ENDIF
\ENDFOR
\STATE \textbf{Note:} $\mathbf{y}_i$ is personalized and never communicated. \textcolor{blue}{Blue steps are executed exclusively for the VR-SGDA variant.}
\end{algorithmic}
\end{minipage}
}
\end{flushleft}
\end{algorithm}

\subsection{Problem Formulation and Algorithm}
We study the federated minimax problem over $n$ clients,
\begin{equation}
\min_{\mathbf{x}} \max_{\mathbf{y}_1,\dots,\mathbf{y}_n}
f(\mathbf{x}, \mathbf{Y})
= \frac{1}{n}\sum_{i=1}^{n} f_i(\mathbf{x}, \mathbf{y}_i),
\label{eq:minimax}
\end{equation}
where $\mathbf{Y} \coloneqq (\mathbf{y}_1,\dots,\mathbf{y}_n)$ and $\mathbf{y}_i  \in \Y_i,$
is the personalized maximization variable of client $i$, constrained to a
compact convex set $\Y_i$ determined by that client's own preferences. Adopting the
Stackelberg concept, we seek a stationary point of the value function $\Phi(\mathbf{x})$.
\begin{align}
\Phi(\mathbf{x})=\max_{\mathbf{Y}} f(\mathbf{x},\mathbf{Y})
&=\frac{1}{n}\sum_{i=1}^{n} \max_{\mathbf{y}_i \in \Y_i} f_i(\mathbf{x},\mathbf{y}_i) \notag\\
&=\frac{1}{n}\sum_{i=1}^{n} \Phi_i(\mathbf{x}).
\label{eq:phi}
\end{align}
The second equality makes the
personalized configuration tractable rather than merely well-motivated. Since
$\mathbf{y}_i$ appears only in $f_i$, the joint maximization separates
completely across clients: each client's follower problem is decoupled from
every other's, and $\Phi$ is the average of $n$ independent value functions. No
consensus over $\mathbf{Y}$ is required, none is meaningful, and as we made
precise in Section~\ref{sec:related} that none is properly defined when the $\Y_i$ differ.
The leader's problem is federated; the followers' problems are local.
Algorithm~\ref{alg:merged_sgda} instantiates this directly. The leader variable
$\mathbf{x}$ is updated by local descent and periodically averaged, exactly as in
FedAvg. Each follower variable $\mathbf{y}_i$ is updated by \emph{projected}
local ascent onto $\Y_i$ and never leaves the client. 
The two step sizes are kept in the two-timescale regime
$\eta_x \ll \eta_y$ of \citet{lin2025twotimescale}, which supplies the inner
quasi-stationarity the analysis needs. We analyze two regimes: one
nonconvex in $\mathbf{x}$ and concave in $\mathbf{Y}$, the other nonconvex in $\mathbf{x}$ and PL in $\mathbf{Y}$.

\section{Convergence Analysis}

We first give overview of all the assumptions

\begin{assumption}[Smoothness]\label{assm:lip-sm}
There exists a constant $L > 0$ such that for all $i \in [n]$ and all pairs $(\mathbf{x}_1, \mathbf{y}_1), (\mathbf{x}_2, \mathbf{y}_2) \in \mathbb{R}^{d_x} \times \mathbb{R}^{d_y}$, the gradients are Lipschitz continuous, satisfying $\lVert \nabla f_i(\mathbf{x}_1, \mathbf{y}_1) - \nabla f_i(\mathbf{x}_2, \mathbf{y}_2) \rVert \le L \lVert (\mathbf{x}_1, \mathbf{y}_1) - (\mathbf{x}_2, \mathbf{y}_2) \rVert$.
\end{assumption}

\begin{assumption}[Unbiased and Bounded Variance]\label{assm:bd_var}
For all $i \in [n]$, the local stochastic gradients are unbiased and have bounded variance. That is, for any sample $\xi$, $\mathbb{E}[\nabla_x f_i(\mathbf{x}, \mathbf{y}; \xi)] = \nabla_x f_i(\mathbf{x}, \mathbf{y})$ and $\mathbb{E}[\|\nabla_x f_i(\mathbf{x}, \mathbf{y}; \xi) - \nabla_x f_i(\mathbf{x}, \mathbf{y})\|^2] \le \sigma_x^2$. Identical unbiasedness and variance bounds (denoted $\sigma_y^2$) hold for the stochastic $\mathbf{y}$ gradients. For the NC-C setting, we assume that $\sigma^2_x = \sigma^2_y = \sigma^2$. 
\end{assumption}
\begin{assumption}[Lipschitz Continuity in $\mathbf{x}$]
$f(\mathbf{x}, \mathbf{y})$ is said to be $G_x$-Lipschitz in $\mathbf{x}$ if
$\forall \mathbf{x}, \mathbf{x}' \in \mathbb{R}^{d_x}:
\quad
\lVert f(\mathbf{x}, \mathbf{y}) - f(\mathbf{x}', \mathbf{y}) \rVert 
\le G_x \lVert \mathbf{x} - \mathbf{x}' \rVert.$
\label{assm:lip_cont}
\end{assumption}
\begin{definition}[Gradient Mapping]\label{def:gradmap}
For a step size $\lambda > 0$ and a closed convex set $\mathcal{Y} \subseteq \mathbb{R}^{d_y}$, the gradient mapping of $F$ in $\mathbf{y}$ is defined as $G_{\lambda}(\mathbf{x}, \mathbf{y}) := \frac{1}{\lambda}\big(\mathcal{P}_{\mathcal{Y}}(\mathbf{y} + \lambda \nabla_{\mathbf{y}} F(\mathbf{x}, \mathbf{y})) - \mathbf{y}\big)$, where $\mathcal{P}_{\mathcal{Y}}$ denotes the Euclidean projection onto $\mathcal{Y}$.
\end{definition}

\begin{assumption}[Gradient-Mapping PL~\cite{li2018simple}]\label{assm:pl}
Fix $\lambda \in (0, 1/L]$. The function $F(\mathbf{x}, \mathbf{y})$ satisfies the gradient-mapping Polyak-Łojasiewicz (PL) condition in $\mathbf{y}$ over $\mathcal{Y}$ with modulus $\mu > 0$ if, for all $\mathbf{x} \in \mathbb{R}^{d_x}$ and $\mathbf{y} \in \mathcal{Y}$, it holds that $\tfrac{1}{2} \lVert G_{\lambda}(\mathbf{x}, \mathbf{y}) \rVert^2 \ge \mu\big(F(\mathbf{x}, \mathbf{y}^{*}(\mathbf{x})) - F(\mathbf{x}, \mathbf{y})\big)$, where $\mathbf{y}^{*}(\mathbf{x}) \in \argmax_{\mathbf{y} \in \mathcal{Y}} F(\mathbf{x}, \mathbf{y})$. 

%This assumes that the unconstrained optimum belongs to the set $\mathcal{Y}$. Otherwie, one can find a counter example violating this condition. ($f(y)=-y^2, y \in [1,2] $). 
\end{assumption}

\begin{remark}
When $\Y = \mathbb{R}^{d_y}$ the projection is the identity and
${G}_{\lambda}(\mathbf{x}, \mathbf{y}) = \nabla_{\mathbf{y}} F(\mathbf{x}, \mathbf{y})$
for every $\lambda$, so Assumption~\ref{assm:pl} recovers the classical PL
condition. 
\end{remark}

\begin{assumption}[Non-Convex Concave]
$f(\mathbf{x}, \mathbf{y})$ is assumed to be non-convex in $\mathbf{x}$ and concave in $\mathbf{y}$. This implies for each $\mathbf{x}$ the function is concave in $\mathbf{y}$.
\label{assm:nc_c}
\end{assumption}

\begin{assumption}[Component-wise smoothness]
\label{assump:svrg-smooth}
For each client $i$ and each sample $k \in \mathcal{D}_i$, the per-sample function $F_{i,k}(\mathbf{x}, \mathbf{y})$ has $L$-Lipschitz gradients:
$
\|\nabla F_{i,k}(\mathbf{x}_1, \mathbf{y}_1) - \nabla F_{i,k}(\mathbf{x}_2, \mathbf{y}_2)\| \le L\|\mathbf{x}_1 - \mathbf{x}_2\| + L\|\mathbf{y}_1 - \mathbf{y}_2\|.
$
\end{assumption}

\textit{We define the following quantity to measure the gradient dissimilarity between the local and the global loss~\cite{pillutla2022federated}:}
\begin{equation*}
\zeta_x = 
\sup_{(\mathbf{x}, \mathbf{y}) \in \mathbb{R}^{d_x} \times \mathbb{R}^{d_y}}
\frac{1}{n} \sum_{i=1}^{n}
\left\| \nabla_x f_i(\mathbf{x}, \mathbf{Y}) - \nabla_x f(\mathbf{x}, \mathbf{Y}) \right\|^2
\end{equation*}
As a consequence of the above definition we have 
\begin{equation}
\frac{1}{n} \sum_{i=1}^{n}
\left\| \nabla_x f_i(\mathbf{x}, \mathbf{Y}) - \nabla_x f(\mathbf{x}, \mathbf{Y}) \right\|^2 \leq \zeta_{x}
\end{equation}
\subsection{Convergence Analysis of NC-C Setting}
\noindent
The Moreau envelope~\cite{rockafellar1998variational} of the function $\Phi(\mathbf{x})$ is defined as below

\begin{equation}
\Phi_{1/2L}(\mathbf{x}) \coloneqq 
\min\limits_{\mathbf{z}} \left[ 
\Phi(\mathbf{z}) + L \lVert \mathbf{z} - \mathbf{x} \rVert^2 
\right]
\label{eq:me_def}
\end{equation}
The minimizer is denoted as $\text{prox}_{\phi/2L}(\mathbf{x})$.
\begin{equation}
\text{prox}_{\phi/2L}(\mathbf{x}) \coloneqq 
\arg\min\limits_{\mathbf{z}} \left[ 
\Phi(\mathbf{z}) + L \lVert \mathbf{z} - \mathbf{x} \rVert^2 
\right]
\label{eq:prox_def}
\end{equation}
our aim is to minimize the following potential
\begin{equation}
U = \mathbb{E}{\lVert \nabla \Phi_{1/2L}(\mathbf{x}) \rVert}^2 
\end{equation}
In the nonconvex-concave regime, the value function
$\Phi(\mathbf{x}) = \max_{y_1,\ldots,y_n} f(\mathbf{x}, \mathbf{Y})$ is generally
\emph{non-smooth}, even when each $f_i$ is smooth in both arguments.
Non-smoothness arises because the maximizer
$\mathbf{y}_i^\star(x) = \argmax_{y_i \in \mathcal{Y}_i} f_i(\mathbf{x}, \mathbf{y}_i)$ can
change discontinuously with $\mathbf{x}$ when $f_i$ is merely concave (not
strongly concave) in $\mathbf{y}_i$-the argmax of a linear function over a
polytope jumps between vertices. By Danskin's theorem~\citep{danskin1967theory,bertsekas1999nonlinear}, which would
deliver $\nabla\Phi$ via the envelope theorem, requires the maximizer
to be unique and continuous; mere concavity does not guarantee this. 

Since $\nabla\Phi(\mathbf{x})$ may not exist, the standard stationarity
criterion $\|\nabla\Phi(\mathbf{x})\|^2 \le \epsilon^2$ is not well-defined.
Following \cite{davis2019stochastic,linjinjordan2020}, we instead minimize the
\emph{Moreau envelope} $\Phi_{1/2L}(\mathbf{x})$ which is $L$-smooth regardless of whether $\Phi$ itself is smooth,
%and satisfies $\nabla\Phi_{1/2L}(x) = 2L(x - \mathrm{prox}_{\Phi/2L}(x))$.
and the convergence criterion becomes
$\|\nabla\Phi_{1/2L}(\mathbf{x})\|^2 \le \epsilon^2$, the standard measure
for nonsmooth nonconvex optimization
\citep{davis2019stochastic, linjinjordan2020}.

\begin{theorem}[Convergence for NonConvex-Concave]\label{thm:ncc_convg}
The iterates generated by the projected SGDA Algorithm.~\ref{alg:merged_sgda} satisfy the following convergence bound for the Moreau envelope: $\frac{1}{T}\sum_{t=1}^{T}\mathbb{E}\left[ \left\|\nabla \Phi_{1/2L}\left(\mathbf{x}^{(t-1)}\right)\right\|^2 \right]  \leq \frac{8\Delta_{\phi}}{\eta_x T} + 8L\eta_x\left(G_x^2 + \frac{\sigma^2}{n}\right) + 160L^2\left(2\tau^2\eta_x^2\sigma^2 + \tau^2\eta_x^2\zeta_x^2\right) \nonumber + \frac{8LD^2}{\tau \eta_y}+ 16L\eta_y\sigma^2 + 64L\eta_x G_x\sqrt{G_x^2+\sigma^2}.$ If we select the primal stepsize as: $\eta_x = \min\Bigg\{
  \frac{\epsilon^2}{8L\!\left(G_x^2 + \tfrac{\sigma^2}{n}\right) + 64 L G_x\sqrt{G_x^2+\sigma^2}},\;\;
  \frac{(G_x^2+\sigma^2)\,\epsilon^4}{2560\, D^2 L^3 \sigma^2 (2\sigma^2 + \zeta_x^2)},\\
  \frac{\epsilon^6}{1024\, D^2 L^3 \sigma^2 (G_x^2+\sigma^2)}
\Bigg\}.$
,the dual stepsize is strictly separated as $\eta_y = \min\left(\frac{1}{2L},\frac{\epsilon^2}{16L\sigma^2}\right)$ and $\tau = \frac{D}{\sqrt{\eta_x\,\eta_y\,(G_x^2+\sigma^2)}}.$ then we have:

$\frac{1}{T}\sum_{t=1}^{T}\mathbb{E}\left[ \left\|\nabla \Phi_{1/2L}\left(\mathbf{x}^{(t-1)}\right)\right\|^2 \right] \leq \frac{8\Delta_{\phi}}{\eta_x T} + 4 \epsilon^2.$
This establishes that achieving an $\mathcal{O}(\epsilon^2)$ stationary point requires $T \geq \mathcal{O}\left(\frac{1}{\epsilon^8}\right)$ gradient computations and $\mathcal{O}\left(\frac{1}{\epsilon^4}\right)$ communication rounds.
\end{theorem}

Theorem~\ref{thm:ncc_convg} says that personalization and projection together cost
nothing in the nonconvex-concave regime: the leading term matches the
centralized two-timescale rate of \citep{lin2025twotimescale}, and the federated overhead
enters only through the usual lower-order terms in the local-step count $\tau$
and the heterogeneity $\zeta_x^2$. Two features of the bound are: First, the client-drift contribution splits into two terms carrying
different variance scales, $\sigma^2/n$ and $\sigma^2$:
the first is the averaged $\mathbf{x}$-noise, which the $n$ clients suppress, while the
second is the un-averaged $\mathbf{y}_i$-noise, which they cannot, and this asymmetry is
the analytical signature of the personalized configuration. Second, the compact
domain enters only through projected sets, a genuine problem constant fixed
by the ambiguity set, not a bound assumed on the iterates. 
%Crucially we dont have stale-iterate or snapshot-reconciliation anywhere in SGDA-Algorithm~\ref{alg:merged_sgda}, because locality of $\mathbf{y}_i$eliminates dual client drift by construction, so the mismatched-best-responseproblem that motivates SGDA+ in shared-$y$  analyses~\citep{deng2021local} never arises. 
The detailed proof is provided in the Appendix under Theorem~\ref{app:thm:ncc_convg}.

\subsection{Convergence Analysis of NC-PL}
Under the PL condition~\ref{assm:pl}, it can be shown that $\Phi$ is smooth and no envelope is needed. The stationarity measure is $\|\nabla\Phi(\mathbf{x})\|^2$ directly. We show that $\Phi$ is smooth in the Appendix in Lemma~\ref{lem:danskin_constrained}.

\begin{theorem}[Non-convex PL Convergence]\label{thm:pl_main}
Suppose Assumptions~\ref{assm:lip-sm} to~\ref{assm:pl} hold, and suppose the step sizes satisfy
(a) $\eta_y \le \tfrac{1}{2L}$ , (b) $\tfrac{\eta_x}{\eta_y} \le \tfrac{\mu^2}{16 L^2}$, (c) $10\tau^2 \eta_x^2 L^2 \le \tfrac{1}{2}$ (d) $\eta_x \le \tfrac{1}{\beta}$ 
and (e) $\frac{8\eta_x^2 L^2}{\mu^2\eta_y}\!\left[\frac{L+\beta}{2} + 6\eta_y L^2\right] \le \frac{1}{2}\big(1 - \beta\eta_x\big)$. (a),(b),(c),(d) arises due to smoothness, two-timescale, consenus-stability, outer-smoothness respectively. With $T \ge T_0 = \max\{4L^2,\frac{25\mu^8}{4096L^4}, \frac{\mu^2}{256}\} $.  The iterates of Algorithm~1 satisfy $\frac{1}{T}\sum_{t=0}^{T-1} \mathbb{E}\|\nabla\Phi(\mathbf{x}^t)\|^2 
\le \frac{4\Delta_\Phi}{\eta_x T} + \frac{32 L^2 A_0}{\mu^2 \eta_y T} + \frac{96 L^2 \sigma_y^2}{\mu^2} + \frac{2\beta \eta_x \sigma_x^2}{n} 
+ \frac{32 L^2 \eta_x^2 \sigma_x^2}{\mu^2 \eta_y n}\!\left(\tfrac{L+\beta}{2} + 6\eta_y L^2\right) \\
+ \left(4 L^2 + \tfrac{32 L^4 (\eta_x + 6\eta_y)}{\mu^2 \eta_y}\right)\cdot 10\tau \eta_x^2\!\left[\sigma_x^2\!\left(1 + \tfrac{1}{n}\right) + \zeta_x\right].
$Choosing  $\eta_y = \Theta(1/\sqrt{T})$, $\eta_x = \tfrac{\mu^2}{16L^2}\eta_y
= \Theta(1/\sqrt{T})$, we obtain the rate

$\frac{1}{T}\sum_{t=0}^{T-1} \mathbb{E}\|\nabla\Phi(\mathbf{x}^t)\|^2 \le  O\!\Big(\tfrac{L^2(\Delta_\Phi + A_0) +\mu\sigma_x^2}{\mu^2\sqrt{T}}\Big)
      + {O\!\Big(\tfrac{L^2\sigma_y^2}{\mu^2}\Big)}
      + O\!\Big(\tfrac{L^2\mu^2(\sigma_x^2+\zeta_x)}{T^{3/4}}\Big) + O(\frac{1}{T})$
\end{theorem}

Theorem~\ref{thm:pl_main} (vanilla projected SGDA under gradient-mapping PL) establishes a rate with a non-vanishing $\mathcal{O}\!\left(\frac{L^2 \sigma_y^2}{\mu^2}\right)$ noise floor. This floor arises intrinsically from the personalized-$\mathbf{y}_i$ structure: since $\mathbf{y}_i$ is never communicated, the stochastic-gradient variance of 
y on the inner ascent cannot be reduced through server-side averaging, unlike in shared-y analyses. The floor quantifies the statistical cost of personalization under SGDA. The proof details are provided in the Appendix under Theorem.~\ref{thm:app:pl_main}.

Inspired from SVRG~\citep{Johnson2013AcceleratingSG,reddi2016stochastic,NEURIPS2020_0cc6928e}, we eliminate the floor via a per-client gradient correction on the inner ascent. Each client $i$ maintains a snapshot of its $\mathbf{y}_i$ and $\mathbf{x}_i$ variables at every synchronization round and uses a variance-reduced
gradient estimator during the intervening local updates. Towards this end we propose the correction to the y-updates in the algorithm. Following the standard convention in the SVRG literature Algorithm~2 requires component-wise smoothness of the per-sample functions:

\medskip
Assumption~\ref{assump:svrg-smooth} implies that any batch mean $\tfrac{1}{|B|}\sum_{k \in B}F_{i,k}$ is also $L$-smooth (convex combination of $L$-smooth functions), so the population objective $f_i$ is $L$-smooth as well. In machine learning applications where each ``component'' corresponds to a training data point, this assumption holds whenever the per-sample loss is smooth in the parameters.

\medskip
 The two stochastic gradient evaluations $\nabla_{\mathbf{y}} f_i(\mathbf{x}_i^t, \mathbf{y}_i^t; \xi_i^t)$ and $\nabla_{\mathbf{y}} f_i(\bar{\mathbf{x}}, \bar{\mathbf{y}}_i; \xi_i^t)$ use the \emph{same} realization $\xi_i^t$. This shared randomness is what enables the variance reduction: the two noisy terms are perfectly coupled through $\xi_i^t$, and their difference reduces to a smoothness-bounded spatial change rather than a sum of independent noises.

\begin{theorem}[Non-convex PL Convergence under VR-SGDA]\label{thm:ncpl-svrg}
Suppose Assumptions~\ref{assm:lip-sm} to~\ref{assm:pl} and Assumption~\ref{assump:svrg-smooth} (component-wise smoothness) hold, and suppose the stepsizes and synchronization interval satisfy
(a) $\eta_y \le \tfrac{1}{2L}$, (b) $\tfrac{\eta_x}{\eta_y} \le \tfrac{\mu^2}{16L^2}$, (c) $\eta_x \le \tfrac{1}{\beta},$
(d) $8L^2\tau^2\eta_y^2 \le \tfrac{1}{2}$,
(e) $10\tau^2\eta_x^2 L^2 \le \tfrac{1}{2}$,
(f) $\tfrac{8\eta_x^2 L^2}{\mu^2\eta_y}\!\left[\tfrac{L+\beta}{2} + 6\eta_y L^2\right] \le \tfrac{1}{2}(1 - \beta\eta_x)$.
Where (a),(b),(c),(d) arises because of smoothness, two-timescale, outer smoothness, block length and consensus stability respectively.  Then VR-SGDA Algorithm~\ref{alg:merged_sgda} satisfies
$\frac{1}{T}\sum_{t=0}^{T-1}\mathbb{E}\|\nabla\Phi(\mathbf{x}^t)\|^2
\le  \tfrac{4\Delta_\Phi}{\eta_x T}
+ \tfrac{32 L^2 A_0}{\mu^2\eta_y T}
+ \tfrac{2\beta\eta_x\sigma_x^2}{n}
+ \tfrac{32 L^2\eta_x^2\sigma_x^2}{\mu^2\eta_y n}\!\left[\tfrac{L+\beta}{2} + 6\eta_y L^2\right]
 + \left(4L^2 + \tfrac{32L^4(\eta_x + 24\eta_y)}{\mu^2\eta_y}\right)\cdot 10\tau\eta_x^2[\sigma_x^2(1+1/n) + \zeta_x]
+ \tfrac{384 L^4}{\mu^2}\cdot\tau^2\eta_x^2(G_x^2 + \sigma_x^2/n)
 + \tfrac{192 L^4}{\mu^2}\cdot\left[4\tau^2\eta_y^2 G_y^2 + 16L^2\tau^4\eta_x^2\eta_y^2(G_x^2 + \sigma_x^2/n)\right].$ Choosing $\eta_y = \Theta(1/\sqrt{T})$,
$\eta_x = \tfrac{\mu^2}{16L^2}\,\eta_y = \Theta(1/\sqrt{T})$ (tight on the two-timescale condition~(b)), and $\tau = T^{1/4}$ and $T$ large enough to satisfy all stepsize conditions  ( $T \ge T_0 =  \max\Big(4L^2,\; 256L^4,\; \tfrac{25\mu^8}{4096 L^4}\Big).$), we obtain the asymptotic rate $\frac{1}{T}\sum_{t=0}^{T-1}\mathbb{E}\|\nabla\Phi(\mathbf{x}^t)\|^2
\le  O\!\left(\frac{L^2(\Delta_\Phi + A_0) +\mu \sigma_x^2 + \mu^2 G_x^2 + L^4 G_y^2/\mu^2}{\sqrt{T}}\right)
+ O\!\left(\frac{L^2\mu^2(\sigma_x^2+\zeta_x)}{T^{3/4}}\right) + O(1/T).$
\end{theorem}
Theorem~\ref{thm:ncpl-svrg} restores the $\mathcal{O}(1/\sqrt{T})$ rate with
no additive floor, under projection and personalization both. 
The correction works for the reason identified above: the floor is a
constant proportional to $\sigma_y^2$ that no choice of stepsizes can remove, since it enters the inner-tracking recursion without an $\eta_y$ factor. Any fix must therefore replace $\sigma_y^2$ with a quantity that itself vanishes under the two-timescale regime.

Coupling the current-iterate and snapshot gradients to a common sample
$\xi_i^t$ does exactly this: the two stochastic terms share their randomness,
so their difference is controlled by the iterate drift from the snapshot,
$\|\mathbf{x}_i^t - \bar{\mathbf{x}}\|$ and $\|\mathbf{y}_i^t - \bar{\mathbf{y}}_i\|$,
rather than by the noise level. The residual variance therefore inherits the
drift's decay instead of remaining a constant. This differs
from off-the-shelf variance reduction as  the snapshot is necessarily client-side and over $\mathbf{y}_i$ alone. Since $\mathbf{y}_i$ is never communicated and no server-side full-gradient anchor exists like SCAFFOLD~\citep{karimireddy2020scaffold} style control variates, which correct
drift on the {shared} variable, address a different quantity.
 
%The proof requires component-wise $L$-smoothness (Assumption~\ref{assump:svrg-smooth}), the standard condition throughout the non-convex SVRG literature \citep{reddi2016stochastic,yang2020global}, which our finite-sum applications satisfy. 

The cost is one extra gradient evaluation per client per
step and $\mathcal{O}(d_y)$ local storage, both paid entirely on-device; the
communication profile of Algorithm~\ref{alg:merged_sgda} is unchanged.  This establishes that achieving an ${O}(\epsilon^2)$ stationary point requires $T = {O}(1/\epsilon^4)$ gradient computations and $K = {O}(1/\epsilon^3)$ communication rounds. Our Theorem~\ref{thm:ncpl-svrg} achieves the same leading rate as \citep{pmlr-v162-sharma22c} under the strictly personalized-$\mathbf{y}_i$ setting (no $\mathbf{y}$-averaging), due to per-client VR-SGDA. This closes the personalization/statistics gap identified in Theorem~\ref{thm:pl_main}.
The detailed proof is provided in the appendix under the Theorem~\ref{app:thm:ncpl-svrg}.

Personalizing the ascent keeps each client's maximization variable $\mathbf{y}_i$ entirely local, so it is never transmitted to the server. Only the shared model $\mathbf{x}$ is communicated, reducing per-round communication to the model alone, independent of the adversary's size.

\begin{table}[t]
\centering
\caption{Comprehensive Subpopulation Group Composition of the Colored MNIST Dataset}
\label{tab:cmnist_group_composition}
\small
\resizebox{0.5\textwidth}{!}{
\begin{tabular}{ccccccc}
\toprule
\textbf{Group} & \textbf{\begin{tabular}[c]{@{}c@{}}Shape\\ (True Label)\end{tabular}} & \textbf{\begin{tabular}[c]{@{}c@{}}Color\\ (Spurious)\end{tabular}} & \textbf{Train Count} & \textbf{Train \%} & \textbf{Test Count} & \textbf{Test \%}  \\ 
\midrule
Group 0 & Low (0--4)  & Red   & $\sim$24,000 & 40\%  & $\sim$2,500 & 25\%   \\
Group 1 & Low (0--4)  & Green & $\sim$6,000  & 10\%  & $\sim$2,500 & 25\%   \\
Group 2 & High (5--9) & Red   & $\sim$6,000  & 10\%  & $\sim$2,500 & 25\%  \\
Group 3 & High (5--9) & Green & $\sim$24,000 & 40\%  & $\sim$2,500 & 25\%   \\ 
\bottomrule

\end{tabular}
}
\end{table}
\textit{Scope:} We currently consider full participation setting and   extending this personalized framework to partial client participation introduces challenges centered on asynchronous primal-dual coupling. The issue is that when a dormant client is eventually re-sampled, its local ascent step must be evaluated using a severely stale inner variable $\mathbf{y}_i^{t-d}$ against a global model $\mathbf{x}^t$ that has drifted across the network for $d$ communication rounds. This mismatch potentially breaks the tracking bounds established in the current analysis. 

\section{Experiments}

\subsection{Federated Adversarial Training}
\label{sec:fed-adv}

The main goal of this experiment is to  provide evidence for our modeling claim, personalizing the maximizing parameters rather than a sharing, improves the robustness. We refer to \cite{rusak2020simple} to design experiments for adversarial training. The federated adversarial training objective is formulated as a minimax optimization problem, where the task model minimizes the classification loss while the noise generator maximizes it by producing adversarial perturbations in the embedding space. Formally, the objective is
\begin{equation}
\min_{\theta}\max_{\phi}
\mathcal{L}\!\left(f_{\theta}(\mathbf{x}+G_{\phi}(\boldsymbol{\xi})),\mathbf{y}\right),
\end{equation}
where $f_{\theta}$ denotes the BERT model with parameters $\theta$, $G_{\phi}$ is the noise generator parameterized by $\phi$ (a two layer fully connected network  with ReLU), $\mathbf{x}$ denotes the token embeddings produced by the embedding layer, $y$ is the ground-truth label, and $\boldsymbol{\xi}\sim\mathcal{N}(0,I)$ is a Gaussian noise tensor. During local training on each federated client, the model parameters are updated using stochastic gradient descent (SGD). Conversely, the noise generator solves the inner maximization problem using projected stochastic gradient ascent. 

We evaluate the proposed method on intent classification benchmarks: the 20-class variant of Banking77 . We first select the 20 most frequent intent classes to obtain a balanced subset of the original dataset. The training data are then partitioned among 10 federated clients using a Dirichlet distribution with concentration parameter $\alpha=0.1$. This partitioning induces a highly non-IID data distribution, resulting in significant heterogeneity in the class distributions across clients. For all experiments, federated training is performed for 100 communication rounds with full client participation in every round. 
The BERT model (minimizer) is optimized using SGD with a learning rate of $10^{-3}$, while the adversarial noise generator (maximizer) is optimized using a learning rate of $10^{-1}$. After each communication round, the model parameters are aggregated using the standard Federated Averaging (FedAvg) algorithm.

We compare the proposed personalized adversarial training framework against two baseline methods. The first baseline is \textit{Empirical Risk Minimization (ERM)}, which is standard federated learning objective without any adversarial perturbations. The second baseline employs adversarial training with a \textit{shared noise generator}, where a single generator is collaboratively optimized across all clients and aggregated along with the model parameters during each round. Our approach maintains \textit{personalized noise generator} for every client, allowing each client to learn adversarial perturbations that are tailored to its local data distribution while only the task model is globally aggregated. The results on Banking-77 are shown in Table \ref{tab:nlp_results}.

Performance is evaluated using two complementary metrics. The first is the \emph{average test accuracy}, computed by averaging the test accuracies over all participating clients. The second is the \emph{worst-client test accuracy}, defined as the minimum test accuracy among all clients, which assesses robustness under highly heterogeneous data distributions and reflects the performance on the most challenging client. 

%Local SGDA+ experience collapses as freezing a shared model snapshot anchors the inner adversary to a heavily biased local subset. The unconstrained adversary then over-exploits this stale representation, generating extreme perturbations that destabilize the global model during server aggregation. The controlled comparison between SGDA and SGDA+ is identical except for the snapshot, isolating the snapshot device as the cause.
\begin{table}[htp]
\centering
\caption{{Federated-Adversarial performance evaluated on clean samples across Banking-77 dataset.}}
\label{tab:nlp_results}
\resizebox{\columnwidth}{!}{ % <--- Fits inside a single column
\begin{tabular}{lcccc}
\textbf{Framework} & \textbf{Worst-Client (\%)} & \textbf{Mean (\%)}\\ 
\midrule
ERM (Baseline)             & 34.09 (0.00) & 55.56 (1.02) \\ 
Shared Model (SGDA)        & 34.09 (1.61) & \textbf{55.60} (0.89)  \\ 
%Shared Model (SGDA+)        & 28.66 (3.60) & \textbf{62.30} (1.21)  \\ 
Personalized Model (Ours)  & \textbf{38.18} (1.90) & 54.50 (0.33) \\ 
\bottomrule
\end{tabular}
}
\end{table}

\subsection{Federated Group DRO}
\vspace{-0.05in}
\label{sec:dro-formulation}
We demonstrate the DRO experiment for the NC-C setting, we instantiate Algorithm~\ref{alg:merged_sgda} for federated \emph{group} DRO,
in which robustness is sought against the worst reweighting of predefined
subpopulations on each client. Let the data domain be partitioned into
$G$ disjoint groups (defined by, e.g., a demographic attribute crossed with
the label), and let client $i$ hold $m_{i,g}$ samples in group $g$, with
$\mathcal{S}_i \subseteq \{1,\ldots,G\}$ denoting the set of groups
\emph{observed} on client $i$ (some groups may have zero samples on some
clients). Let $\mathbf{x} = \theta$ be the shared model. The personalized maximization variable is a reweighting
$\mathbf{y}_i = p_i \in \Delta_{|\mathcal{S}_i|}$ over client $i$'s observed groups,
constrained to the per-client $\chi^2$-ambiguity set
$\mathcal{P}_i
  \;=\;
  \Big\{ p_i \in \Delta_{|\mathcal{S}_i|} :\;
    D_{\chi^2}\!\big(p_i \,\|\, u_i\big) \le \rho_i \Big\},
  \qquad
  u_i \;=\; \tfrac{1}{|\mathcal{S}_i|}\mathbf{1},$
where $u_i$ is uniform over the observed groups and $\rho_i$ is a
per-client radius. Restricting the reweighting to $\mathcal{S}_i$ rather
than to the full group set is essential: allocating mass to a group with
zero samples on client $i$ would leak $\chi^2$-budget onto unobservable
directions and inflate the effective adversary. The local objective is the worst-case group-reweighted risk, $f_i(x, p_i)
  \;=\;
  \sum_{g \in \mathcal{S}_i} p_{i,g}\,\widehat{\mathcal{L}}_{i,g}(x),
  \
  \widehat{\mathcal{L}}_{i,g}(x)
  \;=\;
  \tfrac{1}{m_{i,g}}\sum_{j:\, \mathrm{group}(\xi_{i,j}) = g}
    \ell\!\left(x;\xi_{i,j}\right),$ where $\widehat{\mathcal{L}}_{i,g}(x)$ is client $i$'s empirical group-$g$
loss and $\ell(\cdot;\xi)$ is a per-sample loss (cross-entropy in our
experiments), $L$-smooth in $x$. The federated objective is $\min_{x}\;
  \tfrac{1}{n}\sum_{i=1}^n \max_{p_i \in \mathcal{P}_i} f_i(x, p_i) =
  \min_{x}\;
  \tfrac{1}{n}\sum_{i=1}^n
    \max_{p_i \in \mathcal{P}_i}
    \sum_{g \in \mathcal{S}_i} p_{i,g}\,\widehat{\mathcal{L}}_{i,g}(x).$

The derivation of gradient updates and implementation details are in the appendix in Section~\ref {sec:dro-projection}.

The Colored MNIST (CMNIST) dataset is engineered to evaluate optimization robustness against simplicity bias under severe subpopulation shifts. The original digit space is mapped to a binary label where the causal feature is the structural shape categorized as Low ($Y=0$ for digits 0–4) or High ($Y=1$ for digits 5–9), and the spurious feature is the foreground color, painted either Red or Green. The label-color Cartesian product yields $G=4$ subpopulation groups. Training data features an 80\% spurious correlation ($P(\text{Red} \mid Y=0) = 0.8$, $P(\text{Green} \mid Y=1) = 0.8$), creating a shortcut that standard ERM exploits over shape features. Testing severs this to a uniform 50\% split, causing a distribution shift that penalizes color-reliant models on minority trap groups. To simulate federated heterogeneity, training data is partitioned across 10 clients via a Dirichlet distribution ($\alpha=2.0$), ensuring sufficient local support for all groups (Table~\ref{tab:cmnist_group_composition}). We use Resnet-20 as client model for this experiment, with $100$ rounds, with descent and scent learning rates of $10^{-3}$ and $10^{-2}$ respectively.

\begin{table}[htp]
\centering
\caption{Generalization and subpopulation robustness on Federated Colored MNIST (mean $\pm$ std over three seeds). Baseline ERM memorizes the spurious color correlation, collapsing on minority groups (0.19\%). The Shared Model (Global DRO) stays near chance ($\sim$52\%). Our Fully Personalized Min-Max framework leverages decoupled local Stackelberg adversaries to unlearn color biases and extract invariant shape features, achieving a superior 72.37\% worst-group accuracy.}
\label{tab:empirical_results}
\resizebox{0.5\textwidth}{!}{
\begin{tabular}{lcc}
\toprule
\textbf{Framework} & \textbf{Worst-Group Accuracy (\%)} & \textbf{Average Accuracy (\%)} \\ 
\midrule
ERM (Baseline)             & 0.19 (0.20)                        & 53.00 (4.57)                   \\ 
Shared Model(SGDA)               & 52.23 (2.19)                       & 57.23 (0.87)                   \\ 
%Shared Model(SGDA+)               & 62.73 (5.6)                       & 71.40 (1.26)                   \\ 
Personalized Model (Ours)         & \textbf{72.37} (0.88)                       & \textbf{75.63} (1.25)                   \\ 
\bottomrule
\end{tabular}
}
\end{table}
\vspace{-0.1in}
\begin{figure}[htbp]
    \centering
    \includegraphics[width=0.45\textwidth, height=6cm]{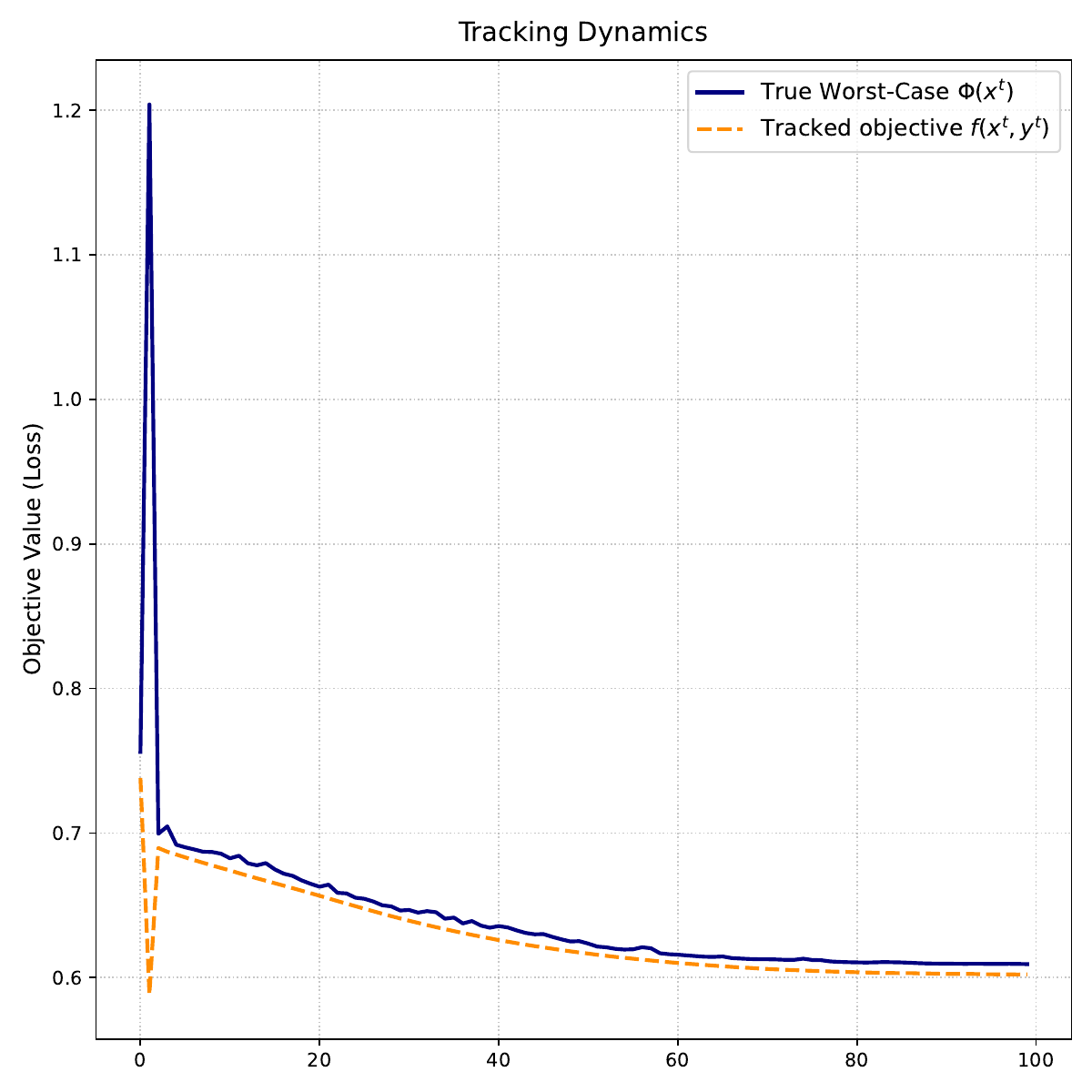}
    \caption{Dual-tracking analysis showing the convergence of $\Phi(x^t)$.}
    \label{fig:dual_tracking}
\end{figure}

Figure~\ref{fig:dual_tracking} tracks how our model improves during training. Initially, the algorithm targets the model's weaknesses by focusing on the hardest data examples, causing a brief, sharp spike in errors. However, the model adapts, reducing this performance gap to near zero. Table~\ref{tab:empirical_results} captures the final outcome of this process: while standard methods completely fail on the hardest data groups, scoring only $0.19\%$, and older shared methods stall at $52.23\%$, our personalized approach forces the model to ignore misleading shortcuts. As a result, it achieves a superior $72.37\%$ accuracy on the absolute hardest cases, alongside a $75.63\%$ average overall score.

\subsection{NC-PL }

To verfy our theory we perform the controlled synthetic experiment to mimic the NC-PL conditions.   
\label{app:ncpl-floor}
 Each client solves $f_i(\mathbf{x},\mathbf{y}_i)
  = -\,a\,\psi\!\big(\mathbf{y}_i - b_i(\mathbf{x})\big) + g(\mathbf{x}),
  \qquad \mathbf{y}_i \in [-R,R]^{d_y},$ where $\psi(\mathbf{u})=\sum_j\big(\tfrac12 u_j^2+\tfrac32\sin^2 u_j\big)$ is
the PL-but-nonconcave function of \citet{karimi2016linear}
($\psi_1''\in[-2,4]$, so $-\psi$ is nonconcave, yet PL holds. $b_i(\mathbf{x})=A_i\tanh(W\mathbf{x})$ is a client-specific moving target that the inner variable must track; and $g(\mathbf{x})=\tfrac{\alpha}{4}\|\mathbf{x}\|^4-\tfrac{\beta}{2}\|\mathbf{x}\|^2$ is nonconvex in $\mathbf{x}$ (double-well potential). Since $\psi_1$ is even and unimodal, the
constrained maximizer is $\mathbf{y}_i^\star(\mathbf{x})=
\Pi_{\mathcal{Y}}(b_i(\mathbf{x}))$ and $\nabla\Phi$ follows in closed form by
Danskin's theorem, so the stationarity measure is computed exactly. 

\begin{figure}[t]
  \centering
  \includegraphics[width=1.0\linewidth]{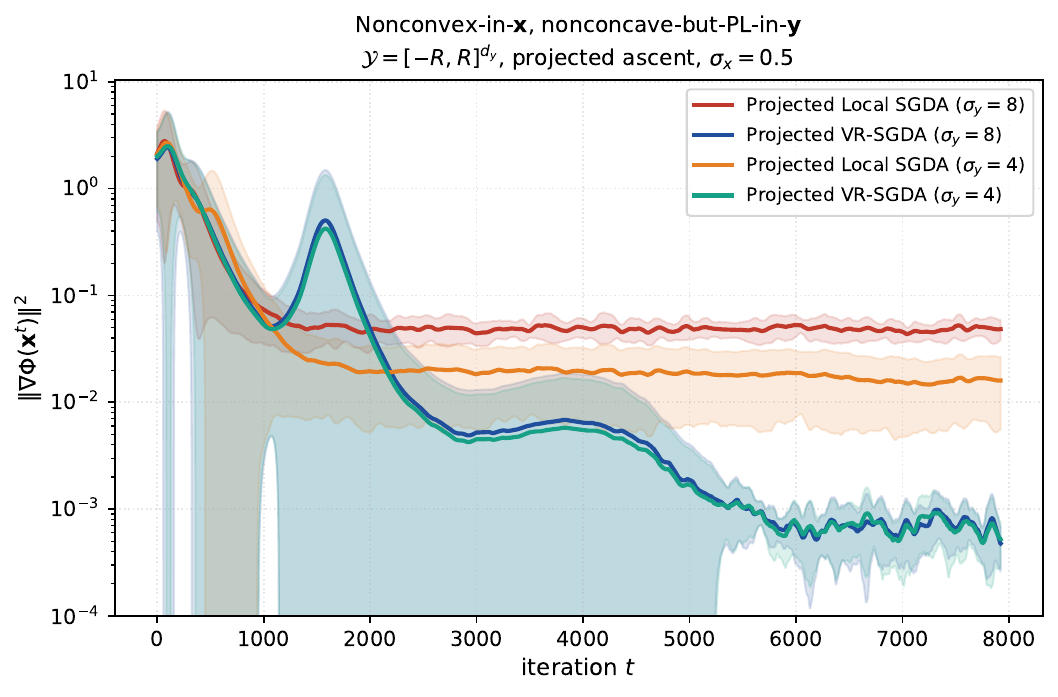}
  \caption{$\|\nabla\Phi(\mathbf{x}^t)\|^2$ for projected Local~SGDA vs.\
  Personalized VR-SGDA (mean of $5$ seeds,
  band $\pm1$ s.d.). SGDA plateaus at the noise floor; VR-SGDA removes it.}
  \label{fig:ncpl-floor}
\end{figure}

The NC-PL conditions are verified in Section.~\ref{nc-pl-verif}.  We use $d_x = 5, d_y = 3,  \text{number of clients} (n)= 10, R = 0.25, \alpha =1, \beta = 2$. $W$ is drawn once at construction as a random Gaussian matrix, $W \in \mathbb{R}^{d_x \times d_x}$ with entries $\sim \mathcal{N}(0, 1/d_x)$. 
To mimic SGD, the inner gradient carries unbiased state-dependent noise
$\widetilde\nabla_y f_i=\nabla_y f_i+\sigma_y\,\xi\odot(\mathbf{1}+\gamma\sin\mathbf{y}_i)$,
$\xi\sim\mathcal{N}(0,I)$, and the outer gradient carries additive noise of
scale $\sigma_x$.  Figure~\ref{fig:ncpl-floor} plots $\|\nabla\Phi(\mathbf{x}^t)\|^2$ for two levels of $\sigma_y$. Under the gradient-mapping PL condition, vanilla projected SGDA settles at a nonzero, stepsize-invariant floor, whereas Personalized VR-SGDA removes the $\sigma_y$-dependent term and settles roughly two orders of magnitude lower. 

The bump in the middle of the plot appears only on the VR-SGDA curves because it lies on the part of the descent that only VR-SGDA reaches. It comes from the shape of the double-well objective: as the iterate moves toward the minimizer it passes through a region where the gradient temporarily grows before shrinking again, producing a brief rise in the curve. VR-SGDA travels through this region on its way down, so the rise shows up. SGDA has already plateaued at its floor by then and never travels through that part of the landscape, so its curve stays flat and shows no bump. 

\iffalse
\begin{table}[htp]
\centering
\caption{{Clean and adversarial performance on CLINC-151.}
We pick a lower learning rate of 0.01, as suggested by our theory for svrg varaiant.
The personalized model without variance reduction exhibits a lower worst-group accuracy despite achieving competitive average performance. This behavior is consistent with our theoretical analysis, where the convergence rate contains a non-vanishing optimization floor arising from gradient variance. By incorporating SVRG update, this optimization floor is eliminated, enabling more accurate convergence to the desired solution, validating our theoretical analyses on personalized projected SVRG.}
\label{tab:adv_results}
\resizebox{\columnwidth}{!}{ % <--- Fits inside a single column
\begin{tabular}{lcccc}
\toprule
 & \multicolumn{2}{c}{\textbf{Clean}} & \multicolumn{2}{c}{\textbf{Adversarial}} \\
\cmidrule(lr){2-3} \cmidrule(lr){4-5}
\textbf{Framework / Model} & \textbf{Worst-Client (\%)} & \textbf{Mean (\%)} & \textbf{Worst-Client (\%)} & \textbf{Mean (\%)} \\ 
\midrule
ERM (Baseline)             &  31.46 (1.08) & 76.32 (0.17)  & 35.19 (0.13)  & 75.88 (0.08) \\ 
Shared Model               & 30.79 (0.57) & \textbf{76.61} (0.12) & 35.19 (0.14)  & 75.93 (0.05) \\ 
Personalized Model  & 29.80 (1.43)  & 76.13 (0.11)  & 34.89 (0.11)  & 76.13 (0.04) \\ 
Personalized+SVRG & \textbf{33.44} (1.49) & 76.46 (0.14) & \textbf{36.08} (0.09) & \textbf{76.52} (0.04)\\
\bottomrule
\end{tabular}
}
\end{table}
\fi 

\section{Conclusion}

We introduce a practical framework for federated minimax problems by keeping the maximization variables personalized to each client and using projected ascent.  This approach  succeeds in standard nonconvex-concave settings without requiring snapshot mechanism. In the nonconvex-pl regime, the standard ascent projection step creates a persistent noise floor, a baseline level of error that cannot be fixed simply by tuning the learning rate. We empirically demonstrate that this is a  limitation of the  algorithm. To fix this, we introduce Personalized VR-SGDA. By using local snapshots of the variables, our method smoothly phases out this constant noise and restores the optimal convergence speed. Finally, this solution works perfectly when all clients participate. In the partial participation setting when inactive clients are sampled, their outdated local variables severely clash with the newly updated global model, potentially breaking the current techniques and the extension remains open.

\iffalse
\section*{AI Use Declaration}
In accordance with AISTATS 2027 policy, 
We disclose the use of generative AI in preparing this manuscript. AI assistance was used to enhance author-written
text (rephrasing, compression, and readability); to help with presentation,
including formatting tables and generating plotting code; to assist with code implementation for correctness and readability; and to verify algebra as an additional check on top of human verification. AI was not used for idea generation, proof techniques, experiment design, or the interpretation of results and conclusions. The core writing, research originality, and all
technical content of the manuscript are the authors' own.
\fi

\bibliography{main}
\bibliographystyle{plainnat}
\clearpage

\clearpage
\onecolumn

%%%%%%%%%%%%%%%%%%%%%%%%%%%%%%%%%%%%%%%%%%%%%%%%%%%%%%%%%%%%
\appendix

\section{Technical appendices and supplementary material}

\input{supplementary_file}

%%%%%%%%%%%%%%%%%%%%%%%%%%%%%%%%%%%%%%%%%%%%%%%%%%%%%%%%%%%%

%\newpage
%\input{checklist.tex}

\end{document}

%% file: supplementary_file.tex
\addtocontents{toc}{\protect\setcounter{tocdepth}{2}}
\tableofcontents

\subsection{Notation}
\label{sec:notation}

\paragraph{Problem and variables.}
$n$: number of clients; $[n] = \{1,\dots,n\}$.
$\mathbf{x} \in \mathbb{R}^{d_x}$: shared minimization variable (model).
$\mathbf{y}_i \in \mathcal{Y}_i$: personalized maximization variable of client $i$;
$\mathbf{Y} = (\mathbf{y}_1,\dots,\mathbf{y}_n)$.
$\mathcal{Y}_i$: compact convex constraint set of client $i$;
$D = \operatorname{diam}(\mathcal{Y}_i)$.

\paragraph{Group DRO.}
$G$: number of groups (subpopulations).
$\mathcal{S}_i \subseteq \{1,\ldots,G\}$: groups observed on client $i$ (support).
$m_{i,g}$: samples in group $g$ on client $i$.
$p_i \in \Delta_{|\mathcal{S}_i|}$: personalized group reweighting.
$u_i = \mathbf{1}/|\mathcal{S}_i|$: uniform reference on client $i$'s support.
$\rho_i$: per-client $\chi^2$-ambiguity radius;
$r^2 = \rho_i / |\mathcal{S}_i|$: rescaled radius.
$\mathcal{P}_i = \{p \in \Delta_{|\mathcal{S}_i|} : D_{\chi^2}(p \| u_i) \le \rho_i\}$:
ambiguity set, with $D_{\chi^2}(p\|u) = |\mathcal{S}_i|\|p - u_i\|^2$.
$\widehat{\mathcal{L}}_{i,g}(x) = \tfrac{1}{m_{i,g}}\sum_{j:\mathrm{group}(\xi_{i,j})=g} \ell(x;\xi_{i,j})$:
empirical group-$g$ loss.
$L_i = (\widehat{\mathcal{L}}_{i,g})_{g \in \mathcal{S}_i}$: group-loss vector.
$\Pi_{\Delta_C}$: simplex projection;
$\Pi_{\mathcal{P}_i}$: projection onto the $\chi^2$-restricted simplex.

\paragraph{Objectives.}
$f_i(\mathbf{x}, \mathbf{y}_i)$: local objective of client $i$.\\
$f(\mathbf{x}, \mathbf{Y}) = \tfrac{1}{n}\sum_i f_i(\mathbf{x}, \mathbf{y}_i)$: global objective.
$\Phi(\mathbf{x}) = \tfrac{1}{n}\sum_i \Phi_i(\mathbf{x})$: value function, with
$\Phi_i(\mathbf{x}) = \max_{\mathbf{y}_i \in \mathcal{Y}_i} f_i(\mathbf{x}, \mathbf{y}_i)$.\\
$\mathbf{y}_i^\star(\mathbf{x}) = \arg\max_{\mathbf{y}_i \in \mathcal{Y}_i} f_i(\mathbf{x}, \mathbf{y}_i)$:
per-client maximizer.
$\Phi_{1/2L}(\mathbf{x}) = \min_z \{\Phi(z) + L\|z - x\|^2\}$: Moreau envelope (NC-concave only).
$\Delta_\Phi$: initialization gap.

\paragraph{Personalized Local SGDA and Personalized VR-SGDA}.
$t = 0,\ldots,T{-}1$: iteration index; $T$: total gradient steps.
$\tau$: synchronization interval; $K = T/\tau$: communication rounds.
$r$: round index; $s = r\tau$: block-start index.
$\eta_x, \eta_y$: primal (slow) and dual (fast) stepsizes.
$\mathbf{x}_i^{(t)}, \mathbf{y}_i^{(t)}$: local iterates;
$\mathbf{x}^{(t)} = \tfrac{1}{n}\sum_i \mathbf{x}_i^{(t)}$: virtual average. $\xi_i^t$: stochastic sampling realization. $\bar{\mathbf{x}}, \bar{\mathbf{y}}_i$: snapshots at synchronization boundary.
$\mu_y = \nabla_y f(\bar{x}, \bar{y})$: full-batch anchor gradient.
$\widetilde{\nabla}_y^{\mathrm{VR}} = \nabla_y f(x^t, y^t; \xi) - \nabla_y f(x^t, \bar{y}; \xi) + \mu_y$:
variance-reduced inner gradient.

\paragraph{Assumptions and constants.}
$L$: smoothness (A1).
$\sigma_x^2, \sigma_y^2$: gradient variance bounds (A2).
$G_x$: Lipschitz bound in $\mathbf{x}$ (A3).
$\mu$: gradient-mapping PL modulus (A4);
$\kappa = L/\mu$: condition number;
$\beta = L(1{+}\kappa)$: smoothness of $\Phi$ under PL.
$\zeta_x^2$: gradient dissimilarity bound.

\paragraph{Convergence quantities.}
$\gamma^t = \tfrac{1}{n}\sum_i \|\mathbf{x}^t - \mathbf{x}_i^t\|^2$: consensus error.
$\Delta_i^t = \Phi_i(\mathbf{x}^t) - f_i(\mathbf{x}^t, \mathbf{y}_i^t)$: per-client tracking gap.
$A_t = \mathbb{E}[\Phi(\mathbf{x}^t) - F(\mathbf{x}^t, \mathbf{Y}^t)]$: averaged tracking gap.
$\epsilon$: target stationarity accuracy.

\subsection{DRO Implementation details}
\label{sec:dro-projection}
For clarity and reproducibility, we clearly derive the DRO solution that is being implemented.\\
\textit{SGDA Descent step ( Algorithm~\ref{alg:merged_sgda}).}
The gradient with respect to the model is a $p_i$-reweighted mixture of
per-group empirical gradients,
\begin{equation}\label{eq:groupdro-desc-grad}
\begin{aligned}
  \nabla_x f_i(x, p_i)
 &  \;=\;
  \sum_{g \in \mathcal{S}_i}
    p_{i,g}\,\nabla_x \widehat{\mathcal{L}}_{i,g}(x)\\
  &\;=\;
  \sum_{g \in \mathcal{S}_i}
    \tfrac{p_{i,g}}{m_{i,g}}
    \sum_{j:\,\mathrm{group}(\xi_{i,j})=g}
      \nabla_x \ell\!\left(x;\xi_{i,j}\right).
\end{aligned}      
\end{equation}
In practice, the stochastic version is computed on a mini-batch by
weighting each sample $j$ by $p_{i,\mathrm{group}(\xi_{i,j})} / m_{i,\mathrm{group}(\xi_{i,j})}$,
which is unbiased for \eqref{eq:groupdro-desc-grad}. Then
\begin{equation}\label{eq:groupdro-desc-step}
  \mathbf{x}_i^{(t+1)}
  \;=\;
  \mathbf{x}_i^{(t)}
  \;-\; \eta_x\,
    \widetilde{\nabla}_x f_i\!\left(\mathbf{x}_i^{(t)}, p_i^{(t)};\,\xi_i^t\right),
\end{equation}
which increases the model's weight on groups the personalized adversary has
concentrated mass on, and does so client-locally between synchronizations.

\textit{SGDA-Ascent step (Algorithm~\ref{alg:merged_sgda}).}
The gradient with respect to the group reweighting is the vector of per-group
empirical losses,
\begin{equation}\label{eq:groupdro-asc-grad}
  \nabla_{p_i} f_i(x, p_i)
  \;=\;
  \Big(\widehat{\mathcal{L}}_{i,g}(x)\Big)_{g \in \mathcal{S}_i}
  \;\in\; \mathbb{R}^{|\mathcal{S}_i|},
\end{equation}
so ascent on $p_i$ moves mass toward groups on which the current model
incurs the highest loss. projected ascent step of SGDA Algorithm~\ref{alg:merged_sgda} is the

\begin{equation}\label{eq:groupdro-asc-step}
  p_i^{(t+1)}
  \;=\;
  \mathcal{P}_{\mathcal{P}_i}\!\left(\, p_i^{(t)}
    \;+\; \eta_y\,
    \widetilde{\nabla}_{p_i} f_i\!\left(\mathbf{x}_i^{(t)}, p_i^{(t)};\,\xi_i^t\right)\right),
\end{equation}
where $\mathcal{P}_{\mathcal{P}_i}$ denotes projection onto the support-restricted
$\chi^2$-ball-simplex ambiguity set defined in Section~\ref{sec:dro-formulation}. This
projection couples the probability simplex and the $\chi^2$-ball centred at
the uniform-on-support $u_i$; it has no coordinate-wise closed form.

Because $p_i$ is personalized, maintained locally on client $i$ and never
communicated, each client's ascent step operates on its own
$|\mathcal{S}_i|$-dimensional group simplex, with a potentially different radius $\rho_i$.

The projection $\mathcal{P}_Y$ appearing in  Algorithm~\ref{alg:merged_sgda} is, for the DRO it is projection onto the
$\chi^2$-ball-restricted simplex
\[
  \mathcal{P}_i \;=\;
  \Big\{\, q \in \mathbb{R}^{|\mathcal{S}_i|}\,:\; q \ge 0,\; \mathbf{1}^\top q = 1,\;
  D_{\chi^2}(q \,\|\, u_i) \le \rho_i \,\Big\},
\]
where $u_i = \mathbf{1}/|\mathcal{S}_i|$ denotes the uniform reference distribution on
client $i$ and $D_{\chi^2}(q\|u_i) = |\mathcal{S}_i|\|q - u_i\|^2$. The projection couples
two constraints, the probability simplex $\Delta_{|\mathcal{S}_i|}$ and the
$\chi^2$-ball centred at $u_i$ and admits no coordinate-wise closed form.
We derive it explicitly via the KKT conditions and because the derivation
is the same on every client, we suppress the client index $i$ throughout this
subsection and write $C := |\mathcal{S}_i|$, $u := u_i$, and $r^2 := \rho_i / C$ so that
the $\chi^2$-constraint reads $\|q - u\|^2 \le r^2$.

Given the raw ascent iterate $\tilde p := p^{t} + \eta_y \nabla_p f(x, p^t)$
(where $\nabla_p f$ is the vector of per-group losses), the projection step in the SGDA
Algorithm~\ref{alg:merged_sgda} solves

\begin{equation}
\label{eq:dro-proj-problem}
  p^{\star} = \operatorname*{argmin}_{q \in \mathbb{R}^{C}}\;
  \tfrac{1}{2}\,\|q - \tilde{p}\|^2
  \quad\text{s.t.}\quad
  q \ge 0,\; \mathbf{1}^\top q = 1,\; \|q - u\|^2 \le r^2 .
\end{equation}
This is a convex quadratic program with linear equality, non-negativity, and a
single quadratic ball constraint. It is feasible ($u$ itself satisfies all
three constraints) and strictly convex, so $p^\star$ is unique.

Introduce dual variables $\mu \in \mathbb{R}$ for the sum-to-one constraint,
$\lambda \in \mathbb{R}_{\ge 0}^{C}$ for the non-negativity constraints, and
$\nu \ge 0$ for the $\chi^2$-ball constraint. The Lagrangian is
\[
  \mathcal{L}(q,\mu,\lambda,\nu)
  \;=\;
  \tfrac{1}{2}\|q - \tilde p\|^2
  \;-\; \mu\big(\mathbf{1}^\top q - 1\big)
  \;-\; \lambda^\top q
  \;+\; \tfrac{\nu}{2}\big(\|q - u\|^2 - r^2\big) .
\]
Stationarity in $q$ gives, coordinate-wise,
\[
  q_c - \tilde p_c - \mu - \lambda_c + \nu(q_c - u_c) \;=\; 0
  \quad\Longleftrightarrow\quad
  (1 + \nu)\,q_c \;=\; \tilde p_c + \mu + \nu u_c + \lambda_c ,
\]
so
\begin{equation}\label{eq:dro-kkt-q}
  q_c
  \;=\;
  \frac{\tilde p_c + \mu + \nu u_c + \lambda_c}{1 + \nu} .
\end{equation}
Complementary slackness $\lambda_c\,q_c = 0$ with $\lambda_c \ge 0$ and
$q_c \ge 0$ enforces $\lambda_c = 0$ on the active support ($q_c > 0$) and
$\lambda_c \ge 0$ off-support ($q_c = 0$), giving the positive-part form
\[
  q_c \;=\; \frac{1}{1+\nu}\Big(\tilde p_c + \mu + \nu u_c\Big)_{\!+} .
\]

The KKT form \eqref{eq:dro-kkt-q} can be rewritten in a more useful shape.
Substituting $u_c = 1/C$ and rearranging,
\[
  q_c
  \;=\;
  u_c \;+\; \frac{1}{1+\nu}\big(\tilde p_c - u_c\big) \;+\; \frac{\mu + \lambda_c}{1+\nu} .
\]
The scalar $\mu$ and the non-negativity multipliers $\lambda_c$ are exactly
what the simplex projection determines: the projection
$\Pi_{\Delta_C}(v)$ of a vector $v$ onto the probability simplex has the
form $\big(v_c - \mu^\star\big)_+$ with $\mu^\star$ chosen by the
sum-to-one constraint. It follows that the KKT solution to
\eqref{eq:dro-proj-problem} equals
\begin{equation}\label{eq:dro-shrinkage}
  q^\star \;=\;
  \Pi_{\Delta_C}\!\Big(\, u \;+\; \tfrac{1}{1+\nu}(\tilde p - u)\,\Big) ,
\end{equation}
that is: shrink $\tilde p$ toward the uniform distribution $u$ by a factor
$\tfrac{1}{1+\nu}$, then project onto the simplex. The dual variable $\nu \ge 0$
is determined by the $\chi^2$-constraint via complementary slackness
$\nu\big(\|q^\star - u\|^2 - r^2\big) = 0$: either the ball constraint is slack
($\nu = 0$, no shrinkage) or it is active ($\nu > 0$, shrinkage tuned so the
constraint binds).

Introduce the scalar shrinkage parameter
\[
  t \;:=\; \frac{\nu}{1+\nu} \;\in\; [0,1),
\]
so that $\tfrac{1}{1+\nu} = 1 - t$. Then \eqref{eq:dro-shrinkage} becomes
\begin{equation}\label{eq:dro-qstar-t}
  q^\star(t) \;=\; \Pi_{\Delta_C}\!\Big(\, u + (1-t)(\tilde p - u) \,\Big) ,
\end{equation}
with $t = 0$ corresponding to the unshrunk case ($\nu = 0$) and $t \to 1$
corresponding to full collapse toward $u$. The projection map is now
parameterized by a single scalar on the bounded interval $[0,1]$.

Define
$g(t) := \|q^\star(t) - u\|^2$.
As $t$ increases from $0$ to $1$, the pre-projection vector
$u + (1-t)(\tilde p - u)$ moves linearly from $\tilde p$ (at $t=0$) toward
$u$ (at $t=1$). Because the projection onto a convex set is non-expansive
(1-Lipschitz), $g(t)$ is non-increasing in $t$, with $g(1) = 0$. Consequently:
\begin{itemize}
  \item If $g(0) = \|\Pi_{\Delta_C}(\tilde p) - u\|^2 \le r^2$, the
    $\chi^2$-constraint is already satisfied by the simplex projection alone
    and $p^\star = \Pi_{\Delta_C}(\tilde p)$ (with $t^\star = 0$, $\nu = 0$).
  \item Otherwise $g(0) > r^2 > g(1) = 0$, and by monotonicity there is a
    unique $t^\star \in (0,1]$ with $g(t^\star) = r^2$. This $t^\star$ can be
    found by bisection on $[0,1]$.
\end{itemize}

{\textit{Exact Solution for computing the tracking error}}\\

Given a group-loss vector $L = (\widehat{\mathcal{L}}_{i,g}(x))_{g \in
\mathcal{S}_i} \in \mathbb{R}^C$ (with $C = |\mathcal{S}_i|$), solve
\begin{equation}\label{eq:linmax-problem}
  p^\star \;=\; \argmax_{q \in \mathbb{R}^C}\; \langle L, q\rangle
  \quad\text{s.t.}\quad
  q \ge 0,\;\; \mathbf{1}^\top q = 1,\;\; \|q - u\|^2 \le r^2,
\end{equation}
where $u = \mathbf{1}/C$ and $r^2 = \rho/C$ as before. This is a
linear program over the intersection of the simplex and a Euclidean
ball, a convex program with a unique maximizer (the feasible set is
strictly convex due to the ball constraint).

\textit{KKT conditions.}
Since we are \emph{maximizing} a linear objective, equivalently
\emph{minimizing} $-\langle L, q\rangle$, the Lagrangian is
\[
  \mathcal{L}(q, \mu, \lambda, \nu)
  \;=\;
  -\langle L, q\rangle
  \;-\; \mu\big(\mathbf{1}^\top q - 1\big)
  \;-\; \lambda^\top q
  \;+\; \tfrac{\nu}{2}\big(\|q - u\|^2 - r^2\big),
\]
with $\mu \in \mathbb{R}$ for sum-to-one, $\lambda_c \ge 0$ for
non-negativity, and $\nu \ge 0$ for the $\chi^2$-ball. Stationarity
$\nabla_q \mathcal{L} = 0$ gives
\[
  -L_c - \mu - \lambda_c + \nu(q_c - u_c) \;=\; 0
  \quad\Longrightarrow\quad
  \nu\, q_c \;=\; L_c + \mu + \lambda_c + \nu u_c.
\]
For $\nu > 0$ (constraint active), dividing through:
\begin{equation}\label{eq:linmax-kkt-q}
  q_c \;=\; u_c \;+\; \frac{L_c}{\nu}
          \;+\; \frac{\mu + \lambda_c}{\nu}.
\end{equation}

As in the projection derivation,
the terms $\mu$ and $\lambda_c$ are exactly the Lagrange multipliers
that a simplex projection would produce. Grouping the terms
involving $L_c$ and $u_c$ separately from the simplex multipliers:
\[
  q_c \;=\; \underbrace{u_c + \frac{L_c}{\nu}}_{v_c(\nu)}
         \;+\; \frac{\mu + \lambda_c}{\nu}.
\]
For fixed $\nu > 0$, the remaining KKT conditions (complementary
slackness $\lambda_c q_c = 0$, feasibility $\sum_c q_c = 1$,
$q_c \ge 0$, $\lambda_c \ge 0$) are identical to the KKT conditions
for projecting $v(\nu)$ onto $\Delta_C$. Therefore
\begin{equation}\label{eq:linmax-shrinkage}
  p^\star \;=\; \Pi_{\Delta_C}\!\Big(\, u + \tfrac{1}{\nu}\, L \,\Big),
\end{equation}
with $\nu > 0$ chosen so the $\chi^2$-constraint binds:
$\|p^\star - u\|^2 = r^2$.

Define $s := 1/\nu \in (0, \infty)$ so that
\begin{equation}\label{eq:linmax-s}
  p^\star(s) \;=\; \Pi_{\Delta_C}\!\big(\, u + s\, L \,\big).
\end{equation}
The parameter $s$ controls how aggressively the solution points in the
direction of $L$: at $s = 0$ we get $p^\star = u$ (uniform, zero
objective improvement), and as $s \to \infty$ the solution approaches
the unconstrained simplex maximizer $\Pi_{\Delta_C}(L)$ (all mass on
the highest-loss group).

\textit{Two cases.} 
\begin{itemize}
  \item {Constraint inactive ($\nu = 0$).} If
    $\|\Pi_{\Delta_C}(L) - u\|^2 \le r^2$, then the unconstrained
    simplex maximizer already lies inside the $\chi^2$-ball. In this
    case $p^\star = \Pi_{\Delta_C}(L)$ and the $\chi^2$-constraint
    does not bind. In the code this is checked by evaluating
    $\Pi_{\Delta_C}(u + s_{\max} L)$ for a large $s_{\max}$.
  \item {Constraint active ($\nu > 0$).} Otherwise, there
    exists a unique $s^\star \in (0, \infty)$ such that
    $\|p^\star(s^\star) - u\|^2 = r^2$, found by bisection.
\end{itemize}

Define $h(s) := \|p^\star(s) - u\|^2 = \|\Pi_{\Delta_C}(u + sL) - u\|^2$.
As $s$ increases from $0$, the pre-projection vector $u + sL$ moves
away from $u$ in the direction of $L$, and by non-expansiveness of
projection:
\[
  h(s) \;=\; \|\Pi_{\Delta_C}(u + sL) - \Pi_{\Delta_C}(u)\|^2
       \;\le\; \|sL\|^2 \;=\; s^2\|L\|^2,
\]
so $h(0) = 0$ and $h$ is non-decreasing. (More precisely, $h$ is
non-decreasing because scaling $s$ up moves the pre-projection
vector monotonically away from $u$, and projection onto a convex
set containing $u$ can only increase the distance from $u$.)
Therefore:
\begin{itemize}
  \item $h(0) = 0 < r^2$, and
  \item for sufficiently large $s$, $h(s) \ge r^2$ (unless $L = 0$,
    in which case $p^\star = u$ trivially).
\end{itemize}
By monotonicity, the equation $h(s) = r^2$ has a unique root
$s^\star > 0$, which bisection on $[0, s_{\max}]$ finds to arbitrary
precision.

\subsection{Verification of NC-PL for the synthetic FL problem}
\label{nc-pl-verif}
\textit{The synthetic inner function is nonconcave but PL.}
Write the inner objective at fixed $\mathbf{x}$ in the
shifted variable $\mathbf{u}=\mathbf{y}-b_i(\mathbf{x})$. Since
$\psi(\mathbf{u})=\sum_{j}\psi_1(u_j)$ is separable with
$\psi_1(v)=\tfrac12 v^2+\tfrac32\sin^2 v$, both claims reduce to the scalar
function $\psi_1$.

\emph{Nonconcavity.} The maximization objective is $-\psi_1$, with
\[
  \psi_1''(v) = 1 + 3\cos(2v) \in [-2,\,4].
\]
Since $\psi_1''(v)<0$ whenever $\cos(2v)<-\tfrac13$, $-\psi_1$ has regions of positive curvature and is therefore
\emph{not concave}; in particular it is not strongly concave. The same Hessian
bound gives the smoothness constant $L_y=a\sup_v|\psi_1''(v)|=4a$.

 We show $\psi_1$ satisfies the PL inequality
$\tfrac12\psi_1'(v)^2 \ge \mu_1\big(\psi_1(v)-\psi_1^\star\big)$ with
$\psi_1^\star=\psi_1(0)=0$, which transfers to $f_i$ with modulus $\mu=a\mu_1$.

 With $\psi_1'(v)=v+\tfrac32\sin(2v)$: for $0<v\le\tfrac{\pi}{2}$ we have
$\sin(2v)\ge0$, so $\psi_1'(v)\ge v>0$; for $v>\tfrac{\pi}{2}>\tfrac32$,
$\psi_1'(v)\ge v-\tfrac32>0$. Hence $\psi_1'>0$ on $(0,\infty)$, and by
oddness $\psi_1'<0$ on $(-\infty,0)$, so $v=0$ is the unique stationary point
and the global minimum. 

Let $R(v):=\tfrac12\psi_1'(v)^2/\psi_1(v)$ for $v\neq0$.
\begin{itemize}
  \item For $|v|\ge3$: $|\psi_1'(v)|\ge|v|-\tfrac32\ge\tfrac{|v|}{2}$ and
    $\psi_1(v)\le\tfrac12 v^2+\tfrac32\le\tfrac23 v^2$, so
    $R(v)\ge\tfrac12(\tfrac{v}{2})^2/(\tfrac23 v^2)=\tfrac{3}{16}=0.1875$.
  \item For $0<|v|<3$: $R$ is continuous and strictly positive (numerator and
    denominator are smooth, the denominator positive), and extends
    continuously to $v=0$ with $R(0^+)=4$ since
    $\psi_1(v)=2v^2+O(v^4)$, $\psi_1'(v)=4v+O(v^3)$. Thus $R$ attains a
    positive minimum on the compact set $[-3,3]$.
\end{itemize}
Taking the smaller of the two gives $\mu_1=\inf_v R(v)>0$. 
%Numerically $\mu_1=0.088$, attained at $v\approx2.20$ (inside the compact regime, consistent with the analytic bound $0.1875$ holding outside it). 
Hence $\psi_1$, and therefore the inner objective, satisfies PL with $\mu=a\mu_1$, while being nonconcave.

\textit{Gradient-mapping PL holds for appropriately large $R$.}
We show that when the box radius $R$ is large enough that the unconstrained
inner maximizer lies strictly interior, the gradient-mapping PL condition on
$\mathcal{Y}=[-R,R]^{d_y}$ coincides with the ordinary PL condition, which we
established for $\psi_1$.

\emph{Setup.} At fixed $\mathbf{x}$, write the inner objective in the shifted
variable $\mathbf{u}=\mathbf{y}-b_i(\mathbf{x})$. The maximization objective
is $-\psi(\mathbf{u})$, whose unconstrained maximizer is $\mathbf{u}=0$, i.e.
$\mathbf{y}=b_i(\mathbf{x})$. The gradient mapping with stepsize $\lambda$ is
\[
  \mathcal{G}_\lambda(\mathbf{y})
  = \tfrac{1}{\lambda}\big( \Pi_{\mathcal{Y}}
      (\mathbf{y}+\lambda\nabla_{\mathbf{y}}f_i(\mathbf{x},\mathbf{y}))- \mathbf{y} \big),
  \qquad
  \nabla_{\mathbf{y}}f_i = -a\,\psi'\!\big(\mathbf{y}-b_i(\mathbf{x})\big).
\]

\emph{Interior condition.} Let $B := \sup_{\mathbf{x},i,j}|b_{i,j}(\mathbf{x})|$;
since $b_i(\mathbf{x})=A_i\tanh(W\mathbf{x})$ and $\tanh\in(-1,1)$, $B$ is
finite and bounded by $\max_i\|A_i\|_\infty$. The per-coordinate gradient is
bounded, $|\nabla_{y_j}f_i|\le a\,G_\psi$ with
$G_\psi:=\sup_{|v|\le 2R}|\psi_1'(v)|<\infty$. Fix
any $\lambda\le 1/L_y=1/(4a)$ and choose
\begin{equation}\label{eq:R-condition}
  R \;>\; B + \lambda\,a\,G_\psi.
\end{equation}

Under \eqref{eq:R-condition} the constrained maximizer is the interior point
$\mathbf{y}_i^\star(\mathbf{x})=b_i(\mathbf{x})$, with
$|y^\star_{i,j}|\le B < R$. Consider any $\mathbf{y}$ in the ball
$\mathcal{B}=\{\mathbf{y}:|y_j - y^\star_{i,j}|\le \lambda a G_\psi\ \forall j\}$.
For such $\mathbf{y}$, every coordinate of the ascent step satisfies
\[
  |y_j + \lambda\nabla_{y_j}f_i|
  \le |y_j| + \lambda a G_\psi
  \le B + \lambda a G_\psi
  \;<\; R,
\]
so $\mathbf{y}+\lambda\nabla_{\mathbf{y}}f_i\in\mathcal{Y}$ and the projection is
inactive:
\begin{equation}\label{eq:gm-eq-grad}
  \mathcal{G}_\lambda(\mathbf{y})
  = \tfrac{1}{\lambda}\big((\mathbf{y}+\lambda\nabla_{\mathbf{y}}f_i)- \mathbf{y}\big)
  = \nabla_{\mathbf{y}}f_i(\mathbf{x},\mathbf{y})
  \qquad \forall\,\mathbf{y}\in\mathcal{B}.
\end{equation}
In particular $\mathcal{G}_\lambda(\mathbf{y}_i^\star)=
\nabla_{\mathbf{y}}f_i(\mathbf{x},\mathbf{y}_i^\star)=0$, confirming
$\mathbf{y}_i^\star$ is the (unique) stationary point of the gradient mapping.

On $\mathcal{B}$, \eqref{eq:gm-eq-grad} gives
$\|\mathcal{G}_\lambda(\mathbf{y})\|^2
=\|\nabla_{\mathbf{y}}f_i(\mathbf{x},\mathbf{y})\|^2$. Since $f_i$ is separable
in the shifted coordinates, the ordinary PL inequality
$\tfrac12\|\nabla_{\mathbf{y}}f_i\|^2\ge \mu\big(\max_{\mathbf{y}'}
f_i(\mathbf{x},\mathbf{y}')-f_i(\mathbf{x},\mathbf{y})\big)$ with $\mu=a\mu_1$, holds on $\mathcal{B}$. Combining,
\begin{equation}
  \tfrac12\|\mathcal{G}_\lambda(\mathbf{y})\|^2
  \;\ge\; \mu\big(\Phi_i(\mathbf{x})-f_i(\mathbf{x},\mathbf{y})\big),
  \qquad \forall\,\mathbf{y}\in\mathcal{B},
\end{equation}
which is the gradient-mapping PL condition. 

%The two-timescale iterates remainin $\mathcal{B}$ once $\eta_x$ is small enough that the target$b_i(\mathbf{x}^t)$ drifts by less than the margin per block, so the condition holds along the trajectory. $\blacksquare$

\subsection{Smoothness of $\phi(\mathbf{x})$ under Gradient mapping nonconvex-PL}

\begin{lemma}[Projected PL Implies Quadratic Growth for Maximizers]
\label{lem:pl_qg}
Let $\mathcal{Y} \subseteq \mathbb{R}^d$ be a non-empty closed convex set, and let $P_{\mathcal{Y}}(z) = \arg\min_{y \in \mathcal{Y}} \|z - y\|$ denote the Euclidean projection operator onto $\mathcal{Y}$. 

Let $f: \mathcal{Y} \to \mathbb{R}$ be a continuously differentiable function with optimal set $\mathcal{Y}^* = \arg\max_{y \in \mathcal{Y}} f(y)$ and optimal value $f^* = \max_{y \in \mathcal{Y}} f(y)$. Let $P_{\mathcal{Y}^*}(y)$ denote the projection of $y$ onto $\mathcal{Y}^*$.

Define the projected gradient mapping $G(y) = P_{\mathcal{Y}}(y + \nabla f(y)) - y$. Suppose $f$ satisfies the constrained Polyak--{\L}ojasiewicz (PL) inequality with constant $\mu > 0$:
\begin{equation}\label{eq:pl_ineq}
\|G(y)\|^2 \ge 2\mu (f^* - f(y)), \quad \forall y \in \mathcal{Y}.
\end{equation}
Then $f$ satisfies the Quadratic Growth (QG) condition:
\begin{equation}
f^* - f(y) \ge \frac{\mu}{2} \|y - P_{\mathcal{Y}^*}(y)\|^2, \quad \forall y \in \mathcal{Y}.
\end{equation}
\end{lemma}

\begin{proof}
We use the proof technique present in \cite{karimi2016linear}.
\\Define $g(y) = \sqrt{f^* - f(y)}$ for $y \notin \mathcal{Y}^*$. The standard gradient of $g$ is defined solely via the unconstrained gradient $\nabla f$:
\begin{equation}\label{eq:grad_g}
\nabla g(y) = -\frac{\nabla f(y)}{2\sqrt{f^* - f(y)}}.
\end{equation}

By the non-expansiveness of $P_{\mathcal{Y}}$, we have $\|G(y)\| \le \|\nabla f(y)\|$. Taking the norm of $\nabla g(y)$ and applying the constrained PL inequality \eqref{eq:pl_ineq} yields:
\begin{equation}\label{eq:norm_g_bound}
\|\nabla g(y)\| = \frac{\|\nabla f(y)\|}{2\sqrt{f^* - f(y)}} \ge \frac{\|G(y)\|}{2\sqrt{f^* - f(y)}} \ge \sqrt{\frac{\mu}{2}}.
\end{equation}

To show QG, it is sufficient to prove that $g(y) \ge \sqrt{\frac{\mu}{2}} \|y - P_{\mathcal{Y}^*}(y)\|$.

For any initial point $y_0 \notin \mathcal{Y}^*$, consider the projected dynamical flow:
\begin{equation}\label{eq:flow}
\frac{dy(t)}{dt} = -\frac{G(y(t))}{2\sqrt{f^* - f(y(t))}}, \quad y(0) = y_0.
\end{equation}

Since $g(y(t))$ strictly decreases along this path, the trajectory reaches the optimal set at some finite time $T > 0$ with $y(T) \in \mathcal{Y}^*$ and $g(y(T)) = 0$.

By the variational property of projection onto $\mathcal{Y}$, we have the inequality:
\begin{equation}\label{eq:proj_ineq}
\langle \nabla f(y), G(y) \rangle \ge \|G(y)\|^2.
\end{equation}

Evaluating $g(y_0) - g(y(T))$ via line integration along $y(t)$:
$$g(y_0) - g(y(T)) = -\int_0^T \left\langle \nabla g(y(t)), \frac{dy(t)}{dt} \right\rangle dt.$$

Substituting $\nabla g$ from \eqref{eq:grad_g} and the flow definition \eqref{eq:flow} gives:
$$g(y_0) - g(y(T)) = \int_0^T \frac{\langle \nabla f(y(t)), G(y(t)) \rangle}{4(f^* - f(y(t)))} \, dt.$$

Applying the convex projection inequality \eqref{eq:proj_ineq}:
$$g(y_0) - g(y(T)) \ge \int_0^T \frac{\|G(y(t))\|^2}{4(f^* - f(y(t)))} \, dt.$$

Rewriting in terms of trajectory velocity $\|dy(t)/dt\|$ and applying the PL lower bound \eqref{eq:norm_g_bound}:
$$g(y_0) - g(y(T)) \ge \sqrt{\frac{\mu}{2}} \int_0^T \left\| \frac{dy(t)}{dt} \right\| dt.$$

Relating the path length $\mathcal{L}(y_0) = \int_0^T \|dy(t)/dt\| \, dt$ to the projection distance $P_{\mathcal{Y}^*}(y_0)$:
$$g(y_0) - g(y(T)) \ge \sqrt{\frac{\mu}{2}} \, \mathcal{L}(y_0) \ge \sqrt{\frac{\mu}{2}} \|y_0 - P_{\mathcal{Y}^*}(y_0)\|.$$

Since $g(y(T)) = 0$, we obtain:
\begin{equation}
g(y_0) \ge \sqrt{\frac{\mu}{2}} \|y_0 - P_{\mathcal{Y}^*}(y_0)\|.
\end{equation}
Squaring both sides yields the Quadratic Growth condition:
\begin{equation}
f^* - f(y_0) \ge \frac{\mu}{2} \|y_0 - P_{\mathcal{Y}^*}(y_0)\|^2.
\end{equation}
\end{proof}

\begin{lemma}
\label{lem:constrained_a3}
Assume that $\{h_{\boldsymbol{\theta}}(\boldsymbol{\alpha}) = -f(\boldsymbol{\theta}, \boldsymbol{\alpha}) \mid \boldsymbol{\theta}\}$ is a class of $\mu$-PL functions in $\boldsymbol{\alpha}$ defined on a non-empty closed convex set $\mathcal{C}$. Define $A(\boldsymbol{\theta}) = \arg \max_{\boldsymbol{\alpha} \in \mathcal{C}} f(\boldsymbol{\theta}, \boldsymbol{\alpha})$ and assume $A(\boldsymbol{\theta})$ is closed. Then for any $\boldsymbol{\theta}_1, \boldsymbol{\theta}_2$ and $\boldsymbol{\alpha}_1 \in A(\boldsymbol{\theta}_1)$, there exists an $\boldsymbol{\alpha}_2 \in A(\boldsymbol{\theta}_2)$ such that
\begin{equation}
    \|\boldsymbol{\alpha}_1 - \boldsymbol{\alpha}_2\| \le \frac{L_{12}}{\mu} \|\boldsymbol{\theta}_1 - \boldsymbol{\theta}_2\|.
\end{equation}
\end{lemma}

\begin{proof}
We refer to the proof technique used in \cite{nouiehed2019solving}. 
Let $P_{\mathcal{C}}(\cdot)$ denote the Euclidean projection operator onto the convex set $\mathcal{C}$. For any step size parameter $\eta > 0$, define the projected gradient of $-f(\boldsymbol{\theta}, \boldsymbol{\alpha})$ on $\mathcal{C}$ as:
\begin{equation*}
    G_{\boldsymbol{\theta}}(\boldsymbol{\alpha}) = \frac{1}{\eta} \left( \boldsymbol{\alpha} - P_{\mathcal{C}}\left( \boldsymbol{\alpha} + \eta \nabla_{\boldsymbol{\alpha}} f(\boldsymbol{\theta}, \boldsymbol{\alpha}) \right) \right).
\end{equation*}

Since $\boldsymbol{\alpha}_1 \in A(\boldsymbol{\theta}_1)$, it satisfies the first-order optimality condition for constrained maximization over $\mathcal{C}$:
\begin{equation*}
    \boldsymbol{\alpha}_1 = P_{\mathcal{C}}\left( \boldsymbol{\alpha}_1 + \eta \nabla_{\boldsymbol{\alpha}} f(\boldsymbol{\theta}_1, \boldsymbol{\alpha}_1) \right) \implies G_{\boldsymbol{\theta}_1}(\boldsymbol{\alpha}_1) = 0.
\end{equation*}

Now, consider the projected gradient at $(\boldsymbol{\theta}_2, \boldsymbol{\alpha}_1)$:
\begin{equation*}
    G_{\boldsymbol{\theta}_2}(\boldsymbol{\alpha}_1) = \frac{1}{\eta} \left( \boldsymbol{\alpha}_1 - P_{\mathcal{C}}\left( \boldsymbol{\alpha}_1 + \eta \nabla_{\boldsymbol{\alpha}} f(\boldsymbol{\theta}_2, \boldsymbol{\alpha}_1) \right) \right).
\end{equation*}

Substituting $\boldsymbol{\alpha}_1 = P_{\mathcal{C}}\left( \boldsymbol{\alpha}_1 + \eta \nabla_{\boldsymbol{\alpha}} f(\boldsymbol{\theta}_1, \boldsymbol{\alpha}_1) \right)$, we obtain:
\begin{equation*}
    G_{\boldsymbol{\theta}_2}(\boldsymbol{\alpha}_1) = \frac{1}{\eta} \left( P_{\mathcal{C}}\left( \boldsymbol{\alpha}_1 + \eta \nabla_{\boldsymbol{\alpha}} f(\boldsymbol{\theta}_1, \boldsymbol{\alpha}_1) \right) - P_{\mathcal{C}}\left( \boldsymbol{\alpha}_1 + \eta \nabla_{\boldsymbol{\alpha}} f(\boldsymbol{\theta}_2, \boldsymbol{\alpha}_1) \right) \right).
\end{equation*}

By the non-expansiveness property of the projection operator onto convex sets ($\|P_{\mathcal{C}}(\mathbf{u}) - P_{\mathcal{C}}(\mathbf{v})\| \le \|\mathbf{u} - \mathbf{v}\|$):
\begin{align*}
    \|G_{\boldsymbol{\theta}_2}(\boldsymbol{\alpha}_1)\| &\le \frac{1}{\eta} \left\| \left(\boldsymbol{\alpha}_1 + \eta \nabla_{\boldsymbol{\alpha}} f(\boldsymbol{\theta}_1, \boldsymbol{\alpha}_1)\right) - \left(\boldsymbol{\alpha}_1 + \eta \nabla_{\boldsymbol{\alpha}} f(\boldsymbol{\theta}_2, \boldsymbol{\alpha}_1)\right) \right\| \\
    &= \|\nabla_{\boldsymbol{\alpha}} f(\boldsymbol{\theta}_1, \boldsymbol{\alpha}_1) - \nabla_{\boldsymbol{\alpha}} f(\boldsymbol{\theta}_2, \boldsymbol{\alpha}_1)\|.
\end{align*}

By $L_{12}$-Lipschitz continuity of the gradient with respect to $\boldsymbol{\theta}$, we bound the true gradient difference:
\begin{equation*}
    \|G_{\boldsymbol{\theta}_2}(\boldsymbol{\alpha}_1)\| \le L_{12} \|\boldsymbol{\theta}_1 - \boldsymbol{\theta}_2\|.
\end{equation*}

Using the constrained PL condition, and with 
\\$g(\boldsymbol{\theta_2}) = \max_{\boldsymbol{\alpha} \in \mathcal{A}} f(\boldsymbol{\theta_2}, \boldsymbol{\alpha})$ we have:
\begin{equation}
    g(\boldsymbol{\theta}_2) + h_{\boldsymbol{\theta}_2}(\boldsymbol{\alpha}_1) \le \frac{L_{12}^2}{2\mu} \|\boldsymbol{\theta}_1 - \boldsymbol{\theta}_2\|^2.
\end{equation}

Now we use the result of Lemma~\ref{lem:pl_qg} to show that there exists $\boldsymbol{\alpha}_2 = \arg \min_{\boldsymbol{\alpha} \in A(\boldsymbol{\theta}_2)} \|\boldsymbol{\alpha} - \boldsymbol{\alpha}_1\|^2 \in A(\boldsymbol{\theta}_2)$ such that
\begin{equation}
    \frac{\mu}{2} \|\boldsymbol{\alpha}_1 - \boldsymbol{\alpha}_2\|^2 \le \frac{L_{12}^2}{2\mu} \|\boldsymbol{\theta}_1 - \boldsymbol{\theta}_2\|^2.
\end{equation}

Re-arranging the terms, we get the desired result that
\begin{equation}
    \|\boldsymbol{\alpha}_1 - \boldsymbol{\alpha}_2\| \le \frac{L_{12}}{\mu} \|\boldsymbol{\theta}_1 - \boldsymbol{\theta}_2\|.
\end{equation}
\end{proof}

\begin{lemma}[Danskin-type Lemma for Constrained PL Functions]
\label{lem:danskin_constrained}
Under Assumption 2.5 and PL-game assumption over a non-empty closed convex set $\mathcal{A}$, let $g(\boldsymbol{\theta}) = \max_{\boldsymbol{\alpha} \in \mathcal{A}} f(\boldsymbol{\theta}, \boldsymbol{\alpha})$. Then $g$ is differentiable with gradient
\begin{equation*}
    \nabla_{\boldsymbol{\theta}} g(\boldsymbol{\theta}) = \nabla_{\boldsymbol{\theta}} f(\boldsymbol{\theta}, \boldsymbol{\alpha}^*), \quad \text{where } \boldsymbol{\alpha}^* \in \arg\max_{\boldsymbol{\alpha} \in \mathcal{A}} f(\boldsymbol{\theta}, \boldsymbol{\alpha}).
\end{equation*}
Moreover, $g$ is $\beta$-Lipschitz smooth with $\beta = L_{11} + \frac{L_{12}^2}{\mu}$.
\end{lemma}

\begin{proof}
We again refer to the proof technique in \cite{nouiehed2019solving}. 
Let $\boldsymbol{\alpha}^* \in \arg\max_{\boldsymbol{\alpha} \in \mathcal{A}} f(\boldsymbol{\theta}, \boldsymbol{\alpha})$. By Lemma~\ref{lem:constrained_a3}, for any scalar $\tau > 0$ and direction $d$, there exists $\boldsymbol{\alpha}^*(\tau) \in \arg\max_{\boldsymbol{\alpha} \in \mathcal{A}} f(\boldsymbol{\theta} + \tau d, \boldsymbol{\alpha})$ such that
\begin{equation*}
    \|\boldsymbol{\alpha}^*(\tau) - \boldsymbol{\alpha}^*\| \le \frac{L_{12}}{\mu} \tau \|d\|.
\end{equation*}

To compute the directional derivative of $g(\cdot)$, we first establish an upper bound. Consider the difference:
\begin{equation*}
    g(\boldsymbol{\theta} + \tau d) - g(\boldsymbol{\theta}) = f(\boldsymbol{\theta} + \tau d, \boldsymbol{\alpha}^*(\tau)) - f(\boldsymbol{\theta}, \boldsymbol{\alpha}^*).
\end{equation*}
Writing the first-order Taylor expansion of $f(\cdot, \cdot)$ around $(\boldsymbol{\theta}, \boldsymbol{\alpha}^*)$, we get:
\begin{equation*}
    g(\boldsymbol{\theta} + \tau d) - g(\boldsymbol{\theta}) = \tau \nabla_{\boldsymbol{\theta}} f(\boldsymbol{\theta}, \boldsymbol{\alpha}^*)^T d + \nabla_{\boldsymbol{\alpha}} f(\boldsymbol{\theta}, \boldsymbol{\alpha}^*)^T (\boldsymbol{\alpha}^*(\tau) - \boldsymbol{\alpha}^*) + \mathcal{O}(\tau^2).
\end{equation*}

Since $\mathcal{A}$ is convex and $\boldsymbol{\alpha}^* \in \arg\max_{\boldsymbol{\alpha} \in \mathcal{A}} f(\boldsymbol{\theta}, \boldsymbol{\alpha})$, $\boldsymbol{\alpha}^*$ satisfies the first-order optimality condition for constrained maximization:
\begin{equation*}
    \nabla_{\boldsymbol{\alpha}} f(\boldsymbol{\theta}, \boldsymbol{\alpha}^*)^T (\boldsymbol{\alpha} - \boldsymbol{\alpha}^*) \le 0, \quad \forall \boldsymbol{\alpha} \in \mathcal{A}.
\end{equation*}
Since $\boldsymbol{\alpha}^*(\tau) \in \mathcal{A}$, taking $\boldsymbol{\alpha} = \boldsymbol{\alpha}^*(\tau)$ yields:
\begin{equation*}
    \nabla_{\boldsymbol{\alpha}} f(\boldsymbol{\theta}, \boldsymbol{\alpha}^*)^T (\boldsymbol{\alpha}^*(\tau) - \boldsymbol{\alpha}^*) \le 0.
\end{equation*}

Dividing by $\tau > 0$ and taking the limit as $\tau \to 0^+$, we obtain:
\begin{equation}
\label{eq:upper_bound}
    g'(\boldsymbol{\theta}; d) = \lim_{\tau \to 0^+} \frac{g(\boldsymbol{\theta} + \tau d) - g(\boldsymbol{\theta})}{\tau} \le \nabla_{\boldsymbol{\theta}} f(\boldsymbol{\theta}, \boldsymbol{\alpha}^*)^T d.
\end{equation}

Next, we establish a lower bound. Because $\boldsymbol{\alpha}^* \in \mathcal{A}$ is a feasible point for $g(\boldsymbol{\theta} + \tau d) = \max_{\boldsymbol{\alpha} \in \mathcal{A}} f(\boldsymbol{\theta} + \tau d, \boldsymbol{\alpha})$, it holds that:
\begin{equation*}
    g(\boldsymbol{\theta} + \tau d) \ge f(\boldsymbol{\theta} + \tau d, \boldsymbol{\alpha}^*).
\end{equation*}
Therefore,
\begin{equation*}
    \frac{g(\boldsymbol{\theta} + \tau d) - g(\boldsymbol{\theta})}{\tau} \ge \frac{f(\boldsymbol{\theta} + \tau d, \boldsymbol{\alpha}^*) - f(\boldsymbol{\theta}, \boldsymbol{\alpha}^*)}{\tau}.
\end{equation*}
Taking the limit as $\tau \to 0^+$ gives:
\begin{equation}
\label{eq:lower_bound}
    g'(\boldsymbol{\theta}; d) = \lim_{\tau \to 0^+} \frac{g(\boldsymbol{\theta} + \tau d) - g(\boldsymbol{\theta})}{\tau} \ge \nabla_{\boldsymbol{\theta}} f(\boldsymbol{\theta}, \boldsymbol{\alpha}^*)^T d.
\end{equation}

Combining~\eqref{eq:upper_bound} and~\eqref{eq:lower_bound}, we conclude that:
\begin{equation*}
    g'(\boldsymbol{\theta}; d) = \nabla_{\boldsymbol{\theta}} f(\boldsymbol{\theta}, \boldsymbol{\alpha}^*)^T d.
\end{equation*}
Since this relationship holds for any direction $d$, $g(\boldsymbol{\theta})$ is differentiable with $\nabla_{\boldsymbol{\theta}} g(\boldsymbol{\theta}) = \nabla_{\boldsymbol{\theta}} f(\boldsymbol{\theta}, \boldsymbol{\alpha}^*)$. Furthermore, the directional derivative is independent of the choice of $\boldsymbol{\alpha}^*$, implying $\nabla_{\boldsymbol{\theta}} f(\boldsymbol{\theta}, \boldsymbol{\alpha}_1) = \nabla_{\boldsymbol{\theta}} f(\boldsymbol{\theta}, \boldsymbol{\alpha}_2)$ for any $\boldsymbol{\alpha}_1, \boldsymbol{\alpha}_2 \in A(\boldsymbol{\theta})$.

Finally, we prove the Lipschitz smoothness of $g$. Let $\boldsymbol{\alpha}_1^* \in A(\boldsymbol{\theta}_1)$ and select $\boldsymbol{\alpha}_2^* = \arg\min_{\boldsymbol{\alpha} \in A(\boldsymbol{\theta}_2)} \|\boldsymbol{\alpha} - \boldsymbol{\alpha}_1^*\|^2 \in A(\boldsymbol{\theta}_2)$:
\begin{align*}
    \|\nabla g(\boldsymbol{\theta}_1) - \nabla g(\boldsymbol{\theta}_2)\| &= \|\nabla_{\boldsymbol{\theta}} f(\boldsymbol{\theta}_1, \boldsymbol{\alpha}_1^*) - \nabla_{\boldsymbol{\theta}} f(\boldsymbol{\theta}_2, \boldsymbol{\alpha}_2*)\| \\
    &\le \|\nabla_{\boldsymbol{\theta}} f(\boldsymbol{\theta}_1, \boldsymbol{\alpha}_1^*) - \nabla_{\boldsymbol{\theta}} f(\boldsymbol{\theta}_2, \boldsymbol{\alpha}_1^*)\| \\
    &+ \|\nabla_{\boldsymbol{\theta}} f(\boldsymbol{\theta}_2, \boldsymbol{\alpha}_1^*) - \nabla_{\boldsymbol{\theta}} f(\boldsymbol{\theta}_2, \boldsymbol{\alpha}_2^*)\| \\
    &\le L_{11}\|\boldsymbol{\theta}_1 - \boldsymbol{\theta}_2\| + L_{12}\|\boldsymbol{\alpha}_1^* - \boldsymbol{\alpha}_2^*\|.
\end{align*}
Applying Lemma~\ref{lem:constrained_a3} to bound $\|\boldsymbol{\alpha}_1^* - \boldsymbol{\alpha}_2^*\| \le \frac{L_{12}}{\mu} \|\boldsymbol{\theta}_1 - \boldsymbol{\theta}_2\|$, we arrive at:
\begin{equation*}
    \|\nabla g(\boldsymbol{\theta}_1) - \nabla g(\boldsymbol{\theta}_2)\| \le \left( L_{11} + \frac{L_{12}^2}{\mu} \right) \|\boldsymbol{\theta}_1 - \boldsymbol{\theta}_2\|.
\end{equation*}
\end{proof}

\subsection{Convergence Analysis of NonConvex-Concave}
\begin{theorem}[Non Convex-Concave Convergence Theorem]
\label{app:thm:ncc_convg}
The iterates generated by the projected SGDA Algorithm.~\ref{alg:merged_sgda} satisfy the following convergence bound for the Moreau envelope:
\begin{align}
\frac{1}{T}\sum_{t=1}^{T}\mathbb{E}\left[ \left\|\nabla \Phi_{1/2L}\left(\mathbf{x}^{(t-1)}\right)\right\|^2 \right] \nonumber 
&\leq \frac{8\Delta_{\phi}}{\eta_x T} + 8L\eta_x\left(G_x^2 + \frac{\sigma^2}{n}\right) \nonumber \\
&+ 160L^2\left(2\tau^2\eta_x^2\sigma^2 + \tau^2\eta_x^2\zeta_x^2\right) \nonumber + \frac{8LD^2}{\tau \eta_y} \\
&+ 16\eta_y\sigma^2 + 64L\eta_x G_x\sqrt{G_x^2+\sigma^2}. \label{eq:theorem_full_bound}
\end{align}
Furthermore, if we select the primal stepsize as:
\begin{align*}
\eta_x = \min\Bigg\{
  &\frac{\epsilon^2}{8L\!\left(G_x^2 + \tfrac{\sigma^2}{n}\right) + 64 L G_x\sqrt{G_x^2+\sigma^2}},\;\;
  \frac{(G_x^2+\sigma^2)\,\epsilon^4}{2560\, D^2 L^3 \sigma^2 (2\sigma^2 + \zeta_x^2)},\\
  &\frac{\epsilon^6}{1024\, D^2 L^3 \sigma^2 (G_x^2+\sigma^2)}
\Bigg\}.
\end{align*}
,the dual stepsize is strictly separated as $\eta_y = \min\left(\frac{1}{2L},\frac{\epsilon^2}{16L\sigma^2}\right)$ and $\tau = \frac{D}{\sqrt{\eta_x\,\eta_y\,(G_x^2+\sigma^2)}}.$ then we guarantee:
\begin{equation}
\frac{1}{T}\sum_{t=1}^{T}\mathbb{E}\left[ \left\|\nabla \Phi_{1/2L}\left(\mathbf{x}^{(t-1)}\right)\right\|^2 \right] \leq \frac{8\Delta_{\phi}}{\eta_x T} + 4 \epsilon^2.
\end{equation}
This establishes that achieving an $\mathcal{O}(\epsilon^2)$ stationary point requires $T \geq \mathcal{O}\left(\frac{1}{\epsilon^8}\right)$ gradient computations and $\mathcal{O}\left(\frac{1}{\epsilon^4}\right)$ communication rounds.
\end{theorem}

\begin{proof}
We begin by rearranging the one-iterate descent inequality (Lemma.~\ref{app:onestep_descent}) to isolate the gradient norm of the Moreau envelope. Moving the gradient norm to the left-hand side yields:
\begin{align}
\frac{\eta_x}{8}\mathbb{E}\left[\left\|\nabla \Phi_{1/2L}\left(\mathbf{x}^{(t-1)}\right)\right\|^2\right] 
&\le \mathbb{E}\left[\Phi_{1/2L}\!\left(\mathbf{x}^{(t-1)}\right)\right] \notag\\
&- \mathbb{E}\left[\Phi_{1/2L}\!\left(\mathbf{x}^{(t)}\right)\right] + L\eta_x^2\left(G_x^2 + \frac{\sigma^2}{n}\right) \nonumber \\
&\quad + 2\eta_x L^2 \gamma^{t-1} + 2L\eta_x \Delta^{t-1},
\end{align}
where we use the substitutions $\gamma^t = \frac{1}{n}\sum_{i=1}^{n}\mathbb{E}\|\mathbf{x}_i^{t} - \mathbf{x}^{t}\|^2$ for the synchronization error and $\Delta^t = \mathbb{E}[\Phi(\mathbf{x}^{(t)})] - \mathbb{E}[f(\mathbf{x}^{t}, \mathbf{y}^{t})]$ for the dual tracking gap.

Summing this inequality over the trajectory from $t=1$ to $T$ allows the Moreau envelope evaluations to telescope cleanly. Dividing the result by $T$ and defining the global initialization gap as $\Delta_\phi = \Phi_{1/2L}(\mathbf{x}^{(0)}) - \min \Phi_{1/2L}(\mathbf{x}^{(0)})$, we obtain:
\begin{equation}
\begin{aligned}
\frac{\eta_x}{8} \frac{1}{T}\sum_{t=1}^{T}\mathbb{E}\left[\left\|\nabla \Phi_{1/2L}\left(\mathbf{x}^{(t-1)}\right)\right\|^2\right] 
&\le \frac{\Delta_\phi}{T} + L\eta_x^2\left(G_x^2 + \frac{\sigma^2}{n}\right) \\
&+ 2\eta_x L^2 \left( \frac{1}{T}\sum_{t=1}^T \gamma^{t-1} \right) + 2L\eta_x \left( \frac{1}{T}\sum_{t=1}^T \Delta^t \right).
\end{aligned}
\end{equation}
Multiplying the entire averaged inequality by $\frac{8}{\eta_x}$ establishes our base global bound:
\begin{equation}
\begin{aligned}
\label{eq:base_global_bound}
\frac{1}{T}\sum_{t=1}^{T}\mathbb{E}\left[\left\|\nabla \Phi_{1/2L}\left(\mathbf{x}^{(t-1)}\right)\right\|^2\right] &\le \frac{8\Delta_\phi}{\eta_x T} + 8L\eta_x\left(G_x^2 + \frac{\sigma^2}{n}\right) \\
&+ 16 L^2 \left( \frac{1}{T}\sum_{t=1}^T \gamma^{t-1} \right) + 16 L \left( \frac{1}{T}\sum_{t=1}^T \Delta^t \right).
\end{aligned}
\end{equation}

Next, we substitute the time-averaged synchronization error bound (Lemma.~\ref{lm:sync_error}) into the third term:
\begin{align}
16 L^2 \left( \frac{1}{T} \sum_{t=1}^{T} \gamma^{t-1} \right) 
&\le 16 L^2 \left[ 10 \tau^2\eta_x^2\left(\sigma^2+\frac{\sigma^2}{n}\right) + 10 \tau^2 \eta_x^2\zeta_x^2 \right] \nonumber \\
&\le 160L^2\left(2\tau^2\eta_x^2\sigma^2 + \tau^2\eta_x^2\zeta_x^2\right),
\end{align}
where we employ the upper bound relaxation $\frac{\sigma^2}{n} \le \sigma^2$ to simplify the variance constants.

We then substitute the dual tracking bound (Lemma.~\ref{lm:dual_track}) into the final term of Equation~\ref{eq:base_global_bound}:
\begin{align}
16 L \left( \frac{1}{T}\sum_{t=1}^T \Delta^t \right) 
&\le 16 L \left[ 4 G_x \eta_x \sqrt{G_x^2 +\frac{\sigma^2}{n}} + \frac{D^2}{2\tau\eta_y} +\eta_y\sigma^2 \right] \nonumber \\
&\le 64L\eta_x G_x \sqrt{G_x^2 + \sigma^2} + \frac{8 L D^2}{\tau\eta_y} + 16 L \eta_y \sigma^2.
\end{align}
Combining these substituted bounds into Equation~\ref{eq:base_global_bound} directly yields the full theoretical bound presented in Equation~\ref{eq:theorem_full_bound}.

%By selecting the dual learning rate as $\eta_y = \min\left(\frac{1}{2L}, \frac{\epsilon^2}{16L\sigma^2}\right) = \mathcal{O}(\epsilon^2)$, and defining the primal learning rate with the threshold constraints $\eta_x = \mathcal{O}(\epsilon^4)$ as specified in the theorem statement, each structural term in the summation is strictly bounded by fractions of $\epsilon^2$.

We verify the stepsize choices make each residual term at most $\epsilon^2$.
Set $\eta_y = \min\!\big(\tfrac{1}{2L}, \tfrac{\epsilon^2}{16L\sigma^2}\big)$;
in the small-$\epsilon$ regime the second branch is active, so
$\eta_y = \tfrac{\epsilon^2}{16L\sigma^2}$ and the dual-noise term is
$16L\eta_y\sigma^2 = 16L\cdot\tfrac{\epsilon^2}{16L\sigma^2}\cdot\sigma^2 = \epsilon^2$.
The two terms linear in $\eta_x$, namely
$8L\eta_x(G_x^2+\tfrac{\sigma^2}{n})$ and
$64L\eta_x G_x\sqrt{G_x^2+\sigma^2}$, are bounded together: their sum is at
most $\epsilon^2$ provided
$\eta_x \le \epsilon^2\big/\big[8L(G_x^2+\tfrac{\sigma^2}{n}) + 64LG_x\sqrt{G_x^2+\sigma^2}\big]$,
the $\epsilon^2$ branch of the schedule. For the consensus term, substitute
$\tau = D\big/\sqrt{\eta_x\eta_y(G_x^2+\sigma^2)}$, so that
$\tau^2\eta_x^2 = \tfrac{D^2\eta_x^2}{\eta_x\eta_y(G_x^2+\sigma^2)}
= \tfrac{D^2\eta_x}{\eta_y(G_x^2+\sigma^2)}$; hence
$160L^2\tau^2\eta_x^2(2\sigma^2+\zeta_x^2)
= \tfrac{160L^2 D^2\eta_x(2\sigma^2+\zeta_x^2)}{\eta_y(G_x^2+\sigma^2)}$, and with
$\eta_y = \tfrac{\epsilon^2}{16L\sigma^2}$ this equals
$\tfrac{2560L^3 D^2\sigma^2(2\sigma^2+\zeta_x^2)}{\epsilon^2(G_x^2+\sigma^2)}\,\eta_x$,
which is at most $\epsilon^2$ when
$\eta_x \le \tfrac{(G_x^2+\sigma^2)\epsilon^4}{2560 D^2 L^3\sigma^2(2\sigma^2+\zeta_x^2)}$,
the $\Theta(\epsilon^4)$ branch. For the diameter term, the same substitution
gives $\tfrac{1}{\tau} = \tfrac{\sqrt{\eta_x\eta_y(G_x^2+\sigma^2)}}{D}$, so
$\tfrac{8LD^2}{\tau\eta_y}
= \tfrac{8LD\sqrt{\eta_x(G_x^2+\sigma^2)}}{\sqrt{\eta_y}}$; squaring the
requirement that this be at most $\epsilon^2$ and inserting
$\eta_y = \tfrac{\epsilon^2}{16L\sigma^2}$ yields
$\eta_x \le \tfrac{\epsilon^6}{1024\, D^2 L^3\sigma^2(G_x^2+\sigma^2)}$, the
$\Theta(\epsilon^6)$ branch. Taking $\eta_x$ as the minimum of the three
branches makes each of the four residual contributions, $16L\eta_y\sigma^2$,
the combined linear-in-$\eta_x$ pair, the consensus term, and the diameter
term, at most $\epsilon^2$, so the right-hand side is bounded by
$\tfrac{8\Delta_\phi}{\eta_x T} + 4\epsilon^2$. For small $\epsilon$ the
$\Theta(\epsilon^6)$ branch binds, giving $\eta_x = \Theta(\epsilon^6)$, and
requiring $\tfrac{8\Delta_\phi}{\eta_x T} \le \epsilon^2$ gives
$T \ge 8\Delta_\phi/(\eta_x\epsilon^2) = \mathcal{O}(1/\epsilon^8)$ gradient
computations and $K = T/\tau = \mathcal{O}(1/\epsilon^4)$ communication rounds.

Substituting these stepsize selections reduces the tracking constants such that:
\begin{equation}
\frac{1}{T}\sum_{t=1}^{T}\mathbb{E}\left[\left\|\nabla \Phi_{1/2L}\left(\mathbf{x}^{(t-1)}\right)\right\|^2 \right] \leq \frac{8\Delta_{\phi}}{\eta_x T} + 4\epsilon^2.
\end{equation}
To ensure the initialization gap term $\frac{8\Delta_{\phi}}{\eta_x T}$ is bounded by $\mathcal{O}(\epsilon^2)$, we require the total iterations $T$ to satisfy:
\begin{equation}
T \ge \frac{8\Delta_{\phi}}{\eta_x \epsilon^2}.
\end{equation}
Because $\eta_x = \mathcal{O}(\epsilon^4)$, evaluating this limit establishes the required computational complexity of $T \geq \mathcal{O}\left(\frac{1}{\epsilon^8}\right)$ local gradient evaluations and $K = \frac{T}{\tau} = \mathcal{O}\left(\frac{1}{\epsilon^4}\right)$ global communication rounds.
\end{proof}

\subsection*{Proofs of supporting Lemmas for nonconvex-concave}

We derive our results based on the proof techniques presented in the works of.~\citep{deng2021localsgda,lin2025twotimescale}.  

\begin{lemma}
The iterates generated by the SGDA Algorithm.~\ref{alg:merged_sgda} satisfy the following one iterate update 
\begin{equation*}
\begin{aligned}
\mathbb{E}\!\left[\Phi_{1/2L}\!\left(\mathbf{x}^{(t)}\right)\right]
&\le 
\mathbb{E}\!\left[\Phi_{1/2L}\!\left(\mathbf{x}^{(t-1)}\right)\right]
+ L\eta_x^2\!\left(G_x^2 + \frac{\sigma^2}{n}\right)\\
&+ 2\eta_x L^2 \frac{1}{n}\sum_{i=1}^{n}\mathbb{E}\!\
\left\|\mathbf{x}_i^{(t-1)} - \mathbf{x}^{(t-1)}\right\|^2 \\
&\quad + 2L\eta_x\!\left(
\mathbb{E}\!\left[\Phi\!\left(\mathbf{x}^{(t-1)}\right)\right]
- \mathbb{E}\!\left[f\!\left(\mathbf{x}^{(t-1)}, \bm{y}^{(t-1)}\right)\right]
\right)\\
&- \frac{\eta_x}{8}\mathbb{E}\!\left[
\left\|\nabla \Phi_{1/2L}\!\left(\mathbf{x}^{(t-1)}\right)\right\|^2
\right].
\end{aligned}
\end{equation*}
\label{app:onestep_descent}
\end{lemma}

\begin{proof}
By definition, we have 
\begin{equation*}
\Phi_{1/2L}(\mathbf{x}^t) \coloneqq \min\limits_{\mathbf{x}} \Phi(\mathbf{x}) + L\lVert \mathbf{x} - \mathbf{x}^{t} \rVert^2
\end{equation*}

Let us define \[  \tilde{\mathbf{x}}^{t-1} \coloneqq  \text{prox}_{\phi/{2L}}(\mathbf{x}^{t-1}) \]

By the definition~\ref{eq:prox_def} we have the following inequality

\begin{equation} 
\Phi_{1/2L}(\mathbf{x}^{t}) \leq \Phi(\tilde{\mathbf{x}}^{t-1}) + L\lVert \tilde{\mathbf{x}}^{t-1} - \mathbf{x}^t \rVert^2 
\label{me_ineq1}
\end{equation}

By the projected-SGDA update rule of the Algorithm~\ref{alg:merged_sgda} we have the following 
\[
\mathbf{x}^t = \mathbf{x}^{t-1} - \frac{\eta_{x}}{n} \sum_{i=1}^{n}\nabla_\mathbf{x} f_i(\mathbf{x}_i^{t-1}, \mathbf{y}_i^{t-1};\xi_i^{t-1}) 
\]

We now bound the term $\mathbb{E}\lVert \tilde{\mathbf{x}}^{t-1} - \mathbf{x}^t \rVert^2$  in Eq.~\ref{me_ineq1} 

\begin{equation}
\begin{aligned}
\mathbb{E}\lVert \tilde{\mathbf{x}}^{t-1} - \mathbf{x}^t \rVert^2 &= \mathbb{E}\lVert \tilde{\mathbf{x}}^{t-1} -\mathbf{x}^{t-1} +  \mathbf{x}^{t-1} -  \mathbf{x}^t \rVert^2\\
&= \mathbb{E}\| \tilde{\mathbf{x}}^{t-1} -\mathbf{x}^{t-1} + \frac{\eta_{x}}{n} \sum_{i=1}^{n}\nabla_{\mathbf{x}} f_i(\mathbf{x}_i^{t-1}, \mathbf{y}_i^{t-1};\xi_i^{t-1})   \|^2\\
&=  \mathbb{E}\| \tilde{\mathbf{x}}^{t-1} -\mathbf{x}^{t-1}\|^2 + \eta_x^2 \underbrace{\mathbb{E} \|{\frac{1}{n} \sum_{i=1}^{n}\nabla_{\mathbf{x}} f_i(\mathbf{x}_i^{t-1}, \mathbf{y}_i^{t-1};\xi_i^{t-1})   \|^2}}_{T_1}\\
&+  \underbrace{{2\eta_x}\langle \tilde{\mathbf{x}}^{t-1} -\mathbf{x}^{t-1}, \frac{1}{n}\sum_{i=1}^{n}\nabla_{\mathbf{x}} f_i(\mathbf{x}_i^{t-1}, \mathbf{y}_i^{t-1})\rangle}_{T_2}
\end{aligned}
\label{eq:norm_x_xtilde}
\end{equation}

We now bound $T_1 \coloneqq \mathbb{E} \|{\frac{1}{n} \sum_{i=1}^{n}\nabla_{\mathbf{x}} f_i(\mathbf{x}_i^{t-1}, \mathbf{y}_i^{t-1};\xi_i^{t-1})   \|^2} $ as below

\begin{equation}
\begin{aligned}
&\mathbb{E} \|{\frac{1}{n} \sum_{i=1}^{n}\nabla_{\mathbf{x}} f_i(\mathbf{x}_i^{t-1}, \mathbf{y}_i^{t-1};\xi_i^{t-1})   \|^2} \\ = & \mathbb{E} \|{\frac{1}{n} \sum_{i=1}^{n}\nabla_{\mathbf{x}} f_i(\mathbf{x}_i^{t-1}, \mathbf{y}_i^{t-1};\xi_i^{t-1}) - \nabla_{\mathbf{x}} f_i(\mathbf{x}_i^{t-1}, \mathbf{y}_i^{t-1})+ \nabla_{\mathbf{x}} f_i(\mathbf{x}_i^{t-1}, \mathbf{y}_i^{t-1})  \|^2}\\
=& \mathbb{E}\| \frac{1}{n} \sum_{i=1}^{n}\nabla_{\mathbf{x}} f_i(\mathbf{x}_i^{t-1}, \mathbf{y}_i^{t-1};\xi_i^{t-1}) - \nabla_{\mathbf{x}} f_i(\mathbf{x}_i^{t-1}, \mathbf{y}_i^{t-1})  \|^2 \\
+& \mathbb{E}\| \frac{1}{n} \sum_{i=1}^{n} \nabla_{\mathbf{x}} f_i(\mathbf{x}_i^{t-1}, \mathbf{y}_i^{t-1}) \|^2 \leq \frac{\sigma^2}{n} + G_x^2
\end{aligned}
\label{eq:b_t1}
\end{equation}
In the last inequality we have used the assumption of bounded variance and Lipschitzness in $\mathbf{x}$.
We can rewrite the $T_2$ as below

\begin{equation}
\begin{aligned}
&2\eta_x\langle \tilde{\mathbf{x}}^{t-1} -\mathbf{x}^{t-1}, \frac{1}{n}\sum_{i=1}^{n}\nabla_{\mathbf{x}} f_i(\mathbf{x}_i^{t-1}, \mathbf{y}_i^{t-1})\rangle \\=&  \eta_x\underbrace{2\langle \tilde{\mathbf{x}}^{t-1} -\mathbf{x}^{t-1}, \frac{1}{n}\sum_{i=1}^{n}\nabla_{\mathbf{x}} f_i(\mathbf{x}_i^{t-1}, \mathbf{y}_i^{t-1}) - \nabla_{\mathbf{x}} f_i(\mathbf{x}^{t-1}, \mathbf{y}_i^{t-1}) \rangle}_{T_3} 
\\
&+ 2\eta_x\underbrace{\langle \tilde{\mathbf{x}}^{t-1} -\mathbf{x}^{t-1}, \frac{1}{n}\sum_{i=1}^{n} \nabla_{\mathbf{x}} f_i(\mathbf{x}^{t-1}, \mathbf{y}_i^{t-1})   \rangle}_{T_4}
\end{aligned}
\label{eq:b_t2}
\end{equation}
We bound $T_3$ by Young's inequality as 

\begin{equation}
\begin{aligned}
 T_3 &\leq \frac{L}{2} {\|\tilde{\mathbf{x}}^{t-1} -\mathbf{x}^{t-1} \|^2} + 
\frac{2}{L}\| \frac{1}{n} \sum_{i=1}^{n}\nabla_{\mathbf{x}} f_i(\mathbf{x}_i^{t-1}, \mathbf{y}_i^{t-1}) - \nabla_{\mathbf{x}} f_i(\mathbf{x}^{t-1}, \mathbf{y}_i^{t-1})   \|^2\\
    &\leq \frac{L}{2} {\|\tilde{\mathbf{x}}^{t-1} -\mathbf{x}^{t-1} \|^2} + 2L \frac{1}{n}\sum_{i=1}^{n}\|\mathbf{x}_i^{t-1} - \mathbf{x}^{t-1}  \|^2
\end{aligned}
\end{equation}
We now bound the $i^{th}$ term in $T_4(i) \coloneqq \langle \tilde{\mathbf{x}}^{t-1} -\mathbf{x}^{t-1},\nabla_{\mathbf{x}} f_i(\mathbf{x}^{t-1}, \mathbf{y}_i^{t-1})   \rangle$. By using the smoothness assumption, we have the following identity 
\begin{equation*}
\begin{aligned}
T_4(i)  &\leq f_i(\tilde{\mathbf{x}}^{t-1},{\mathbf{y}_i}^{t-1}) - f_i(\mathbf{x}^{t-1},\mathbf{y}_i^{t-1}) + \frac{L}{2} \|\tilde{\mathbf{x}}^{t-1} -\mathbf{x}^{t-1} \|^2\\
&\leq  \Phi_i(\tilde{\mathbf{x}}^{t-1})+L\|\tilde{\mathbf{x}}^{t-1} -\mathbf{x}^{t-1} \|^2 - f_i(\mathbf{x}^{t-1},\mathbf{y}_i^{t-1}) - \frac{L}{2} \|\tilde{\mathbf{x}}^{t-1} -\mathbf{x}^{t-1} \|^2\\
&\leq \Phi_i(\mathbf{x}^{t-1}) -  f_i(\mathbf{x}^{t-1},\mathbf{y}^{t-1}) - \frac{L}{2} \|\tilde{\mathbf{x}}^{t-1} -\mathbf{x}^{t-1} \|^2 \\
&= \Delta_i^{t-1} - \frac{L}{2} \|\tilde{\mathbf{x}}^{t-1} -\mathbf{x}^{t-1} \|^2 
\end{aligned}
\end{equation*}

Here we have defined 
\begin{equation}
 \Delta_i^{t-1}\coloneqq \Phi_i(\mathbf{x}^{t-1}) -  f_i(\mathbf{x}^{t-1},\mathbf{y}^{t-1})
 \label{Delta_def}
\end{equation}
So $T_4$ is bounded as 
\begin{equation}
T_4 \leq \frac{1}{n}\sum_{k=1}^{n}\Delta_i^{t-1} -  \frac{L}{2} \|\tilde{\mathbf{x}}^{t-1} -\mathbf{x}^{t-1} \|^2 
\end{equation}
By substituting the bounds for $T_3$ and $T_4$ in Eq.~\ref{eq:b_t2} i.e $T_2$. Further substituting the $T_1$ and $T_2$ in the Eq.~\ref{eq:norm_x_xtilde} we have the following

By using $T_3$ and $T_4$ we bound the $T_2$ as follows 

\begin{equation*}
T_2 \leq 2\eta_xL\gamma^{t-1} + \frac{2 \eta_x}{n}\sum_{i=1}^{n} \Delta_i^{t-1} -\frac{\eta_xL}{2} \|\tilde{\mathbf{x}}^{t-1} -\mathbf{x}^{t-1} \|^2 
\end{equation*}
Here we have defined 

\begin{equation}
\gamma^t \coloneqq \frac{1}{n}\sum_{i=1}^{n}\|\mathbf{x}_i^t - \mathbf{x}^t  \|^2
\end{equation}

\begin{equation}
\begin{aligned}
\mathbb{E}\lVert \tilde{\mathbf{x}}^{t-1} - \mathbf{x}^t \rVert^2 
&=  \mathbb{E}\| \tilde{\mathbf{x}}^{t-1} -\mathbf{x}^{t-1}\|^2 + \eta_x^2 \bm(\frac{\sigma^2}{n} + G_x)\\
&+ 2\eta_xL\gamma^{t-1} + \frac{2 \eta_x}{n}\sum_{i=1}^{n} \Delta_i^{t-1} -\frac{\eta_xL}{2} \|\tilde{\mathbf{x}}^{t-1} -\mathbf{x}^{t-1} \|^2
\end{aligned}
\label{eq:norm_x_xtilde}
\end{equation}

By substituting the Eq.~\ref{eq:norm_x_xtilde} in Eq.~\ref{me_ineq1} we get the following  

\begin{equation}
\begin{aligned}
\mathbb{E}[\Phi_{1/2L}(\mathbf{x}^{t})] &\leq \mathbb{E}[\Phi(\tilde{\mathbf{x}}^{t-1})] + L\mathbb{E}\| \tilde{\mathbf{x}}^{t-1} -\mathbf{x}^{t-1}\|^2 + L\eta_x^2 \bm(\frac{\sigma^2}{n} + G_x)\\
&+ 2\eta_xL^2 \mathbb{E} [\gamma^{t-1}] + \frac{2L \eta_x}{n}\sum_{i=1}^{n} \mathbb{E}\Delta_i^{t-1} -\frac{\eta_xL^2}{2} \mathbb{E}\|\tilde{\mathbf{x}}^{t-1} -\mathbf{x}^{t-1} \|^2\\
&= \mathbb{E}[\Phi_{1/2L}(\mathbf{x}^{t-1})] + L\eta_x^2 \bm(\frac{\sigma^2}{n} + G_x) + 2\eta_xL^2\mathbb{E}[\gamma^{t-1}] \\ 
&+ \frac{2L \eta_x}{n}\sum_{i=1}^{n} \mathbb{E}[\Delta_i^{t-1}] -\frac{\eta_x}{8} \mathbb{E}\|\nabla \Phi_{1/2L}(\mathbf{x}^{t-1})  \|^2 
\end{aligned}
\end{equation}

In the above, we get the second equality by using the definition of Moreau Envelope~\ref{eq:me_def} and the identity $2L(\tilde{\mathbf{x}}^{t-1} -\mathbf{x}^{t-1}) = \nabla \Phi_{1/2L}(\mathbf{x}^{t-1})$.

\end{proof}

In the following, we now bound the quantities $\Delta^{t}$ and the synchronization error $\gamma^t$.
\begin{lemma}
$ \frac{1}{n} \sum_{i=1}^{n} \frac{1}{T}\sum_{t=1}^{T}
\mathbb{E}\!\left[\Phi_i\!\left(\mathbf{x}^{(t-1)}\right)\right]
- \mathbb{E}\!\left[f_i\!\left(\mathbf{x}^{(t-1)}, \bm{y}_i^{(t-1)}\right)\right]
 \leq 2 G_x \eta_x \sqrt{G_x^2 +\frac{\sigma^2}{n}} + \frac{D^2}{\tau\eta_y} +\eta_y\sigma^2 $
\end{lemma}
\label{lm:dual_track}
\begin{proof}

We decompose the \\
$ \Delta_i^{t-1} = \mathbb{E}\left[\Phi_i\!\left(\mathbf{x}^{(t-1)}\right)\right]
- \mathbb{E}\!\left[f_i\left(\mathbf{x}^{(t-1)}, \bm{y}_i^{(t-1)}\right)\right]$ 

By the definition of $\Phi(.)$ we have the following 
\[ \Delta_i^{t} = \mathbb{E}\left[f_i\left(\mathbf{x}^{t},\mathbf{y}_i^{*}(\mathbf{x}^{t})\right)- f_i\left(\mathbf{x}^{t}, \mathbf{y}_i^{t}\right)\right]\]

\begin{equation}
\begin{aligned}
\mathbb{E}\left[f_i(\mathbf{x}^{t},\mathbf{y}_i^{*}(\mathbf{x}^{t}))- f_i(\mathbf{x}^{t}, \mathbf{y}_i^{t})\right] 
&\leq 
\underbrace{\mathbb{E}\left[f_i(\mathbf{x}^{t},\mathbf{y}_i^{*}(\mathbf{x}^{t}))- f_i(\mathbf{x}^{s}, \mathbf{y}_i^{*}(\mathbf{x}^t))\right]}_{T_1}+ 
\underbrace{\mathbb{E}\left[f_i(\mathbf{x}^{s},\mathbf{y}_i^{*}(\mathbf{x}^{s}))- f_i(\mathbf{x}^{s}, \mathbf{y}_i^{t})\right]}_{T_2}\\
&+ 
\underbrace{\mathbb{E}\left[f_i(\mathbf{x}^{s},\mathbf{y}_i^{t})- f_i(\mathbf{x}^{t}, \mathbf{y}_i^{t})\right]}_{T_3}
\end{aligned}
\label{eq:Delta_t}
\end{equation}

Here we take $s\leq t-1$, the inequality follows from  $f_i(\mathbf{x}^{s}, \mathbf{y}_i^{*}(\mathbf{x}^t)) \leq f_i(\mathbf{x}^{s},\mathbf{y}_i^{*}(\mathbf{x}^{s}))$.

\begin{equation}
\begin{aligned}
T_1+T_3 &\leq 2 G_x\mathbb{E}\lVert \mathbf{x}^t - \mathbf{x}^{s} \rVert \\
        &= 2 G_x \eta_x \mathbb{E}\lVert \frac{1}{n} \sum_{i=1}^{n} \sum_{k=s}^{t-1}\nabla_{\mathbf{x}} f_i(\mathbf{x}_i^{k},\mathbf{y}_i^{k};\xi_i^{k}) \rVert\\
        &\leq 2 G_x \eta_x  \sum_{k=s}^{t-1} \mathbb{E}\lVert \frac{1}{n} \sum_{i=1}^{n} \nabla_{\mathbf{x}} f_i(\mathbf{x}_i^{k},\mathbf{y}_i^{k};\xi_i^{k}) \rVert\\
        &\leq 2 G_x \eta_x \sum_{k=s}^{t-1} \sqrt{\mathbb{E}\lVert \frac{1}{n} \sum_{i=1}^{n} \nabla_\mathbf{x} f_i(\mathbf{x}_i^{k},\mathbf{y}_i^{k};\xi_i^{k}) \rVert^2} \\
        &\leq  2 G_x \eta_x (t-s) \sqrt{G_x^2 +\frac{\sigma^2}{n}} \\
        &\leq  2 G_x \eta_x (t-s) \sqrt{G_x^2 +{\sigma^2}}
\end{aligned}
\label{all_client_drift}
\end{equation}

The first inequality follows from the assumption of Lipchitz continuous in $\mathbf{x}$. The subsequent inequality uses the Jensen's and the bounded variance assumption.

\begin{equation}
\begin{aligned}
T_2 =& \mathbb{E}\left[f_i(\mathbf{x}_i^{(t-1)},\mathbf{y}_i^{*}(\mathbf{x}^{s}))- f_i(\mathbf{x}_i^{(t-1)}, \mathbf{y}_i^{t})\right] \\
+&  \mathbb{E}\left[f_i(\mathbf{x}^s,\mathbf{y}_i^*(\mathbf{x^s})) - f_i(\mathbf{x}_i^{t-1},\mathbf{y}_i^*(\mathbf{x^s})) + f_i(\mathbf{x}_i^{t-1},\mathbf{y}_i^t) - f_i(\mathbf{x}^{s},\mathbf{y}_i^t) \right] 
\end{aligned}
\end{equation}

We now bound the  first expectation, following the lemma.~\ref{lem:sga-descent} we have the following bound with $\eta_y \leq \frac{1}{2L}$

\begin{equation*}
\mathbb{E}[f(\mathbf{x}^{t-1},\mathbf{y}) - f(\mathbf{x}^{t-1},\mathbf{y}^t)]
\leq
\frac{1}{2\eta_y}\Big(\mathbb{E}[\lVert \mathbf{y} - \mathbf{y}^{t-1}\rVert^2]
- \mathbb{E}[\lVert \mathbf{y} - \mathbf{y}^t\rVert^2]\Big)
+ \eta_y \sigma^2.
\end{equation*}

By substituting $\mathbf{y}$ as $ \mathbf{y}_i^*(\mathbf{x}^{s})$ and $\mathbf{y}^{t}$ as $\mathbf{y}_i^t$. And using the fact the diameter of the set $\mathcal{Y}_i$ is $D_i$ and let $D$ be the maximum of all $D_i$, we bound the term $T_2$ as below   

\begin{equation}
\begin{aligned}
&\mathbb{E}\left[f_i(\mathbf{x}_i^{(t-1)},\mathbf{y}_i^{*}(\mathbf{x}^{s}))- f_i(\mathbf{x}_i^{(t-1)}, \mathbf{y}_i^{t})\right]
\leq \frac{1}{2\eta_y}\Big(\mathbb{E}[\lVert \mathbf{y}_i^{*}(\mathbf{x}^{s}) - \mathbf{y}_i^{t-1}\rVert^2]
- \mathbb{E}[\lVert \mathbf{y}_i^{*}(\mathbf{x}^{s}) - \mathbf{y}_i^t\rVert^2]) + \eta_y\sigma^2
\end{aligned}
\end{equation}

\begin{equation}
T_2 \leq \frac{D^2}{2\tau \eta_y} + \eta_y \sigma^2 + 2G_x\lVert \mathbf{x}_i^{t-1} - \mathbf{x}^s\rVert
\end{equation}

we now bound 
\[ \frac{1}{T}\sum_{t=1}^{T} \Delta_i^t = \frac{\tau}{T} \sum_{r=0}^{\frac{T-\tau}{\tau}} \frac{1}{\tau} \sum_{t=r\tau}^{(r+1)\tau-1} \Delta_i^{t} \]

We assume that $\frac{T}{\tau}$ is an integer.
Under the synchronization rule of Algorithm~\ref{alg:merged_sgda} (the reset fires when $\tau \mid (t{+}1)$), consecutive synchronizations
occur at indices $t = \tau-1, 2\tau-1, 3\tau-1, \ldots$, so the fresh blocks
begin at $t = 0, \tau, 2\tau, \ldots$ Writing block $r$ as the index range
$t \in \{r\tau, r\tau+1, \ldots, (r+1)\tau - 1\}$ and the block-start index
as $s = r\tau$, the reset at the start of the block gives
\begin{equation}\label{eq:blockstart}
  \mathbf{x}_i^{s} \;=\; x^{s}\qquad\text{for every client } i,
\end{equation}
an equality that holds \emph{only} at the block-start index $t = s$ (it does
not hold at $t > s$ within the same block, because clients then drift
independently until the next synchronization). Substituting
\eqref{eq:blockstart} into the bridge term replaces the global $x^{s}$ by
client $i$'s own block-start iterate,
\[
  \bigl\lVert x^{s} - \mathbf{x}_i^{t-1} \bigr\rVert
  \;=\; \bigl\lVert \mathbf{x}_i^{s} - \mathbf{x}_i^{t-1} \bigr\rVert,
\]
a single client's drift from its own starting point over the local steps
$k = s, \ldots, t-2$. Since the next synchronization after $s$ occurs at
$s + \tau - 1$, and the step indices $k = s, \ldots, t-2$ satisfy
$k \le t-2 \le s + \tau - 2 < s + \tau - 1$, none of these steps is itself
a synchronization: the range $k = s, \ldots, t-2$ is a single uninterrupted
local trajectory on client $i$. Following the single-client analysis of
Eq.~\eqref{all_client_drift} (variance $\sigma^{2}$),
\begin{equation}
\label{eq:driftB}
\begin{aligned}
&\mathbf{x}_i^{t-1} - \mathbf{x}_i^{s}
    \;=\; -\eta_x \sum_{k=s}^{t-2} \nabla_x f_i\!\bigl(\mathbf{x}_i^{k}, \mathbf{y}_i^{k}; \xi_i^{k}\bigr),\\
&\mathbb{E}\bigl\lVert \mathbf{x}_i^{t-1} - \mathbf{x}_i^{s} \bigr\rVert
    \;\le\; \eta_x (t-1-s) \sqrt{G_x^{2} + \sigma^{2}}
    \;\le\; \eta_x (\tau - 1) \sqrt{G_x^{2} + \sigma^{2}}.
\end{aligned}    
\end{equation}
The drift vanishes at $t-1 = s$ (an empty sum) and grows linearly over the
$\tau - 1$ local steps available within a block. Combining with the global
progress term gives the per-iteration bound
\[
  T_2 \;\le\; 2\eta_x (\tau - 1) \sqrt{G_x^{2} + \sigma^{2}},
\]

\paragraph{Averaging over the trajectory.} Substituting the bounds
$T_1 + T_3$ and $T_2$ into Eq.~\ref{eq:Delta_t} and averaging
$\Delta_i^t$ over $t = 0, \ldots, T-1$ via the block decomposition
\[
  \frac{1}{T} \sum_{t=0}^{T-1} \Delta_i^t
  \;=\; \frac{1}{T} \sum_{r=0}^{T/\tau - 1} \sum_{t=r\tau}^{(r+1)\tau - 1}
       \Delta_i^t,
\]
where $s = r\tau$ is used within each block, gives
\[
  \frac{1}{T} \sum_{t=0}^{T-1} \Delta_i^t
  \;\le\; 4\,G_x\,\eta_x\,\tau\,\sqrt{G_x^{2} + \sigma^{2}}
       + \frac{D^{2}}{2\tau\eta_y}
       + \eta_y \sigma^{2}.
\]
\end{proof}

The block-structured inner analysis of \citet{linjinjordan2020,lin2025twotimescale}, in which the inner variable is updated through $\tau$ consecutive steps with intra-block drift controlled by stepsize separation, was originally introduced as an algorithmic batching device in the centralized setting; the parameter $\tau$
 had no operational meaning beyond internal accounting. We observe that the same algebraic structure admits a natural federated reinterpretation: the block boundary is precisely a synchronization point, and $\tau$
is the communication gap, the number of local steps between successive server-synchronizations. Under this reinterpretation, the algebra (single-client drift over an uninterrupted local trajectory) becomes a direct statement about federated communication architecture, and the centralized two-timescale guarantee transfers additively, with the bound reducing exactly to its centralized form at $\tau = 1$
(full synchronization). The contribution is not in the intra-block analysis itself, which is Lin's, but in identifying that Lin's block primitive is operationally a communication-interval primitive, a reinterpretation that, combined with the personalized maximization variable, removes the consensus-error machinery prior federated minimax analyses required.

\begin{lemma}
$\frac{1}{T} \sum_{t=0}^{T-1} \gamma^t = 10 \tau^2\eta_x^2\bm(\sigma^2+\frac{\sigma^2}{n}) + 10 \tau^2 \eta_x^2\zeta_x^2$. As long as  $\eta_x$ and $\tau$ are chosen to satisfy  $10L^2\tau^2\eta_x^2\leq \frac{1}{2}$
\label{lm:sync_error}
\end{lemma}

\begin{proof}
Let $s = r\tau$ denote the start of the current synchronization interval such that $s \le t < s + \tau$. Under the execution rules of Algorithm 1, all local primal models are synchronized at the beginning of each communication block:
\begin{equation}
    \mathbf{x}_i^s = x^s, \quad \forall i \in [n],
\end{equation}
which implies that the initial consensus error at the block boundary is zero ($\gamma^s = 0$). 

For any iteration $t$ within the current block, the deviation of client $i$'s local model from the network average is determined by the accumulation of local gradients since the last synchronization point $s$:
\begin{equation}
\begin{aligned}
   & x^{t} - x_{i}^{t} \\
   &= \left(x^{s} - \frac{\eta_x}{n}\sum_{k=1}^{n}\sum_{j=s}^{t-1}\nabla_x f_k(x_k^j, y_k^j; \xi_k^j)\right) - \left(x^{s} - \eta_x\sum_{j=s}^{t-1}\nabla_x f_i(\mathbf{x}_i^j, \mathbf{y}_i^j; \xi_i^j)\right).
\end{aligned}
\end{equation}
Regrouping these terms allows us to express the discrepancy as a combination of local and global update trajectories:
\begin{equation}
    x^{t} - x_{i}^{t} = \frac{\eta_x}{n}\sum_{k=1}^{n}\sum_{j=s}^{t-1}\left[\nabla_x f_i(\mathbf{x}_i^j, \mathbf{y}_i^j; \xi_i^j) - \nabla_x f_k(x_k^j, y_k^j; \xi_k^j)\right].
\end{equation}
To isolate the separate effects of sampling noise, local iterate drift, and spatial data heterogeneity, we expand the inner summand by adding and subtracting the true local gradients $\nabla_x f_i(\mathbf{x}_i^j, \mathbf{y}_i^j)$, the globalized client gradients at the synchronized state $\nabla_x f_i(\mathbf{x}^j, \mathbf{y}_i^j)$, and their cross-client counterparts:
\begin{align}
    x^{t} - x_{i}^{t} = \eta_x\sum_{j=s}^{t-1}\Bigg\{ 
    &\left(\nabla_x f_i(\mathbf{x}_i^j, \mathbf{y}_i^j; \xi_i^j) - \nabla_x f_i(\mathbf{x}_i^j, \mathbf{y}_i^j)\right) \nonumber \\
    + &\left(\nabla_x f_i(\mathbf{x}_i^j, \mathbf{y}_i^j) - \nabla_x f_i(x^j, \mathbf{y}_i^j)\right) \nonumber \\
    + &\left(\nabla_x f_i(\mathbf{x}^j, \mathbf{y}_i^j) - \frac{1}{n}\sum_{k=1}^{n}\nabla_x f_k(x^j, y_k^j)\right) \nonumber \\
    + &\frac{1}{n}\sum_{k=1}^{n}\left(\nabla_x f_k(\mathbf{x}^j, \mathbf{y}_k^j) - \nabla_x f_k(\mathbf{x}_k^j, \mathbf{y}_k^j)\right) \nonumber \\
    + &\frac{1}{n}\sum_{k=1}^{n}\left(\nabla_x f_k(\mathbf{x}_k^j, \mathbf{y}_k^j) - \nabla_x f_k(\mathbf{x}_k^j, \mathbf{y}_k^j; \xi_k^j)\right) \Bigg\}.
\end{align}
Taking the squared norm, applying the conditional expectation $\mathbb{E}[\cdot]$, and utilizing the algebraic identity $\|\sum_{m=1}^5 a_m\|^2 \le 5\sum_{m=1}^5 \|a_m\|^2$ alongside Jensen's Inequality over the $(t-s) \le \tau$ steps inside the block outer sum establishes:
\begin{align}
    \mathbb{E}[\|x^t - \mathbf{x}_i^t\|^2] \le 5\tau\eta_x^2\sum_{j=s}^{t-1}\mathbb{E}\Bigg[ 
    &\|\nabla_x f_i(\mathbf{x}_i^j, \mathbf{y}_i^j; \xi_i^j) - \nabla_x f_i(\mathbf{x}_i^j, \mathbf{y}_i^j)\|^2 \nonumber \\
    + &\|\nabla_x f_i(\mathbf{x}_i^j, \mathbf{y}_i^j) - \nabla_x f_i(x^j, \mathbf{y}_i^j)\|^2 \nonumber \\
    + &\|\nabla_x f_i(x^j, \mathbf{y}_i^j) - \nabla_x f(x^j, Y^j)\|^2 \nonumber \\
    + &\frac{1}{n}\sum_{k=1}^{n}\|\nabla_x f_k(x^j, y_k^j) - \nabla_x f_k(x_k^j, y_k^j)\|^2 \nonumber \\
    + &\frac{1}{n}\sum_{k=1}^{n}\|\nabla_x f_k(x_k^j, y_k^j) - \nabla_x f_k(x_k^j, y_k^j; \xi_k^j)\|^2 \Bigg].
\end{align}
Averaging across all clients $i \in [n]$, we bound the individual components by applying our core assumptions:
\begin{enumerate}
    \item \textbf{Bounded Variance ($A_2$):} The local stochastic gradient variance satisfies $\mathbb{E}[\|\nabla_x f_i(\mathbf{x}_i^j, \mathbf{y}_i^j; \xi_i^j) - \nabla_x f_i(\mathbf{x}_i^j, \mathbf{y}_i^j)\|^2] \le \sigma_x^2$, and the network-wide average sampling variance is bounded by $\frac{\sigma_x^2}{n}$.
    \item \textbf{Primal Smoothness ($A_1$):} Under $L$-smoothness, the iterate deviations translate directly to gradient bounds: $\|\nabla_x f_i(\mathbf{x}_i^j, \mathbf{y}_i^j) - \nabla_x f_i(x^j, \mathbf{y}_i^j)\|^2 \le L^2\|\mathbf{x}_i^j - x^j\|^2$.
    \item \textbf{Gradient Dissimilarity:} The spatial dataset heterogeneity across the network is bounded by $\zeta_x^2$.
\end{enumerate}

Substituting these bounds into our consensus metric\\
$\gamma^t := \frac{1}{n}\sum_{i=1}^{n}\mathbb{E}[\|x^{t}-x_{i}^{t}\|^{2}]$ simplifies the recurrence relation to:
\begin{equation}
    \gamma^t \le 5\tau\eta_x^2\sum_{j=s}^{t-1}\left[ 2L^2\gamma^j + \sigma_x^2\left(1 + \frac{1}{n}\right) + \zeta_x^2 \right],
\end{equation}
\begin{equation}
    \gamma^t \le 10L^2\tau\eta_x^2\sum_{j=s}^{t-1}\gamma^j + 5\tau^2\eta_x^2\left[\sigma_x^2\left(1 + \frac{1}{n}\right) + \zeta_x^2\right].
\end{equation}

Summing this relationship over all iterations within the single communication interval block $t = s, \dots, s+\tau-1$ yields:
\begin{equation}
    \sum_{t=s}^{s+\tau-1}\gamma^t \le \sum_{t=s}^{s+\tau-1} \left( 10L^2\tau\eta_x^2\sum_{j=s}^{t-1}\gamma^j \right) + \sum_{t=s}^{s+\tau-1} \left( 5\tau^2\eta_x^2\left[\sigma_x^2\left(1 + \frac{1}{n}\right) + \zeta_x^2\right] \right).
\end{equation}
Because $\gamma^j \ge 0$, we can strictly upper-bound the inner recursive sum by extending its limit to the end of the block ($s+\tau-1$). This removes its dependence on $t$, allowing us to evaluate the outer sums by multiplying by $\tau$:
\begin{equation}
    \sum_{t=s}^{s+\tau-1}\gamma^t \le 10L^2\tau^2\eta_x^2\sum_{t=s}^{s+\tau-1}\gamma^t + 5\tau^3\eta_x^2\left[\sigma_x^2\left(1 + \frac{1}{n}\right) + \zeta_x^2\right].
\end{equation}
By enforcing the step size stability constraint $10L^2\tau^2\eta_x^2 \le \frac{1}{2}$, we move the recursive summation to the left-hand side:
\begin{equation}
    \left(1 - 10L^2\tau^2\eta_x^2\right)\sum_{t=s}^{s+\tau-1}\gamma^t \le 5\tau^3\eta_x^2\left[\sigma_x^2\left(1 + \frac{1}{n}\right) + \zeta_x^2\right],
\end{equation}
\begin{equation}
    \sum_{t=s}^{s+\tau-1}\gamma^t \le 10\tau^3\eta_x^2\left[\sigma_x^2\left(1 + \frac{1}{n}\right) + \zeta_x^2\right].
\end{equation}
To calculate the total consensus error over the entire training envelope, we sum over all $M = \frac{T}{\tau}$ blocks across the horizon:
\begin{align}
    \frac{1}{T}\sum_{t=0}^{T-1}\gamma^t &= \frac{1}{T}\sum_{m=0}^{\frac{T}{\tau}-1}\left(\sum_{t=m\tau}^{(m+1)\tau-1}\gamma^t\right) \nonumber \\
    &\le \frac{1}{T}\left(\frac{T}{\tau}\right) \cdot 10\tau^3\eta_x^2\left[\sigma_x^2\left(1 + \frac{1}{n}\right) + \zeta_x^2\right] \nonumber \\
    &\le 10\tau^2\eta_x^2\left[\sigma_x^2\left(1 + \frac{1}{n}\right) + \zeta_x^2\right].
\end{align}
Distributing the product parameters matches the final analytical goal statement:
\begin{equation}
    \frac{1}{T}\sum_{t=0}^{T-1}\gamma^{t} = 10\tau^2\eta_x^2\left(\sigma_x^2 + \frac{\sigma_x^2}{n}\right) + 10\tau^2\eta_x^2\zeta_x^2.
\end{equation}
\end{proof}

\begin{lemma}[Projected SGA descent for smooth concave functions]
\label{lem:sga-descent}
Let $g : \mathbb{R}^{d_y} \to \mathbb{R}$ be concave and $L$-smooth, $\mathcal{Y}
\subseteq \mathbb{R}^{d_y}$ closed and convex, and let
$\tilde\nabla g(y; \xi)$ be an unbiased stochastic gradient with bounded
variance, i.e.\
$\mathbb{E}[\tilde\nabla g(y;\xi)] = \nabla g(y)$ and
$\mathbb{E}\|\tilde\nabla g(y;\xi) - \nabla g(y)\|^2 \le \sigma^2$.
Consider the projected stochastic gradient ascent update
\[
  y^{t+1} \;=\; \Pi_{\mathcal{Y}}\!\left( y^t + \eta_y\,\tilde\nabla g(y^t; \xi^t) \right).
\]
If the stepsize satisfies $\eta_y \le \tfrac{1}{2L}$, then for every
$y \in \mathcal{Y}$,
\begin{equation}\label{eq:sga-descent}
  \mathbb{E}\!\left[g(y) - g(y^{t+1})\right]
  \;\le\;
  \frac{1}{2\eta_y}\Big(\,\mathbb{E}\|y - y^t\|^2 \;-\; \mathbb{E}\|y - y^{t+1}\|^2\,\Big)
  \;+\; \eta_y \sigma^2 .
\end{equation}
\end{lemma}

\begin{proof}
We proceed in three steps: a projection inequality, a smoothness-and-concavity
bound on the deterministic part, and a noise-control step using Young's
inequality with a constant chosen to be absorbed by the smoothness slack.

For any closed convex $\mathcal{Y}$, the projection $\Pi_{\mathcal{Y}}$
satisfies, for every $z \in \mathbb{R}^{d_y}$ and $y \in \mathcal{Y}$,
\[
  \langle z - \Pi_{\mathcal{Y}}(z),\; y - \Pi_{\mathcal{Y}}(z)\rangle \;\le\; 0.
\]
Applying this with $z = y^t + \eta_y \tilde\nabla g(y^t;\xi^t)$ and
$\Pi_{\mathcal{Y}}(z) = y^{t+1}$ gives
\[
  \big\langle y^t + \eta_y\tilde\nabla g(y^t;\xi^t) - y^{t+1},\; y - y^{t+1}\big\rangle \;\le\; 0,
\]
so that
\begin{equation}\label{eq:proj}
  \langle y^t - y^{t+1},\; y - y^{t+1}\rangle
  \;\le\;
  \eta_y\,\big\langle \tilde\nabla g(y^t;\xi^t),\; y^{t+1} - y\big\rangle .
\end{equation}

Using the three-point identity
$\langle a-b,c-b\rangle = \tfrac12\|a-b\|^2 + \tfrac12\|c-b\|^2 - \tfrac12\|a-c\|^2$
with $a = y^t$, $b = y^{t+1}$, $c = y$, the left side of \eqref{eq:proj} becomes
$\tfrac12\|y^t - y^{t+1}\|^2 + \tfrac12\|y - y^{t+1}\|^2 - \tfrac12\|y - y^t\|^2$.
Multiplying by $2$ and rearranging,
\begin{equation}\label{eq:proj2}
  \|y - y^{t+1}\|^2
  \;\le\;
  \|y - y^t\|^2 \;-\; \|y^t - y^{t+1}\|^2
  \;+\; 2\eta_y\,\big\langle \tilde\nabla g(y^t;\xi^t),\; y^{t+1} - y\big\rangle .
\end{equation}

Write
\[
\begin{aligned}
  \langle \tilde\nabla g(y^t;\xi^t),\; y^{t+1} - y\rangle 
  &= \langle \nabla g(y^t),\; y^{t+1} - y\rangle \\
  &\quad + \langle \tilde\nabla g(y^t;\xi^t) - \nabla g(y^t),\; y^{t+1} - y\rangle,
\end{aligned}
\]
and split the first term using $y^{t+1} - y = (y^{t+1} - y^t) + (y^t - y)$.

For the displacement piece, $L$-smoothness of $g$ gives
\[
  g(y^{t+1}) \;\ge\; g(y^t) + \langle \nabla g(y^t),\; y^{t+1} - y^t\rangle - \tfrac{L}{2}\|y^{t+1} - y^t\|^2,
\]
equivalently
\begin{equation}\label{eq:smooth}
  \langle \nabla g(y^t),\; y^{t+1} - y^t\rangle
  \;\le\;
  g(y^{t+1}) - g(y^t) + \tfrac{L}{2}\|y^{t+1} - y^t\|^2 .
\end{equation}
For the comparator piece, concavity of $g$ gives
$g(y) \le g(y^t) + \langle \nabla g(y^t),\; y - y^t\rangle$, equivalently
\begin{equation}\label{eq:concave}
  \langle \nabla g(y^t),\; y^t - y\rangle \;\le\; g(y^t) - g(y) .
\end{equation}
Adding \eqref{eq:smooth} and \eqref{eq:concave},
\begin{equation}\label{eq:det}
  \langle \nabla g(y^t),\; y^{t+1} - y\rangle
  \;\le\;
  g(y^{t+1}) - g(y) + \tfrac{L}{2}\|y^{t+1} - y^t\|^2 .
\end{equation}

Substituting back into \eqref{eq:proj2},
\begin{align*}
  \|y - y^{t+1}\|^2
  &\;\le\;
  \|y - y^t\|^2 - \|y^t - y^{t+1}\|^2 \\
  &\quad{}+ 2\eta_y\!\left[ g(y^{t+1}) - g(y) + \tfrac{L}{2}\|y^{t+1} - y^t\|^2 \right] \\
  &\quad{}+ 2\eta_y\,\big\langle \tilde\nabla g(y^t;\xi^t) - \nabla g(y^t),\; y^{t+1} - y\big\rangle .
\end{align*}
Collecting the deterministic terms and isolating $g(y) - g(y^{t+1})$,
\begin{align}
  2\eta_y\big[g(y) - g(y^{t+1})\big]
  &\;\le\;
  \|y - y^t\|^2 - \|y - y^{t+1}\|^2\notag\\
  &\;-\; (1 - \eta_y L)\|y^t - y^{t+1}\|^2 \notag\\
  &+ 2\eta_y\,\big\langle \tilde\nabla g(y^t;\xi^t) - \nabla g(y^t),\; y^{t+1} - y\big\rangle .
  \label{eq:det-bound}
\end{align}
The stepsize condition $\eta_y \le \tfrac{1}{2L}$ ensures
$1 - \eta_y L \ge \tfrac{1}{2} > 0$, so the term
$-(1 - \eta_y L)\|y^t - y^{t+1}\|^2$ is nonpositive and provides slack to
absorb the residual that arises from the noise inner product.

Decompose $y^{t+1} - y = (y^{t+1} - y^t) + (y^t - y)$. Conditional on the
history $\mathcal{F}_{t-1}$, the iterate $y^t$ is measurable and the noise
$\tilde\nabla g(y^t;\xi^t) - \nabla g(y^t)$ is zero-mean, so
\[
  \mathbb{E}\!\left[ \big\langle \tilde\nabla g(y^t;\xi^t) - \nabla g(y^t),\; y^t - y\big\rangle \;\Big|\; \mathcal{F}_{t-1} \right] \;=\; 0 .
\]
Only the displacement component contributes after taking expectations. By
Young's inequality with constant $\alpha > 0$,
\[
\begin{aligned}
  2\eta_y\,\big\langle \tilde\nabla g(y^t;\xi^t) - \nabla g(y^t),\; y^{t+1} - y^t\big\rangle
  &\;\le\;
  \frac{\eta_y^2}{\alpha}\,\|\tilde\nabla g(y^t;\xi^t) - \nabla g(y^t)\|^2\\
  &\;+\; \alpha\,\|y^{t+1} - y^t\|^2 .
\end{aligned}
\]
We choose $\alpha = 1 - \eta_y L$ so that the squared-norm residual exactly
cancels the slack in \eqref{eq:det-bound}. Taking expectations and applying the
variance bound,
\[
  \mathbb{E}\!\left[ 2\eta_y\,\big\langle \tilde\nabla g - \nabla g,\; y^{t+1} - y\big\rangle\right]
  \;\le\;
  \frac{\eta_y^2}{1 - \eta_y L}\,\sigma^2
  \;+\; (1 - \eta_y L)\,\mathbb{E}\|y^{t+1} - y^t\|^2 .
\]
Substituting this bound into the expectation of \eqref{eq:det-bound}, the
$(1 - \eta_y L)\|y^t - y^{t+1}\|^2$ slack and residual cancel, leaving
\[
  2\eta_y\,\mathbb{E}\!\left[g(y) - g(y^{t+1})\right]
  \;\le\;
  \mathbb{E}\|y - y^t\|^2 - \mathbb{E}\|y - y^{t+1}\|^2
  \;+\; \frac{\eta_y^2\,\sigma^2}{1 - \eta_y L} .
\]
Under $\eta_y \le \tfrac{1}{2L}$ we have $1 - \eta_y L \ge \tfrac{1}{2}$,
giving $\tfrac{\eta_y^2 \sigma^2}{1 - \eta_y L} \le 2\eta_y^2 \sigma^2$.
Dividing through by $2\eta_y$ yields \eqref{eq:sga-descent}.
\end{proof}

\subsection{Nonconvex-PL Convergence}
\subsection*{Proofs of supporting Lemmas for nonconvex-PL}
\begin{lemma}[One-step outer descent]
\label{lem:outer-descent}
Under Assumptions~\ref{assm:lip-sm} to~\ref{assm:pl}, the  SGDA iterates of Algorithm~\ref{alg:merged_sgda} satisfy
\begin{align*}
\mathbb{E}\!\left[\Phi(\mathbf{x}^{t+1}) - \Phi(\mathbf{x}^t)\right]
&\le -\tfrac{\eta_x}{2}\,\mathbb{E}\!\left\|\nabla\Phi(\mathbf{x}^t)\right\|^2- \left(\tfrac{\eta_x}{2} - \tfrac{\beta\eta_x^2}{2}\right)
\mathbb{E}\!\left\|\tfrac{1}{n}\sum_{i=1}^n \nabla_{\mathbf{x}} f_i(\mathbf{x}_i^t, \mathbf{y}_i^t)\right\|^2 \\
&\quad + \tfrac{2\eta_x L^2}{\mu}\, A_t
+ \tfrac{\eta_x L^2}{n}\sum_{i=1}^n \mathbb{E}\!\left\|\mathbf{x}^t - \mathbf{x}_i^t\right\|^2
+ \tfrac{\beta\eta_x^2 \sigma_x^2}{2n}. 
\end{align*}
where $A_t = \mathbb{E}[\Phi(\mathbf{x}^t) - F(\mathbf{x}^t, \mathbf{Y}^t)]$.
\end{lemma}
 
\begin{proof}
By $\beta$-smoothness of $\Phi$ as proven in \ref{lem:danskin_constrained}, applied at $\mathbf{x}^{t+1}$ and $\mathbf{x}^t$,
\begin{equation}
\Phi(\mathbf{x}^{t+1}) \le \Phi(\mathbf{x}^t)
+ \langle \nabla\Phi(\mathbf{x}^t),\, \mathbf{x}^{t+1} - \mathbf{x}^t \rangle
+ \tfrac{\beta}{2}\left\|\mathbf{x}^{t+1} - \mathbf{x}^t\right\|^2.
\label{eq:smooth}
\end{equation}
Substituting the update rule $\mathbf{x}^{t+1} - \mathbf{x}^t = -\eta_x \cdot \tfrac{1}{n}\sum_{i=1}^n \nabla_{\mathbf{x}} f_i(\mathbf{x}_i^t, \mathbf{y}_i^t; \xi_i^t)$ into~\eqref{eq:smooth},
\begin{equation}
\Phi(\mathbf{x}^{t+1}) \le \Phi(\mathbf{x}^t)
- \eta_x \left\langle \nabla\Phi(\mathbf{x}^t),\, \tfrac{1}{n}\sum_{i=1}^n \nabla_{\mathbf{x}} f_i(\mathbf{x}_i^t, \mathbf{y}_i^t; \xi_i^t) \right\rangle 
+ \tfrac{\beta\eta_x^2}{2}\left\|\tfrac{1}{n}\sum_{i=1}^n \nabla_{\mathbf{x}} f_i(\mathbf{x}_i^t, \mathbf{y}_i^t; \xi_i^t)\right\|^2.
\label{eq:step1}
\end{equation}
 
Taking expectations, using the unbiasedness of stochastic gradients (Assumption~\ref{assm:bd_var}) on the inner-product term and the bias--variance decomposition on the squared-norm term,
\begin{equation}
\mathbb{E}\!\left\|\tfrac{1}{n}\sum_{i=1}^n \nabla_{\mathbf{x}} f_i(\mathbf{x}_i^t, \mathbf{y}_i^t; \xi_i^t)\right\|^2
= \mathbb{E}\!\left\|\tfrac{1}{n}\sum_{i=1}^n \nabla_{\mathbf{x}} f_i(\mathbf{x}_i^t, \mathbf{y}_i^t)\right\|^2
+ \tfrac{\sigma_x^2}{n}.
\label{eq:var}
\end{equation}
Substituting~\eqref{eq:var} into~\eqref{eq:step1} after taking expectations,
\begin{multline}
\mathbb{E}[\Phi(\mathbf{x}^{t+1})]
\le \mathbb{E}[\Phi(\mathbf{x}^t)]
- \eta_x \mathbb{E}\!\left\langle \nabla\Phi(\mathbf{x}^t),\, \tfrac{1}{n}\sum_{i=1}^n \nabla_{\mathbf{x}} f_i(\mathbf{x}_i^t, \mathbf{y}_i^t) \right\rangle
+ \tfrac{\beta\eta_x^2}{2} \mathbb{E}\!\left\|\tfrac{1}{n}\sum_{i=1}^n \nabla_{\mathbf{x}} f_i(\mathbf{x}_i^t, \mathbf{y}_i^t)\right\|^2
+ \tfrac{\beta\eta_x^2 \sigma_x^2}{2n}.
\label{eq:step2}
\end{multline}
 
Applying the identity $-\langle a, b\rangle = \tfrac{1}{2}\|a-b\|^2 - \tfrac{1}{2}\|a\|^2 - \tfrac{1}{2}\|b\|^2$ with $a = \nabla\Phi(\mathbf{x}^t)$ and $b = \tfrac{1}{n}\sum_i \nabla_{\mathbf{x}} f_i(\mathbf{x}_i^t, \mathbf{y}_i^t)$,
\begin{multline}
-\eta_x \mathbb{E}\langle a, b\rangle
= -\tfrac{\eta_x}{2}\mathbb{E}\|\nabla\Phi(\mathbf{x}^t)\|^2
- \tfrac{\eta_x}{2}\mathbb{E}\!\left\|\tfrac{1}{n}\sum_{i=1}^n \nabla_{\mathbf{x}} f_i(\mathbf{x}_i^t, \mathbf{y}_i^t)\right\|^2 
+ \tfrac{\eta_x}{2}\mathbb{E}\!\left\|\nabla\Phi(\mathbf{x}^t) - \tfrac{1}{n}\sum_{i=1}^n \nabla_{\mathbf{x}} f_i(\mathbf{x}_i^t, \mathbf{y}_i^t)\right\|^2.
\label{eq:polar}
\end{multline}
Substituting~\eqref{eq:polar} into~\eqref{eq:step2},

\begin{equation}
\begin{aligned}
\mathbb{E}[\Phi(\mathbf{x}^{t+1}) - \Phi(\mathbf{x}^t)]
&\le -\tfrac{\eta_x}{2}\mathbb{E}\|\nabla\Phi(\mathbf{x}^t)\|^2
- \left(\tfrac{\eta_x}{2} - \tfrac{\beta\eta_x^2}{2}\right)
\mathbb{E}\!\left\|\tfrac{1}{n}\sum_{i=1}^n \nabla_{\mathbf{x}} f_i(\mathbf{x}_i^t, \mathbf{y}_i^t)\right\|^2 \\
&+ \tfrac{\eta_x}{2}\mathbb{E}\!\left\|\nabla\Phi(\mathbf{x}^t) - \tfrac{1}{n}\sum_{i=1}^n \nabla_{\mathbf{x}} f_i(\mathbf{x}_i^t, \mathbf{y}_i^t)\right\|^2
+ \tfrac{\beta\eta_x^2 \sigma_x^2}{2n}.
\label{eq:step3}
\end{aligned}
\end{equation}

\medskip
\noindent{Bounding the gradient-error term.} Decompose the difference through the virtual iterate $\mathbf{x}^t$,
\begin{multline*}
\nabla\Phi(\mathbf{x}^t) - \tfrac{1}{n}\sum_{i=1}^n \nabla_{\mathbf{x}} f_i(\mathbf{x}_i^t, \mathbf{y}_i^t)
= \underbrace{\left[\nabla\Phi(\mathbf{x}^t) - \tfrac{1}{n}\sum_{i=1}^n \nabla_{\mathbf{x}} f_i(\mathbf{x}^t, \mathbf{y}_i^t)\right]}_{E_y \ \text{(inner-tracking error)}}
+ \underbrace{\tfrac{1}{n}\sum_{i=1}^n \left[\nabla_{\mathbf{x}} f_i(\mathbf{x}^t, \mathbf{y}_i^t) - \nabla_{\mathbf{x}} f_i(\mathbf{x}_i^t, \mathbf{y}_i^t)\right]}_{E_x \ \text{(consensus error)}}.
\end{multline*}
By relaxed triangle inequality $\|E_y + E_x\|^2 \le 2\|E_y\|^2 + 2\|E_x\|^2$,
\begin{equation}
\tfrac{\eta_x}{2}\mathbb{E}\!\left\|\nabla\Phi(\mathbf{x}^t) - \tfrac{1}{n}\sum_{i=1}^n \nabla_{\mathbf{x}} f_i(\mathbf{x}_i^t, \mathbf{y}_i^t)\right\|^2
\le \eta_x \mathbb{E}\|E_y\|^2 + \eta_x \mathbb{E}\|E_x\|^2.
\label{eq:split}
\end{equation}
 
\medskip
\noindent{Bounding $\|E_y\|^2$.} By Danskin's theorem, $\nabla\Phi_i(\mathbf{x}) = \nabla_{\mathbf{x}} f_i(\mathbf{x}, \mathbf{y}_i^*(\mathbf{x}))$, so
\[
E_y = \tfrac{1}{n}\sum_{i=1}^n \left[\nabla_{\mathbf{x}} f_i(\mathbf{x}^t, \mathbf{y}_i^*(\mathbf{x}^t)) - \nabla_{\mathbf{x}} f_i(\mathbf{x}^t, \mathbf{y}_i^t)\right].
\]
By Jensen's inequality and $L$-smoothness of $\nabla_{\mathbf{x}} f_i$ in $\mathbf{y}$ (Assumption~\ref{assm:lip-sm}),
\begin{equation}
\begin{aligned}
\|E_y\|^2 \le \tfrac{1}{n}\sum_{i=1}^n \left\|\nabla_{\mathbf{x}} f_i(\mathbf{x}^t, \mathbf{y}_i^*(\mathbf{x}^t)) - \nabla_{\mathbf{x}} f_i(\mathbf{x}^t, \mathbf{y}_i^t)\right\|^2
\le \tfrac{L^2}{n}\sum_{i=1}^n \left\|\mathbf{y}_i^*(\mathbf{x}^t) - \mathbf{y}_i^t\right\|^2.
\label{eq:Ey}
\end{aligned}
\end{equation}
 
\medskip
\noindent{Bounding $\|E_x\|^2$.} By Jensen's inequality and $L$-smoothness of $\nabla_{\mathbf{x}} f_i$ in $\mathbf{x}$,
\begin{equation}
\begin{aligned}
\|E_x\|^2\le \tfrac{1}{n}\sum_{i=1}^n \left\|\nabla_{\mathbf{x}} f_i(\mathbf{x}^t, \mathbf{y}_i^t) - \nabla_{\mathbf{x}} f_i(\mathbf{x}_i^t, \mathbf{y}_i^t)\right\|^2
\le \tfrac{L^2}{n}\sum_{i=1}^n \left\|\mathbf{x}^t - \mathbf{x}_i^t\right\|^2.
\label{eq:Ex}
\end{aligned}
\end{equation}
 
\medskip
The PL condition (Assumption~\ref{assm:pl}) implies the quadratic growth condition (~\cite{karimi2016linear} Appendix~A):
\begin{equation}
\left\|\mathbf{y}_i^t - \mathbf{y}_i^*(\mathbf{x}^t)\right\|^2 \le \tfrac{2}{\mu}\left(\Phi_i(\mathbf{x}^t) - f_i(\mathbf{x}^t, \mathbf{y}_i^t)\right).
\label{eq:QG}
\end{equation}
Substituting~\eqref{eq:QG} into~\eqref{eq:Ey} and averaging,
\begin{equation}
\begin{aligned}
\mathbb{E}\|E_y\|^2 \le  \tfrac{L^2}{n}\sum_{i=1}^n \tfrac{2}{\mu}\mathbb{E}[\Phi_i(\mathbf{x}^t) - f_i(\mathbf{x}^t, \mathbf{y}_i^t)]
= \tfrac{2L^2}{\mu}\mathbb{E}[\Phi(\mathbf{x}^t) - F(\mathbf{x}^t, \mathbf{Y}^t)]
= \tfrac{2L^2}{\mu}\, A_t.
\label{eq:Ey-QG}
\end{aligned}
\end{equation}
 
\medskip
Substituting~\eqref{eq:split},~\eqref{eq:Ex},~\eqref{eq:Ey-QG} into~\eqref{eq:step3} gives desired result
\begin{align*}
\mathbb{E}[\Phi(\mathbf{x}^{t+1}) - \Phi(\mathbf{x}^t)]
&\le -\tfrac{\eta_x}{2}\mathbb{E}\|\nabla\Phi(\mathbf{x}^t)\|^2
- \left(\tfrac{\eta_x}{2} - \tfrac{\beta\eta_x^2}{2}\right)
\mathbb{E}\!\left\|\tfrac{1}{n}\sum_{i=1}^n \nabla_{\mathbf{x}} f_i(\mathbf{x}_i^t, \mathbf{y}_i^t)\right\|^2 \\
&\quad + \tfrac{2\eta_x L^2}{\mu}\, A_t
+ \tfrac{\eta_x L^2}{n}\sum_{i=1}^n \mathbb{E}\left\|\mathbf{x}^t - \mathbf{x}_i^t\right\|^2
+ \tfrac{\beta\eta_x^2 \sigma_x^2}{2n},
\end{align*}

\end{proof}

\begin{lemma}[One-step $\mathbf{y}$-side contraction]
\label{lem:y-contraction}
Under the Assumptions~\ref{assm:lip-sm} to~\ref{assm:pl}, if $\eta_y \le 1/(2L)$, then for each client $i$ the iterates of Algorithm~1 satisfy
\begin{align*}
\mathbb{E}\!\left[\Phi_i(\mathbf{x}^{t+1}) - f_i(\mathbf{x}^{t+1}, \mathbf{y}_i^{t+1})\right]
&\le \left(1 - \tfrac{\mu\eta_y}{2}\right)
\mathbb{E}\!\left[\Phi_i(\mathbf{x}^{t+1}) - f_i(\mathbf{x}^{t+1}, \mathbf{y}_i^t)\right] \\
& + 3\eta_y \sigma_y^2+  6\eta_y L^2\left(
\eta_x^2\, \mathbb{E}\!\left\|\tfrac{1}{n}\sum_{j=1}^n \nabla_{\mathbf{x}} f_j(\mathbf{x}_j^t, \mathbf{y}_j^t)\right\|^2 
+ \tfrac{\eta_x^2 \sigma_x^2}{n}
+ \mathbb{E}\!\left\|\mathbf{x}^t - \mathbf{x}_i^t\right\|^2
\right).
\end{align*}
\end{lemma}
 
\begin{proof}
 Let \[
g_i^t := \nabla_{\mathbf{y}} f_i(\mathbf{x}_i^t, \mathbf{y}_i^t; \xi_i^t),
\quad
\bar g_i^t := \nabla_{\mathbf{y}} f_i(\mathbf{x}_i^t, \mathbf{y}_i^t),
\quad
\hat g_i^t := \nabla_{\mathbf{y}} f_i(\mathbf{x}^{t+1}, \mathbf{y}_i^t).
\]
The stochastic gradient mapping is $\hat G_i^t = (\mathbf{y}_i^{t+1} - \mathbf{y}_i^t)/\eta_y$ (using $g_i^t$ in the projection), and the true gradient mapping used in Assumption~\ref{assm:pl} is $G_i^{\star,t} = \tfrac{1}{\eta_y}[\mathcal{P}_{\mathcal{Y}}(\mathbf{y}_i^t + \eta_y \hat g_i^t) - \mathbf{y}_i^t]$.
 
\medskip
By $L$-smoothness of $f_i$ in $\mathbf{y}$,
\begin{equation}
f_i(\mathbf{x}^{t+1}, \mathbf{y}_i^{t+1})
\ge f_i(\mathbf{x}^{t+1}, \mathbf{y}_i^t)
+ \langle \hat g_i^t,\, \mathbf{y}_i^{t+1} - \mathbf{y}_i^t \rangle
- \tfrac{L}{2}\left\|\mathbf{y}_i^{t+1} - \mathbf{y}_i^t\right\|^2.
\label{eq:y-smooth}
\end{equation}
Rewriting with $\mathbf{y}_i^{t+1} - \mathbf{y}_i^t = \eta_y \hat G_i^t$,
\begin{equation}
f_i(\mathbf{x}^{t+1}, \mathbf{y}_i^{t+1})
\ge f_i(\mathbf{x}^{t+1}, \mathbf{y}_i^t)
+ \eta_y\langle \hat g_i^t,\, \hat G_i^t \rangle
- \tfrac{L\eta_y^2}{2}\|\hat G_i^t\|^2.
\label{eq:y-smooth-2}
\end{equation}
 
\medskip
Since $\mathbf{y}_i^{t+1} = \mathcal{P}_{\mathcal{Y}_i}(\mathbf{y}_i^t + \eta_y g_i^t)$, the first-order optimality of projection gives, for all $\mathbf{y} \in \mathcal{Y}_i$,
\[
\left\langle \mathbf{y}_i^{t+1} - \mathbf{y}_i^t - \eta_y g_i^t,\, \mathbf{y} - \mathbf{y}_i^{t+1} \right\rangle \ge 0.
\]
Taking $\mathbf{y} = \mathbf{y}_i^t$ and simplifying,
\begin{equation}
\|\mathbf{y}_i^{t+1} - \mathbf{y}_i^t\|^2 \le \eta_y \langle g_i^t,\, \mathbf{y}_i^{t+1} - \mathbf{y}_i^t \rangle,
\text{equivalently} 
\langle g_i^t,\, \hat G_i^t \rangle \ge \|\hat G_i^t\|^2.
\label{eq:proj}
\end{equation}
 
\medskip
 Decompose $\hat g_i^t$ into stochastic and $\mathbf{x}$-drift components:
\[
\hat g_i^t = g_i^t + (\bar g_i^t - g_i^t) + (\hat g_i^t - \bar g_i^t),
\]
where the middle term is the stochastic noise at $(\mathbf{x}_i^t, \mathbf{y}_i^t)$ and the last is the $\mathbf{x}$-drift from $\mathbf{x}_i^t$ to $\mathbf{x}^{t+1}$. Then
\[
\eta_y \langle \hat g_i^t, \hat G_i^t \rangle
= \eta_y \langle g_i^t, \hat G_i^t \rangle
+ \eta_y \langle \bar g_i^t - g_i^t, \hat G_i^t \rangle
+ \eta_y \langle \hat g_i^t - \bar g_i^t, \hat G_i^t \rangle.
\]
Using the projection inequality~\eqref{eq:proj} on the first term:
\begin{equation}
\eta_y \langle \hat g_i^t, \hat G_i^t \rangle
\ge \eta_y \|\hat G_i^t\|^2
+ \eta_y \langle \bar g_i^t - g_i^t, \hat G_i^t \rangle
+ \eta_y \langle \hat g_i^t - \bar g_i^t, \hat G_i^t \rangle.
\label{eq:ip-split}
\end{equation}
 
\medskip
\noindent{By Young's inequalities.} Applying $\pm\langle a, b\rangle \le \tfrac{\alpha}{2}\|a\|^2 + \tfrac{1}{2\alpha}\|b\|^2$ with $\alpha = 4$ to both error inner products,
\begin{align}
\eta_y \langle \bar g_i^t - g_i^t, \hat G_i^t \rangle
&\ge -2\eta_y\|\bar g_i^t - g_i^t\|^2 - \tfrac{\eta_y}{8}\|\hat G_i^t\|^2,
\label{eq:Y1}\\
\eta_y \langle \hat g_i^t - \bar g_i^t, \hat G_i^t \rangle
&\ge -2\eta_y\|\hat g_i^t - \bar g_i^t\|^2 - \tfrac{\eta_y}{8}\|\hat G_i^t\|^2.
\label{eq:Y2}
\end{align}
Substituting~\eqref{eq:Y1} and~\eqref{eq:Y2} into~\eqref{eq:ip-split},
\begin{equation}
\eta_y \langle \hat g_i^t, \hat G_i^t \rangle
\ge \tfrac{3\eta_y}{4}\|\hat G_i^t\|^2
- 2\eta_y\|\bar g_i^t - g_i^t\|^2
- 2\eta_y\|\hat g_i^t - \bar g_i^t\|^2.
\label{eq:ip-final}
\end{equation}
Plugging~\eqref{eq:ip-final} into~\eqref{eq:y-smooth-2},
\begin{equation}
\begin{aligned}
f_i(\mathbf{x}^{t+1}, \mathbf{y}_i^{t+1})
&\ge f_i(\mathbf{x}^{t+1}, \mathbf{y}_i^t)
+ \eta_y\left(\tfrac{3}{4} - \tfrac{L\eta_y}{2}\right)\|\hat G_i^t\|^2\\
&- 2\eta_y\|\bar g_i^t - g_i^t\|^2
- 2\eta_y\|\hat g_i^t - \bar g_i^t\|^2.
\label{eq:asc}
\end{aligned}
\end{equation}
Under $\eta_y \le 1/(2L)$, the coefficient satisfies $\tfrac{3}{4} - \tfrac{L\eta_y}{2} \ge \tfrac{1}{2}$, so
\begin{equation}
\begin{aligned}
f_i(\mathbf{x}^{t+1}, \mathbf{y}_i^{t+1})
&\ge f_i(\mathbf{x}^{t+1}, \mathbf{y}_i^t)
+ \tfrac{\eta_y}{2}\|\hat G_i^t\|^2\\
&- 2\eta_y\|\bar g_i^t - g_i^t\|^2
- 2\eta_y\|\hat g_i^t - \bar g_i^t\|^2.
\label{eq:asc-clean}
\end{aligned}
\end{equation}
 
\medskip
By non-expansiveness of the projection,
\[
\begin{aligned}
\|\hat G_i^t - G_i^{\star,t}\|
&= \tfrac{1}{\eta_y}\left\|\mathcal{P}_{\mathcal{Y}}(\mathbf{y}_i^t + \eta_y g_i^t) - \mathcal{P}_{\mathcal{Y}}(\mathbf{y}_i^t + \eta_y \hat g_i^t)\right\|\\
&\le \|g_i^t - \hat g_i^t\|
\le \|g_i^t - \bar g_i^t\| + \|\bar g_i^t - \hat g_i^t\|.
\end{aligned}
\]
Applying the inequality $\|\hat G_i^t\|^2 \ge \tfrac{1}{2}\|G_i^{\star,t}\|^2 - \|\hat G_i^t - G_i^{\star,t}\|^2$ (a consequence of $\|a\|^2 \ge \tfrac{1}{2}\|b\|^2 - \|a-b\|^2$),
\begin{equation}
\|\hat G_i^t\|^2 \ge \tfrac{1}{2}\|G_i^{\star,t}\|^2
- 2\|g_i^t - \bar g_i^t\|^2
- 2\|\bar g_i^t - \hat g_i^t\|^2.
\label{eq:bridge_}
\end{equation}
Substituting~\eqref{eq:bridge_} into~\eqref{eq:asc-clean},
\begin{equation}
\begin{aligned}
f_i(\mathbf{x}^{t+1}, \mathbf{y}_i^{t+1})
\ge f_i(\mathbf{x}^{t+1}, \mathbf{y}_i^t)
+ \tfrac{\eta_y}{4}\|G_i^{\star,t}\|^2
- 3\eta_y\|\bar g_i^t - g_i^t\|^2
- 3\eta_y\|\hat g_i^t - \bar g_i^t\|^2.
\end{aligned}
\label{eq:asc-bridge}
\end{equation}
 
\medskip
 By Assumption~\ref{assm:pl} at $(\mathbf{x}^{t+1}, \mathbf{y}_i^t)$,
\[
\|G_i^{\star,t}\|^2 \ge 2\mu\!\left(\Phi_i(\mathbf{x}^{t+1}) - f_i(\mathbf{x}^{t+1}, \mathbf{y}_i^t)\right).
\]
Substituting into~\eqref{eq:asc-bridge},
\begin{equation*}
f_i(\mathbf{x}^{t+1}, \mathbf{y}_i^{t+1})
\ge f_i(\mathbf{x}^{t+1}, \mathbf{y}_i^t)
+ \tfrac{\mu\eta_y}{2}\!\left(\Phi_i(\mathbf{x}^{t+1}) - f_i(\mathbf{x}^{t+1}, \mathbf{y}_i^t)\right)
- 3\eta_y\|\bar g_i^t - g_i^t\|^2
- 3\eta_y\|\hat g_i^t - \bar g_i^t\|^2.
\end{equation*}
Subtracting both sides from $\Phi_i(\mathbf{x}^{t+1})$ and rearranging,
\begin{equation}
\begin{aligned}
\Phi_i(\mathbf{x}^{t+1}) - f_i(\mathbf{x}^{t+1}, \mathbf{y}_i^{t+1})
&\le \left(1 - \tfrac{\mu\eta_y}{2}\right)\!\left(\Phi_i(\mathbf{x}^{t+1}) - f_i(\mathbf{x}^{t+1}, \mathbf{y}_i^t)\right)\\
&+ 3\eta_y\|\bar g_i^t - g_i^t\|^2
+ 3\eta_y\|\hat g_i^t - \bar g_i^t\|^2.
\label{eq:gap-contract}
\end{aligned}
\end{equation}
 
\medskip
 
By Assumption~\ref{assm:bd_var}, the stochastic noise term is bounded in expectation:
\begin{equation}
\mathbb{E}\|\bar g_i^t - g_i^t\|^2 = \mathbb{E}\|\nabla_{\mathbf{y}} f_i(\mathbf{x}_i^t, \mathbf{y}_i^t) - \nabla_{\mathbf{y}} f_i(\mathbf{x}_i^t, \mathbf{y}_i^t; \xi_i^t)\|^2 \le \sigma_y^2.
\label{eq:noise}
\end{equation}
 
The $\mathbf{x}$-drift term is bounded by $L$-smoothness of $\nabla_{\mathbf{y}} f_i$ in $\mathbf{x}$ (Assumption~\ref{assm:lip-sm}):
\[
\|\hat g_i^t - \bar g_i^t\|^2 = \|\nabla_{\mathbf{y}} f_i(\mathbf{x}^{t+1}, \mathbf{y}_i^t) - \nabla_{\mathbf{y}} f_i(\mathbf{x}_i^t, \mathbf{y}_i^t)\|^2 \le L^2 \|\mathbf{x}^{t+1} - \mathbf{x}_i^t\|^2.
\]
Split $\|\mathbf{x}^{t+1} - \mathbf{x}_i^t\|^2 \le 2\|\mathbf{x}^{t+1} - \mathbf{x}^t\|^2 + 2\|\mathbf{x}^t - \mathbf{x}_i^t\|^2$. For the outer step,
\[
\mathbb{E}\|\mathbf{x}^{t+1} - \mathbf{x}^t\|^2 = \eta_x^2\, \mathbb{E}\!\left\|\tfrac{1}{n}\sum_{j=1}^n \nabla_{\mathbf{x}} f_j(\mathbf{x}_j^t, \mathbf{y}_j^t)\right\|^2 + \tfrac{\eta_x^2 \sigma_x^2}{n}.
\]
Combining,
\begin{equation}
\mathbb{E}\|\hat g_i^t - \bar g_i^t\|^2
\le 2 L^2 \eta_x^2\, \mathbb{E}\!\left\|\tfrac{1}{n}\sum_{j=1}^n \nabla_{\mathbf{x}} f_j(\mathbf{x}_j^t, \mathbf{y}_j^t)\right\|^2
+ \tfrac{2L^2 \eta_x^2 \sigma_x^2}{n}
+ 2L^2\, \mathbb{E}\|\mathbf{x}^t - \mathbf{x}_i^t\|^2.
\label{eq:drift}
\end{equation}
 
\medskip
 Taking expectations in~\eqref{eq:gap-contract} and substituting~\eqref{eq:noise},~\eqref{eq:drift},
\begin{align*}
&\mathbb{E}\!\left[\Phi_i(\mathbf{x}^{t+1}) - f_i(\mathbf{x}^{t+1}, \mathbf{y}_i^{t+1})\right]\\
&\le \left(1 - \tfrac{\mu\eta_y}{2}\right)
\mathbb{E}\!\left[\Phi_i(\mathbf{x}^{t+1}) - f_i(\mathbf{x}^{t+1}, \mathbf{y}_i^t)\right]
+ 3\eta_y \sigma_y^2 \\
&\quad + 6\eta_y L^2\!\left(
\eta_x^2\, \mathbb{E}\!\left\|\tfrac{1}{n}\sum_{j=1}^n \nabla_{\mathbf{x}} f_j(\mathbf{x}_j^t, \mathbf{y}_j^t)\right\|^2
+ \tfrac{\eta_x^2 \sigma_x^2}{n}
+ \mathbb{E}\|\mathbf{x}^t - \mathbf{x}_i^t\|^2
\right),
\end{align*}
which is the claimed inequality.
\end{proof}

\begin{lemma}

Under Assumptions~\ref{assm:lip-sm} to~\ref{assm:pl}, the iterates of Algorithm~1 satisfy
\begin{align*}
\mathbb{E}\!\left[\Phi(\mathbf{x}^{t+1}) - F(\mathbf{x}^{t+1}, \mathbf{Y}^t)\right]
&\le \left(1 + \tfrac{4L^2 \eta_x}{\mu}\right) A_t
+ \tfrac{\eta_x}{2}\, \mathbb{E}\|\nabla\Phi(\mathbf{x}^t)\|^2
+ \tfrac{\eta_x^2 (L+\beta)}{2}\, \mathbb{E}\!\left\|\tfrac{1}{n}\sum_{i=1}^n \nabla_{\mathbf{x}} f_i(\mathbf{x}_i^t, \mathbf{y}_i^t)\right\|^2\\
&+ \tfrac{\eta_x L^2}{n}\sum_{i=1}^n \mathbb{E}\|\mathbf{x}^t - \mathbf{x}_i^t\|^2
+ \tfrac{(L+\beta)\eta_x^2 \sigma_x^2}{2n}.
\end{align*}
Combining this bound with Lemma~\ref{lem:y-contraction} (averaged over clients) yields the coupled recursion
\begin{align*}
A_{t+1} \le \tilde\rho\, A_t
+ \tfrac{\eta_x}{2}\, \mathbb{E}\|\nabla\Phi(\mathbf{x}^t)\|^2
+ \tilde R(t),
\end{align*}
where
\[
\tilde\rho := \left(1 - \tfrac{\mu\eta_y}{2}\right)\!\left(1 + \tfrac{4L^2 \eta_x}{\mu}\right),
\]
and
\begin{align*}
\tilde R(t)
= &\ 3\eta_y \sigma_y^2
+ \eta_x^2\!\left[\tfrac{L+\beta}{2} + 6\eta_y L^2\right] \mathbb{E}\!\left\|\tfrac{1}{n}\sum_{i=1}^n \nabla_{\mathbf{x}} f_i(\mathbf{x}_i^t, \mathbf{y}_i^t)\right\|^2 \\
&+ L^2 (\eta_x + 6\eta_y) \cdot \tfrac{1}{n}\sum_{i=1}^n \mathbb{E}\|\mathbf{x}^t - \mathbf{x}_i^t\|^2
+ \tfrac{\eta_x^2 \sigma_x^2}{n}\!\left[\tfrac{L+\beta}{2} + 6\eta_y L^2\right].
\end{align*}
\label{lem:bridge}
\end{lemma}
 
\begin{proof}
We use the identity
\begin{align*}
\mathbb{E}[\Phi(\mathbf{x}^{t+1}) - F(\mathbf{x}^{t+1}, \mathbf{Y}^t)]
&= \underbrace{\mathbb{E}[\Phi(\mathbf{x}^t) - F(\mathbf{x}^t, \mathbf{Y}^t)]}_{\text{\textcircled{1}} \,=\, A_t}\\ 
&+ \underbrace{\mathbb{E}[\Phi(\mathbf{x}^{t+1}) - \Phi(\mathbf{x}^t)]}_{\text{\textcircled{2}}} \\
&+ \underbrace{\mathbb{E}[F(\mathbf{x}^t, \mathbf{Y}^t) - F(\mathbf{x}^{t+1}, \mathbf{Y}^t)]}_{\text{\textcircled{3}}}.
\end{align*}
Term \textcircled{1} equals $A_t$ by definition. Term \textcircled{2} is bounded by Lemma~\ref{lem:outer-descent}. Term \textcircled{3} is bounded below.
 
\medskip
 Bounding \textcircled{3}. By $L$-smoothness of $F(\cdot, \mathbf{Y}^t)$ in $\mathbf{x}$,
\[
F(\mathbf{x}^{t+1}, \mathbf{Y}^t) \ge F(\mathbf{x}^t, \mathbf{Y}^t) + \langle \nabla_{\mathbf{x}} F(\mathbf{x}^t, \mathbf{Y}^t),\, \mathbf{x}^{t+1} - \mathbf{x}^t \rangle - \tfrac{L}{2}\|\mathbf{x}^{t+1} - \mathbf{x}^t\|^2,
\]
which can be rearranged to
\[
F(\mathbf{x}^t, \mathbf{Y}^t) - F(\mathbf{x}^{t+1}, \mathbf{Y}^t)
\le -\langle \nabla_{\mathbf{x}} F(\mathbf{x}^t, \mathbf{Y}^t),\, \mathbf{x}^{t+1} - \mathbf{x}^t \rangle + \tfrac{L}{2}\|\mathbf{x}^{t+1} - \mathbf{x}^t\|^2.
\]
Substituting $\mathbf{x}^{t+1} - \mathbf{x}^t = -\eta_x \cdot \tfrac{1}{n}\sum_i \nabla_{\mathbf{x}} f_i(\mathbf{x}_i^t, \mathbf{y}_i^t; \xi_i^t)$ and taking expectations, together with the bias-variance decomposition on the squared-norm term,
\begin{equation*}
\text{\textcircled{3}}
\le \eta_x\, \mathbb{E}\!\left\langle \nabla_{\mathbf{x}} F(\mathbf{x}^t, \mathbf{Y}^t),\, \tfrac{1}{n}\sum_{i=1}^n \nabla_{\mathbf{x}} f_i(\mathbf{x}_i^t, \mathbf{y}_i^t)\right\rangle 
+ \tfrac{L\eta_x^2}{2}\, \mathbb{E}\!\left\|\tfrac{1}{n}\sum_{i=1}^n \nabla_{\mathbf{x}} f_i(\mathbf{x}_i^t, \mathbf{y}_i^t)\right\|^2 + \tfrac{L\eta_x^2 \sigma_x^2}{2n}.
\end{equation*}
 
Applying Young's inequality $\langle a, b\rangle \le \tfrac{1}{2}\|a\|^2 + \tfrac{1}{2}\|b\|^2$ on the inner-product term,
\begin{equation}
{\text{\textcircled{3}}}
\le \tfrac{\eta_x}{2}\, \mathbb{E}\|\nabla_{\mathbf{x}} F(\mathbf{x}^t, \mathbf{Y}^t)\|^2
+ \left(\tfrac{\eta_x}{2} + \tfrac{L\eta_x^2}{2}\right) \mathbb{E}\!\left\|\tfrac{1}{n}\sum_{i=1}^n \nabla_{\mathbf{x}} f_i(\mathbf{x}_i^t, \mathbf{y}_i^t)\right\|^2
+ \tfrac{L\eta_x^2 \sigma_x^2}{2n}.
\label{eq:three-a}
\end{equation}
 
\medskip
We split the gradient error term. By relaxed triangle inequality $\|a + b\|^2 \le 2\|a\|^2 + 2\|b\|^2$ with $a = \nabla\Phi(\mathbf{x}^t)$ and $b = \nabla_{\mathbf{x}} F(\mathbf{x}^t, \mathbf{Y}^t) - \nabla\Phi(\mathbf{x}^t)$,
\[
\tfrac{\eta_x}{2}\, \mathbb{E}\|\nabla_{\mathbf{x}} F(\mathbf{x}^t, \mathbf{Y}^t)\|^2
\le \eta_x\, \mathbb{E}\|\nabla\Phi(\mathbf{x}^t)\|^2
+ \eta_x\, \mathbb{E}\|\nabla_{\mathbf{x}} F(\mathbf{x}^t, \mathbf{Y}^t) - \nabla\Phi(\mathbf{x}^t)\|^2.
\]
 
By $L$-smoothness of $f_i$ in $\mathbf{y}$ and Danskin's theorem ($\nabla\Phi_i(\mathbf{x}) = \nabla_{\mathbf{x}} f_i(\mathbf{x}, \mathbf{y}_i^*(\mathbf{x}))$),
\begin{align*}
\mathbb{E}\|\nabla_{\mathbf{x}} F(\mathbf{x}^t, \mathbf{Y}^t) - \nabla\Phi(\mathbf{x}^t)\|^2
&= \mathbb{E}\!\left\|\tfrac{1}{n}\sum_{j=1}^n \!\left(\nabla_{\mathbf{x}} f_j(\mathbf{x}^t, \mathbf{y}_j^t) - \nabla_{\mathbf{x}} f_j(\mathbf{x}^t, \mathbf{y}_j^*(\mathbf{x}^t))\right)\right\|^2 \\
&\le \tfrac{L^2}{n}\sum_{j=1}^n \mathbb{E}\|\mathbf{y}_j^t - \mathbf{y}_j^*(\mathbf{x}^t)\|^2.
\end{align*}
Applying the quadratic-growth consequence of Assumption~\ref{assm:pl}
\begin{align}
\mathbb{E}\|\nabla_{\mathbf{x}} F(\mathbf{x}^t, \mathbf{Y}^t) - \nabla\Phi(\mathbf{x}^t)\|^2 
\le \tfrac{2L^2}{n\mu}\sum_{j=1}^n \mathbb{E}[\Phi_j(\mathbf{x}^t) - f_j(\mathbf{x}^t, \mathbf{y}_j^t)]
= \tfrac{2L^2}{\mu}\, A_t.
\label{eq:F-Phi}
\end{align}
Substituting~\eqref{eq:F-Phi} back into~\eqref{eq:three-a},
\begin{equation}
\begin{aligned}
\text{\textcircled{3}}\le \eta_x\, \mathbb{E}\|\nabla\Phi(\mathbf{x}^t)\|^2
+ \tfrac{2\eta_x L^2}{\mu}\, A_t
+ \left(\tfrac{\eta_x}{2} + \tfrac{L\eta_x^2}{2}\right) \mathbb{E}\!\left\|\tfrac{1}{n}\sum_{i=1}^n \nabla_{\mathbf{x}} f_i(\mathbf{x}_i^t, \mathbf{y}_i^t)\right\|^2
+ \tfrac{L\eta_x^2 \sigma_x^2}{2n}.
\label{eq:three-final}
\end{aligned}
\end{equation}
 
\medskip
 Combining \textcircled{1}+\textcircled{2}+\textcircled{3}. Adding $A_t$, the bound on \textcircled{2} from Lemma~\ref{lem:outer-descent}, and~\eqref{eq:three-final},
 
\emph{Coefficient of $\mathbb{E}\|\nabla\Phi(\mathbf{x}^t)\|^2$:} $-\tfrac{\eta_x}{2} + \eta_x = +\tfrac{\eta_x}{2}$.
 
\emph{Coefficient of $A_t$:} $1 + \tfrac{2\eta_x L^2}{\mu} + \tfrac{2\eta_x L^2}{\mu} = 1 + \tfrac{4\eta_x L^2}{\mu}$.
 
\emph{Coefficient of $\mathbb{E}\|\tfrac{1}{n}\sum_i \nabla_{\mathbf{x}} f_i\|^2$:} $-\left(\tfrac{\eta_x}{2} - \tfrac{\beta\eta_x^2}{2}\right) + \tfrac{\eta_x}{2} + \tfrac{L\eta_x^2}{2} = \tfrac{\eta_x^2 (L+\beta)}{2}$.
 
\emph{Coefficient of $\tfrac{1}{n}\sum_i \mathbb{E}\|\mathbf{x}^t - \mathbf{x}_i^t\|^2$:} $\eta_x L^2$ (from Lemma~\ref{lem:outer-descent}).
 
\emph{Coefficient of $\sigma_x^2/n$:} $\tfrac{\beta\eta_x^2}{2} + \tfrac{L\eta_x^2}{2} = \tfrac{(L+\beta)\eta_x^2}{2}$.
 
Gathering,
\begin{multline}
\mathbb{E}[\Phi(\mathbf{x}^{t+1}) - F(\mathbf{x}^{t+1}, \mathbf{Y}^t)]
\le \left(1 + \tfrac{4L^2 \eta_x}{\mu}\right) A_t
+ \tfrac{\eta_x}{2}\, \mathbb{E}\|\nabla\Phi(\mathbf{x}^t)\|^2 \\
+ \tfrac{\eta_x^2 (L+\beta)}{2}\, \mathbb{E}\!\left\|\tfrac{1}{n}\sum_{i=1}^n \nabla_{\mathbf{x}} f_i(\mathbf{x}_i^t, \mathbf{y}_i^t)\right\|^2
+ \tfrac{\eta_x L^2}{n}\sum_{i=1}^n \mathbb{E}\|\mathbf{x}^t - \mathbf{x}_i^t\|^2
+ \tfrac{(L+\beta)\eta_x^2 \sigma_x^2}{2n},
\label{eq:bridge}
\end{multline}
gives is the first claim.
 
\medskip
Coupling with the $\mathbf{y}$-side contraction. Averaging Lemma~\ref{lem:y-contraction} over clients,
\begin{align}
A_{t+1}
&= \mathbb{E}[\Phi(\mathbf{x}^{t+1}) - F(\mathbf{x}^{t+1}, \mathbf{Y}^{t+1})] \notag\\
&\le \left(1 - \tfrac{\mu\eta_y}{2}\right) \mathbb{E}[\Phi(\mathbf{x}^{t+1}) - F(\mathbf{x}^{t+1}, \mathbf{Y}^t)] + 3\eta_y \sigma_y^2 \notag \\
&+ 6\eta_y L^2\!\left(
\eta_x^2\, \mathbb{E}\!\left\|\tfrac{1}{n}\sum_{i=1}^n \nabla_{\mathbf{x}} f_i(\mathbf{x}_i^t, \mathbf{y}_i^t)\right\|^2+ \tfrac{\eta_x^2 \sigma_x^2}{n}
+ \tfrac{1}{n}\sum_{i=1}^n \mathbb{E}\|\mathbf{x}^t - \mathbf{x}_i^t\|^2
\right).
\label{eq:L2-avg}
\end{align}
Substituting ~\eqref{eq:bridge} into~\eqref{eq:L2-avg} and using $(1 - \tfrac{\mu\eta_y}{2}) \le 1$ on the non-$A_t$ (the $A_t$ coefficient itself remains multiplicative):
 
\emph{Coefficient of $A_t$:} $\left(1 - \tfrac{\mu\eta_y}{2}\right)\!\left(1 + \tfrac{4L^2\eta_x}{\mu}\right) = \tilde\rho$.
 
\emph{Coefficient of $\mathbb{E}\|\nabla\Phi(\mathbf{x}^t)\|^2$:} at most $\tfrac{\eta_x}{2}$.
 
\emph{Coefficient of $\mathbb{E}\|\tfrac{1}{n}\sum_i \nabla_{\mathbf{x}} f_i\|^2$:} at most \\
$\tfrac{\eta_x^2 (L+\beta)}{2} + 6\eta_y L^2 \eta_x^2 = \eta_x^2\!\left[\tfrac{L+\beta}{2} + 6\eta_y L^2\right]$.
 
\emph{Coefficient of $\tfrac{1}{n}\sum_i \mathbb{E}\|\mathbf{x}^t - \mathbf{x}_i^t\|^2$:} at most $\eta_x L^2 + 6\eta_y L^2 = L^2(\eta_x + 6\eta_y)$.
 
\emph{Coefficient of $\sigma_x^2/n$:} at most $\tfrac{(L+\beta)\eta_x^2}{2} + 6\eta_y L^2 \eta_x^2$.
 
\emph{Coefficient of $\sigma_y^2$:} $3\eta_y$.
 
Collecting,
\[
A_{t+1} \le \tilde\rho\, A_t + \tfrac{\eta_x}{2}\, \mathbb{E}\|\nabla\Phi(\mathbf{x}^t)\|^2 + \tilde R(t),
\]
with $\tilde R(t)$ as stated. This proves the second claim.
\end{proof}

We now finally provide the {Convergence Theorem nonconvex-PL} setting.

\begin{theorem}[Non-convex PL Convergence]
\label{thm:app:pl_main}
Suppose Assumptions~\ref{assm:lip-sm} to~\ref{assm:pl} hold, and suppose the step sizes satisfy\\
(a) $\eta_y \le \tfrac{1}{2L}$ , (b) $\tfrac{\eta_x}{\eta_y} \le \tfrac{\mu^2}{16 L^2}$, (c) $10\tau^2 \eta_x^2 L^2 \le \tfrac{1}{2}$ (d) $\eta_x \le \tfrac{1}{\beta}$ \\ 
and (e) $\frac{8\eta_x^2 L^2}{\mu^2\eta_y}\!\left[\frac{L+\beta}{2} + 6\eta_y L^2\right] \le \frac{1}{2}\big(1 - \beta\eta_x\big)$. (a),(b),(c),(d) arises due to smoothness, two-timescale, consenus-stability, outer-smoothness respectively. With $T \ge T_0 = \max\{4L^2,\frac{25\mu^8}{409L^4}, \frac{\mu^2}{256}\} $.  The iterates of Algorithm~1 satisfy
\begin{align*}
\frac{1}{T}\sum_{t=0}^{T-1} \mathbb{E}\|\nabla\Phi(\mathbf{x}^t)\|^2 
&\le \frac{4\Delta_\Phi}{\eta_x T} + \frac{32 L^2 A_0}{\mu^2 \eta_y T} + \frac{96 L^2 \sigma_y^2}{\mu^2} + \frac{2\beta \eta_x \sigma_x^2}{n} 
+ \frac{32 L^2 \eta_x^2 \sigma_x^2}{\mu^2 \eta_y n}\!\left(\tfrac{L+\beta}{2} + 6\eta_y L^2\right) \\
&\quad + \left(4 L^2 + \tfrac{32 L^4 (\eta_x + 6\eta_y)}{\mu^2 \eta_y}\right)\cdot 10\tau \eta_x^2\!\left[\sigma_x^2\!\left(1 + \tfrac{1}{n}\right) + \zeta_x\right].
\end{align*}
Choosing  $\eta_y = \Theta(1/\sqrt{T})$, $\eta_x = \tfrac{\mu^2}{16L^2}\eta_y
= \Theta(1/\sqrt{T})$, we obtain the rate
\begin{align*}
\frac{1}{T}\sum_{t=0}^{T-1} \mathbb{E}\|\nabla\Phi(\mathbf{x}^t)\|^2 &\le  O\!\Big(\tfrac{L^2(\Delta_\Phi + A_0) +\mu\sigma_x^2}{\mu^2\sqrt{T}}\Big)
      + {O\!\Big(\tfrac{L^2\sigma_y^2}{\mu^2}\Big)}
      + O\!\Big(\tfrac{L^2\mu^2(\sigma_x^2+\zeta_x)}{T^{3/4}}\Big) + O(\frac{1}{T})
\end{align*}
\end{theorem}
 
\begin{proof}
 From Lemma~\ref{lem:outer-descent}, telescoping $t = 0, \ldots, T-1$,
\[
\begin{aligned}
\mathbb{E}[\Phi(\mathbf{x}^T) - \Phi(\mathbf{x}^0)]
\le -\tfrac{\eta_x}{2} T G_T
+ \tfrac{2\eta_x L^2}{\mu}\sum_{t=0}^{T-1} A_t
+ \eta_x L^2\!\sum_{t=0}^{T-1} \tfrac{1}{n}\sum_{i=1}^n \mathbb{E}\|\mathbf{x}^t - \mathbf{x}_i^t\|^2
+ \tfrac{\beta \eta_x^2 \sigma_x^2 T}{2n},
\end{aligned}
\]
where we have dropped the (non-positive under $\eta_x \le 1/\beta$) $\|\tfrac{1}{n}\sum \nabla_{\mathbf{x}} f_i\|^2$ term. Rearranging with $\Phi(\mathbf{x}^T) \ge \Phi^*$,
\begin{equation}
\tfrac{\eta_x}{2} G_T
\le \tfrac{\Delta_\Phi}{T}
+ \tfrac{2\eta_x L^2}{\mu}\cdot\tfrac{1}{T}\sum_{t=0}^{T-1} A_t
+ \eta_x L^2\cdot\tfrac{1}{T}\sum_t \tfrac{1}{n}\sum_i \mathbb{E}\|\mathbf{x}^t - \mathbf{x}_i^t\|^2
+ \tfrac{\beta \eta_x^2 \sigma_x^2}{2n}.
\label{eq:outer-sum_pl}
\end{equation}
 
\medskip

Summing the coupled recursion. From Lemma~\ref{lem:bridge},
\[
A_{t+1} \le \tilde\rho A_t + \tfrac{\eta_x}{2}\mathbb{E}\|\nabla\Phi(\mathbf{x}^t)\|^2 + \tilde R(t),
\qquad \tilde\rho = \left(1 - \tfrac{\mu\eta_y}{2}\right)\!\left(1 + \tfrac{4L^2\eta_x}{\mu}\right).
\]

Using the two-timescale condition $\eta_x/\eta_y \le \mu^2/(16L^2)$, which gives
$\tfrac{4L^2\eta_x}{\mu} \le \tfrac{\mu\eta_y}{4}$, and dropping the non-positive
cross term $-\tfrac{\mu\eta_y}{2}\cdot\tfrac{4L^2\eta_x}{\mu}$,
\begin{equation}
\tilde\rho
= \Big(1 - \tfrac{\mu\eta_y}{2}\Big)\Big(1 + \tfrac{4L^2\eta_x}{\mu}\Big)
\le 1 + \tfrac{4L^2\eta_x}{\mu} - \tfrac{\mu\eta_y}{2}
\le 1 + \tfrac{\mu\eta_y}{4} - \tfrac{\mu\eta_y}{2}
= 1 - \tfrac{\mu\eta_y}{4} < 1.
\end{equation}

Unrolling and averaging,
\[
\tfrac{1}{T}\sum_{t=0}^{T-1} A_t
\le \tfrac{A_0}{(1-\tilde\rho)T}
+ \tfrac{\eta_x/2}{1-\tilde\rho} G_T
+ \tfrac{1}{(1-\tilde\rho)T}\sum_{t=0}^{T-1} \tilde R(t).
\]
Using $1 - \tilde\rho \ge \mu\eta_y/4$,
\begin{equation}
\tfrac{1}{T}\sum_{t=0}^{T-1} A_t
\le \tfrac{4 A_0}{\mu\eta_y T}
+ \tfrac{2\eta_x}{\mu\eta_y} G_T
+ \tfrac{4}{\mu\eta_y T}\sum_{t=0}^{T-1} \tilde R(t).
\label{eq:gap-avg}
\end{equation}
 
\medskip
 
Substituting~\eqref{eq:gap-avg} into~\eqref{eq:outer-sum_pl} and solving for $G_T$.
 
Substituting and collecting the $G_T$ terms on the left,
\begin{equation}
\begin{aligned}
\left[\tfrac{\eta_x}{2} - \tfrac{4 L^2 \eta_x^2}{\mu^2 \eta_y}\right] G_T
&\le \tfrac{\Delta_\Phi}{T}
+ \tfrac{8 L^2 \eta_x A_0}{\mu^2 \eta_y T}
+ \tfrac{8 L^2 \eta_x}{\mu^2 \eta_y T}\sum_t \tilde R(t)\\
&+ \eta_x L^2\cdot\tfrac{1}{T}\sum_t \tfrac{1}{n}\sum_i \mathbb{E}\|\mathbf{x}^t - \mathbf{x}_i^t\|^2
+ \tfrac{\beta \eta_x^2 \sigma_x^2}{2n}.
\label{eq:GT-collect}
\end{aligned}
\end{equation}
 
By the two-timescale condition ( $\eta_x/\eta_y \le \mu^2/(16 L^2)$), $\tfrac{4 L^2 \eta_x^2}{\mu^2 \eta_y} \le \tfrac{\eta_x}{4}$, so the coefficient on $G_T$ is at least $\eta_x/4$:
\begin{equation}
\begin{aligned}
\tfrac{\eta_x}{4} G_T
\le \tfrac{\Delta_\Phi}{T}
+ \tfrac{8 L^2 \eta_x A_0}{\mu^2 \eta_y T}
+ \tfrac{8 L^2 \eta_x}{\mu^2 \eta_y T}\sum_t \tilde R(t)
+ \eta_x L^2\cdot\tfrac{1}{T}\sum_t \tfrac{1}{n}\sum_i \mathbb{E}\|\mathbf{x}^t - \mathbf{x}_i^t\|^2
+ \tfrac{\beta \eta_x^2 \sigma_x^2}{2n}.
\label{eq:GT-simplified}
\end{aligned}
\end{equation}
 
\medskip
Substituting $\tilde R(t)$ and the consensus bound~\ref{lm:sync_error}. From Lemma~\ref{lem:bridge},
\begin{multline*}
\tilde R(t)
= 3\eta_y \sigma_y^2
+ \eta_x^2\!\left[\tfrac{L+\beta}{2} + 6\eta_y L^2\right]\mathbb{E}\!\left\|\tfrac{1}{n}\sum_i \nabla_{\mathbf{x}} f_i\right\|^2
+ L^2(\eta_x + 6\eta_y)\cdot \tfrac{1}{n}\sum_i \mathbb{E}\|\mathbf{x}^t - \mathbf{x}_i^t\|^2
+ \tfrac{\eta_x^2 \sigma_x^2}{n}\!\left[\tfrac{L+\beta}{2} + 6\eta_y L^2\right].
\end{multline*}
The $\|\tfrac{1}{n}\sum \nabla_{\mathbf{x}} f_i\|^2$ term is absorbed by the (previously dropped) non-positive coefficient from~\eqref{eq:outer-sum_pl} under the stepsize conditions.
 
Summing and expanding the consensus terms,
\[
\begin{aligned}
\tfrac{8 L^2 \eta_x}{\mu^2 \eta_y T}\sum_t \tilde R(t)\le \tfrac{24 \eta_xL^2 \sigma_y^2}{\mu^2}
+ \tfrac{8 L^2 \eta_x^3 \sigma_x^2}{\mu^2 \eta_y n}\!\left[\tfrac{L+\beta}{2} + 6\eta_y L^2\right]
+ \tfrac{8 L^4 (\eta_x + 6\eta_y)}{\mu^2 \eta_y}\cdot 10\tau \eta_x^3\!\left[\sigma_x^2\!\left(1 + \tfrac{1}{n}\right) + \zeta_x\right],
\end{aligned}
\]
and the consensus term in~\eqref{eq:GT-simplified} contributes an additional
\[
\eta_x L^2 \cdot 10\tau \eta_x^2\!\left[\sigma_x^2\!\left(1 + \tfrac{1}{n}\right) + \zeta_x\right].
\]
 
\medskip
 Multiplying~\eqref{eq:GT-simplified} by $4/\eta_x$ and collecting,
\begin{align*}
G_T
&\le \tfrac{4\Delta_\Phi}{\eta_x T}
+ \tfrac{32 L^2 A_0}{\mu^2 \eta_y T}
+ \tfrac{96 L^2 \sigma_y^2}{\mu^2}
+ \tfrac{2\beta \eta_x \sigma_x^2}{n}
+ \tfrac{32 L^2 \eta_x^2 \sigma_x^2}{\mu^2 \eta_y n}\!\left[\tfrac{L+\beta}{2} + 6\eta_y L^2\right] \\
&\quad + \left(4 L^2 + \tfrac{32 L^4 (\eta_x + 6\eta_y)}{\mu^2 \eta_y}\right) \cdot 10\tau \eta_x^2\!\left[\sigma_x^2\!\left(1 + \tfrac{1}{n}\right) + \zeta_x\right],
\end{align*}
which is the bound stated in the theorem.

\medskip
Choose $\eta_y = \Theta(1/\sqrt{T})$, $\eta_x = \tfrac{\mu^2}{16L^2}\eta_y
= \Theta(1/\sqrt{T})$ (tight on the two-timescale condition~(b)), and
$\tau = O(T^{1/4})$.

computing $T_0$.
Under $\eta_y = 1/\sqrt{T}$, $\eta_x = \tfrac{\mu^2}{16L^2}\eta_y$, and
$\tau = T^{1/4}$, condition (b) holds with equality for all $T$, and the
remaining conditions give thresholds
\[
\text{(a)}\;\; T \ge 4L^2,\qquad
\text{(c)}\;\; T \ge \tfrac{25\mu^8}{4096L^4},\qquad
\text{(d)}\;\; T \ge \tfrac{\mu^2}{256},
\]
with (e) satisfied for any $\eta_x = \Theta(1/\sqrt{T})$. Since $\mu \le L$,
the thresholds from (c) and (d) are dominated by (a), so
$T_0 = \max\{4L^2,\frac{25\mu^8}{409L^4}, \frac{\mu^2}{256}\} $.

All conditions (a)to (e) hold for $T \ge T_0$,
and with $\beta = L(1+\kappa) = O(L^2/\mu)$ each term reduces as follows.
\begin{itemize}
\item Init in $\Phi$: $\tfrac{4\Delta_\Phi}{\eta_x T}
  = O\!\big(L^2\Delta_\Phi/(\mu^2\sqrt{T})\big)$  =  $O(1/\sqrt{T})$.
\item Init in $A_0$: $\tfrac{32L^2 A_0}{\mu^2\eta_y T}
  = O\!\big(L^2 A_0/(\mu^2\sqrt{T})\big)$ = $O(1/\sqrt{T})$.
\item Noise floor: $\tfrac{96L^2\sigma_y^2}{\mu^2}
  = O(L^2\sigma_y^2/\mu^2)$, constant in $T$; does not vanish.
\item Outer stochastic noise: $\tfrac{2\beta\eta_x\sigma_x^2}{n}
  = O\!\big(\mu\sigma_x^2/(n\sqrt{T})\big)$ =  $O(1/(n\sqrt{T}))$.
\item Inner-variance: $\tfrac{32L^2\eta_x^2\sigma_x^2}{\mu^2\eta_y n}
  \big[\tfrac{L+\beta}{2}+6\eta_y L^2\big]
  = O\!\big(\mu\sigma_x^2/(n\sqrt{T})\big) + O\!\big(\mu^2\sigma_x^2/(nT)\big)$, the two bracket pieces give $O(1/(n\sqrt{T}))$ and $O(1/(nT))$.
\item Consensus: $\big(4L^2+\tfrac{32L^4(\eta_x+24\eta_y)}{\mu^2\eta_y}\big)
  10\tau\eta_x^2[\sigma_x^2(1{+}\tfrac1n)+\zeta_x]
  = O\!\big(L^2\mu^2(\sigma_x^2+\zeta_x)/T^{3/4}\big)$  = $O(1/T^{3/4})$.
\end{itemize}
The $O(1/\sqrt{T})$ terms (initialization in $\Phi$ and $A_0$) vanish; the
remaining stochastic and consensus terms vanish faster. The one term that
does \emph{not} vanish is the noise floor $O(L^2\sigma_y^2/\mu^2)$, which is
independent of the stepsizes. Collecting,
\[
  G_T = O\!\Big(\tfrac{L^2(\Delta_\Phi + A_0) +\mu\sigma_x^2}{\mu^2\sqrt{T}}\Big)
      + {O\!\Big(\tfrac{L^2\sigma_y^2}{\mu^2}\Big)}
      + O\!\Big(\tfrac{L^2\mu^2(\sigma_x^2+\zeta_x)}{T^{3/4}}\Big) + O(\frac{1}{T}).
\]
Thus plain projected SGDA converges only to a neighborhood of radius
$O(L^2\sigma_y^2/\mu^2)$.
schedule.

\end{proof}

\subsection{Convergence Proof of VR-SGDA}
 
\begin{lemma}[SVRG variance bound]
\label{lem:svrg-var}
Under Assumption~\ref{assump:svrg-smooth}, the SVRG-corrected gradient $v_i^t$ from Algorithm~\ref{alg:merged_sgda} satisfies
\[
\mathbb{E}\|v_i^t - \bar v_i^t\|^2
\le 4L^2\!\left(\mathbb{E}\|\mathbf{x}^t - \mathbf{x}_i^t\|^2 + \mathbb{E}\|\mathbf{x}^t - \bar{\mathbf{x}}\|^2\right)
+ 2L^2\, \mathbb{E}\|\mathbf{y}_i^t - \bar{\mathbf{y}}_i\|^2,
\]
where $\bar v_i^t := \nabla_{\mathbf{y}} f_i(\mathbf{x}_i^t, \mathbf{y}_i^t)$ is the true gradient at the current iterate, $\mathbf{x}^t = \tfrac{1}{n}\sum_i \mathbf{x}_i^t$ is the virtual-average iterate, and $\bar{\mathbf{x}} = \mathbf{x}^s$ is the snapshot at the last synchronization time.
\end{lemma}
 
\begin{proof}
Let $W(\xi) := \nabla_{\mathbf{y}} f_i(\mathbf{x}_i^t, \mathbf{y}_i^t; \xi) - \nabla_{\mathbf{y}} f_i(\bar{\mathbf{x}}, \bar{\mathbf{y}}_i; \xi)$. Then
\[
v_i^t - \bar v_i^t = W(\xi_i^t) - \mathbb{E}_\xi[W(\xi)],
\]
and
\[
\mathbb{E}\|v_i^t - \bar v_i^t\|^2 = \text{Var}_{\xi_i^t}(W(\xi_i^t)) \le \mathbb{E}\|W(\xi_i^t)\|^2.
\]
 
\medskip
\noindent{Bounding $\|W(\xi)\|^2$ pointwise.} By Assumption~\ref{assump:svrg-smooth},
\[
\|W(\xi)\|^2 \le \big(L\|\mathbf{x}_i^t - \bar{\mathbf{x}}\| + L\|\mathbf{y}_i^t - \bar{\mathbf{y}}_i\|\big)^2
\le 2L^2\|\mathbf{x}_i^t - \bar{\mathbf{x}}\|^2 + 2L^2\|\mathbf{y}_i^t - \bar{\mathbf{y}}_i\|^2,
\]
where the second inequality uses $(a+b)^2 \le 2(a^2 + b^2)$.
 
\medskip
\noindent{Decomposing $\|\mathbf{x}_i^t - \bar{\mathbf{x}}\|^2$,} The quantity $\bar{\mathbf{x}} = \mathbf{x}^s$ is the virtual-average iterate at the last synchronization time; it is \emph{not} the same as either the current local iterate $\mathbf{x}_i^t$ or the current virtual average $\mathbf{x}^t$. By the triangle inequality,
\[
\|\mathbf{x}_i^t - \bar{\mathbf{x}}\|^2 = \|(\mathbf{x}_i^t - \mathbf{x}^t) + (\mathbf{x}^t - \bar{\mathbf{x}})\|^2 \le 2\|\mathbf{x}^t - \mathbf{x}_i^t\|^2 + 2\|\mathbf{x}^t - \bar{\mathbf{x}}\|^2.
\]
The first term is the consensus error (bounded by the draft's Lemma on client drift); the second is the \emph{virtual-average drift-from-snapshot}, bounded in Lemma~\ref{lem:xbar-drift} below.
 
\medskip
 Combining above and taking expectations:
\begin{align*}
\mathbb{E}\|v_i^t - \bar v_i^t\|^2
&\le 2L^2\big(2\mathbb{E}\|\mathbf{x}^t - \mathbf{x}_i^t\|^2 + 2\mathbb{E}\|\mathbf{x}^t - \bar{\mathbf{x}}\|^2\big) + 2L^2\mathbb{E}\|\mathbf{y}_i^t - \bar{\mathbf{y}}_i\|^2 \\
&= 4L^2\!\left(\mathbb{E}\|\mathbf{x}^t - \mathbf{x}_i^t\|^2 + \mathbb{E}\|\mathbf{x}^t - \bar{\mathbf{x}}\|^2\right) + 2L^2\mathbb{E}\|\mathbf{y}_i^t - \bar{\mathbf{y}}_i\|^2,
\end{align*}
which is the claimed bound.
\end{proof}

\begin{lemma}[Virtual-average drift from snapshot]
\label{lem:xbar-drift}
For $t \in [s, s+\tau)$ with $s = \lfloor t/\tau\rfloor \cdot \tau$,
\[
\mathbb{E}\|\mathbf{x}^t - \bar{\mathbf{x}}\|^2 \le \tau\eta_x^2 \sum_{j=s}^{t-1} \mathbb{E}\!\left\|\tfrac{1}{n}\sum_k \nabla_{\mathbf{x}} f_k(\mathbf{x}_k^j, \mathbf{y}_k^j)\right\|^2 + \tfrac{\tau\eta_x^2 \sigma_x^2}{n} \cdot (t - s).
\]
In particular, if the outer gradient magnitude is uniformly bounded by $G_x^2$ (as ensured under gradient-mapping PL and bounded gap),
\[
\max_{t \in [s, s+\tau)} \mathbb{E}\|\mathbf{x}^t - \bar{\mathbf{x}}\|^2 \le \tau^2 \eta_x^2 G_x^2 + \tfrac{\tau^2 \eta_x^2 \sigma_x^2}{n}.
\]
\end{lemma}
 
\begin{proof}
Since $\bar{\mathbf{x}} = \mathbf{x}^s$,
\[
\mathbf{x}^t - \bar{\mathbf{x}} = \sum_{j=s}^{t-1}(\mathbf{x}^{j+1} - \mathbf{x}^j) = -\eta_x \sum_{j=s}^{t-1} \tfrac{1}{n}\sum_k \nabla_{\mathbf{x}} f_k(\mathbf{x}_k^j, \mathbf{y}_k^j; \xi_k^j).
\]
By Cauchy--Schwarz (over $j$),
\[
\|\mathbf{x}^t - \bar{\mathbf{x}}\|^2 \le (t - s)\eta_x^2 \sum_{j=s}^{t-1}\left\|\tfrac{1}{n}\sum_k \nabla_{\mathbf{x}} f_k(\mathbf{x}_k^j, \mathbf{y}_k^j; \zeta_k^j)\right\|^2.
\]
Taking expectations and using the bias--variance decomposition $\mathbb{E}\|\tfrac{1}{n}\sum_k \nabla_{\mathbf{x}} f_k(\cdot; \zeta_k^j)\|^2 = \mathbb{E}\|\tfrac{1}{n}\sum_k \nabla_{\mathbf{x}} f_k(\cdot)\|^2 + \tfrac{\sigma_x^2}{n}$, we obtain the first claim. Bounding the true-gradient term by $G_x^2$ and using $(t-s) \le \tau$ gives the second.
\end{proof}
  
\begin{lemma}[$\mathbf{y}$-drift from snapshot within a synchronization block]
\label{lem:y-drift}
Suppose the block-length constraint $8L^2 \tau^2 \eta_y^2 \le 1/2$ holds. Under Assumptions~\ref{assm:lip-sm}--\ref{assm:pl} and $\eta_y \le 1/(2L)$, and defining $\gamma_y^t := \tfrac{1}{n}\sum_i \mathbb{E}\|\mathbf{y}_i^t - \bar{\mathbf{y}}_i\|^2$, $\gamma^t := \tfrac{1}{n}\sum_i \mathbb{E}\|\mathbf{x}^t - \mathbf{x}_i^t\|^2$, and letting $G_y^2$ be a uniform bound on the true inner gradient norm $\|\bar v_i^j\|^2$, we have
\[
\max_{t \in [s, s+\tau)} \gamma_y^t \le 4\tau^2 \eta_y^2 G_y^2 + 16 L^2 \tau^2 \eta_y^2 \max_{j \in [s, s+\tau)} \gamma^j + 16 L^2 \tau^4 \eta_x^2 \eta_y^2\!\left(G_x^2 + \tfrac{\sigma_x^2}{n}\right).
\]
\end{lemma}
 
\begin{proof}
The proof proceeds in following steps.
 
\medskip
\noindent\textit{One-step decomposition.} 
Since $\bar{\mathbf{y}}_i = \mathbf{y}_i^s$,
\begin{equation}
\mathbf{y}_i^t - \bar{\mathbf{y}}_i = \sum_{j=s}^{t-1}(\mathbf{y}_i^{j+1} - \mathbf{y}_i^j).
\label{eq:y-tel-1}
\end{equation}
By non-expansiveness of $\mathcal{P}_{\mathcal{Y}}$, $\|\mathbf{y}_i^{j+1} - \mathbf{y}_i^j\| \le \eta_y \|v_i^j\|$. Applying Cauchy--Schwarz over $j$:
\begin{equation}
\|\mathbf{y}_i^t - \bar{\mathbf{y}}_i\|^2 \le (t - s)\eta_y^2 \sum_{j=s}^{t-1}\|v_i^j\|^2 \le \tau\eta_y^2 \sum_{j=s}^{t-1}\|v_i^j\|^2.
\label{eq:y-tel-2}
\end{equation}
 
\medskip
\noindent{We Bound $\mathbb{E}\|v_i^j\|^2$} by Splitting via Young's inequality,
\[
\|v_i^j\|^2 \le 2\|\bar v_i^j\|^2 + 2\|v_i^j - \bar v_i^j\|^2.
\]
The first term is bounded uniformly: $\|\bar v_i^j\|^2 \le G_y^2$. The second term is bounded by Lemma~\ref{lem:svrg-var} in expectation:
\[
\mathbb{E}\|v_i^j - \bar v_i^j\|^2 \le 4L^2\!\left(\mathbb{E}\|\mathbf{x}^j - \mathbf{x}_i^j\|^2 + \mathbb{E}\|\mathbf{x}^j - \bar{\mathbf{x}}\|^2\right) + 2L^2\mathbb{E}\|\mathbf{y}_i^j - \bar{\mathbf{y}}_i\|^2.
\]
Combining,
\begin{equation}
\mathbb{E}\|v_i^j\|^2 \le 2G_y^2 + 8L^2\!\left(\mathbb{E}\|\mathbf{x}^j - \mathbf{x}_i^j\|^2 + \mathbb{E}\|\mathbf{x}^j - \bar{\mathbf{x}}\|^2\right) + 4L^2\mathbb{E}\|\mathbf{y}_i^j - \bar{\mathbf{y}}_i\|^2.
\label{eq:v-bound}
\end{equation}
 
\medskip
\noindent\textit{Averaging over clients.} 

Taking expectations in~\eqref{eq:y-tel-2} and averaging over $i$,
\[
\gamma_y^t \le \tau\eta_y^2 \sum_{j=s}^{t-1}\!\left[2G_y^2 + 8L^2\gamma^j + 8L^2\mathbb{E}\|\mathbf{x}^j - \bar{\mathbf{x}}\|^2 + 4L^2\gamma_y^j\right].
\]
Here we used $\tfrac{1}{n}\sum_i \mathbb{E}\|\mathbf{x}^j - \mathbf{x}_i^j\|^2 = \gamma^j$ and $\mathbb{E}\|\mathbf{x}^j - \bar{\mathbf{x}}\|^2$ is common to all clients.
 
\medskip
\noindent\textit{ Taking the maximum over the block.}\\
Define $\gamma_{y,\max} := \max_{t \in [s, s+\tau)} \gamma_y^t$, $\gamma_{\max} := \max_{j \in [s, s+\tau)} \gamma^j$, and \\
$\Delta_{\bar x,\max} := \max_{j \in [s, s+\tau)} \mathbb{E}\|\mathbf{x}^j - \bar{\mathbf{x}}\|^2$. Bounding each summand by the maximum and using $(t-s) \le \tau$:
\[
\gamma_{y,\max} \le \tau \eta_y^2 \cdot \tau \cdot \left[2G_y^2 + 8L^2\gamma_{\max} + 8L^2 \Delta_{\bar x,\max} + 4L^2\gamma_{y,\max}\right]
\]
\[
= 2\tau^2\eta_y^2 G_y^2 + 8L^2\tau^2\eta_y^2\gamma_{\max} + 8L^2\tau^2\eta_y^2\Delta_{\bar x,\max} + 4L^2\tau^2\eta_y^2\gamma_{y,\max}.
\]
 
\medskip
\noindent\textit{ Solving the  inequality.} Rearranging,
\[
\gamma_{y,\max}\big(1 - 4L^2\tau^2\eta_y^2\big) \le 2\tau^2\eta_y^2 G_y^2 + 8L^2\tau^2\eta_y^2(\gamma_{\max} + \Delta_{\bar x,\max}).
\]
Under the block-length constraint $8L^2 \tau^2 \eta_y^2 \le 1/2$ (which implies $4L^2\tau^2\eta_y^2 \le 1/4$, hence $1/(1 - 4L^2\tau^2\eta_y^2) \le 4/3 \le 2$), we obtain
\begin{equation}
\gamma_{y,\max} \le 4\tau^2\eta_y^2 G_y^2 + 16L^2\tau^2\eta_y^2(\gamma_{\max} + \Delta_{\bar x,\max}).
\label{eq:y-drift-final}
\end{equation}
Substituting $\Delta_{\bar x,\max} \le \tau^2\eta_x^2(G_x^2 + \sigma_x^2/n)$ from Lemma~\ref{lem:xbar-drift} yields the claimed bound.
\end{proof}
 
\medskip
The recursion $\gamma_{y,\max} \le \text{const} + \text{const}\cdot \gamma_{y,\max}$ closes only when the self-referential coefficient is strictly less than 1. The block-length constraint $8L^2\tau^2\eta_y^2 \le 1/2$ ensures this. Under the stepsize regime $\eta_y = \Theta(1/\sqrt{T})$, $\tau = O(T^{1/4}/\sqrt{n})$, the constraint becomes $8L^2 \cdot T^{1/2}/n \cdot 1/T = 8L^2/(n\sqrt{T}) \le 1/2$, which holds for $T \ge 256 L^4/n^2$, a mild condition.

%\section*{Updated Inner Contraction Lemma (Lemma 2b)}
 
\begin{lemma}[One-step $\mathbf{y}$-side gap contraction under SVRG]
\label{lem:y-contraction-svrg}
Under Assumptions~1 to 4 and Assumption on component-wise smoothness (Assumption~1'), if $\eta_y \le 1/(2L)$, then Algorithm~2 satisfies
\begin{align*}
\mathbb{E}\!\left[\Phi_i(\mathbf{x}^{t+1}) - f_i(\mathbf{x}^{t+1}, \mathbf{y}_i^{t+1})\right]
&\le \left(1 - \tfrac{\mu\eta_y}{2}\right)\mathbb{E}\!\left[\Phi_i(\mathbf{x}^{t+1}) - f_i(\mathbf{x}^{t+1}, \mathbf{y}_i^t)\right] + 18\eta_y L^2\, \mathbb{E}\|\mathbf{x}^t - \mathbf{x}_i^t\|^2 \\
&+ 12\eta_y L^2\, \mathbb{E}\|\mathbf{x}^t - \bar{\mathbf{x}}\|^2
+ 6\eta_y L^2\, \mathbb{E}\|\mathbf{y}_i^t - \bar{\mathbf{y}}_i\|^2 \\
&\quad + 6\eta_y L^2\eta_x^2\, \mathbb{E}\!\left\|\tfrac{1}{n}\sum_{j=1}^n \nabla_{\mathbf{x}} f_j(\mathbf{x}_j^t, \mathbf{y}_j^t)\right\|^2
+ \tfrac{6\eta_y L^2\eta_x^2\sigma_x^2}{n}.
\end{align*}
\end{lemma}
 
\begin{proof}
The derivation follows the Lemma.~\ref{lem:y-contraction}, with the stochastic gradient $g_i^t$ replaced by the SVRG-corrected estimator $v_i^t$. Two substantive differences: (i) the projection variational inequality now involves $v_i^t$ instead of $g_i^t$; (ii) the ``noise term'' $\|\bar g_i^t - v_i^t\|^2$ is now bounded by Lemma~\ref{lem:y-drift} instead of by $\sigma_y^2$.
  Let \[
g_i^t := \nabla_{\mathbf{y}} f_i(\mathbf{x}_i^t, \mathbf{y}_i^t; \xi_i^t),
\quad
\bar g_i^t := \nabla_{\mathbf{y}} f_i(\mathbf{x}_i^t, \mathbf{y}_i^t),
\quad
\hat g_i^t := \nabla_{\mathbf{y}} f_i(\mathbf{x}^{t+1}, \mathbf{y}_i^t).
\]
\medskip
\noindent\textit{Smoothness at $\mathbf{x}^{t+1}$.} \\
By $L$-smoothness of $f_i$ in $\mathbf{y}$,
\begin{equation}
f_i(\mathbf{x}^{t+1}, \mathbf{y}_i^{t+1})
\ge f_i(\mathbf{x}^{t+1}, \mathbf{y}_i^t)
+ \langle \hat g_i^t, \mathbf{y}_i^{t+1} - \mathbf{y}_i^t\rangle
- \tfrac{L}{2}\|\mathbf{y}_i^{t+1} - \mathbf{y}_i^t\|^2.
\label{eq:smooth-svrg}
\end{equation}
Substituting $\mathbf{y}_i^{t+1} - \mathbf{y}_i^t = \eta_y \hat G_i^t$,
\begin{equation}
f_i(\mathbf{x}^{t+1}, \mathbf{y}_i^{t+1})
\ge f_i(\mathbf{x}^{t+1}, \mathbf{y}_i^t)
+ \eta_y\langle \hat g_i^t, \hat G_i^t\rangle
- \tfrac{L\eta_y^2}{2}\|\hat G_i^t\|^2.
\label{eq:smooth-svrg-2}
\end{equation}
 
\medskip

Since $\mathbf{y}_i^{t+1} = \mathcal{P}_{\mathcal{Y}}(\mathbf{y}_i^t + \eta_y v_i^t)$, first-order optimality of the projection gives
\[
\langle \mathbf{y}_i^{t+1} - \mathbf{y}_i^t - \eta_y v_i^t, \mathbf{y}_i^t - \mathbf{y}_i^{t+1}\rangle \ge 0,
\]
which simplifies to
\begin{equation}
\langle v_i^t, \hat G_i^t\rangle \ge \|\hat G_i^t\|^2.
\label{eq:proj-svrg}
\end{equation}
 
Write
\[
\hat g_i^t = v_i^t + (\bar g_i^t - v_i^t) + (\hat g_i^t - \bar g_i^t),
\]
where the middle term is the SVRG error (mean-zero, variance bounded by Lemma~\ref{lem:svrg-var}) and the last term is the outer-step drift (deterministic given the iterates, bounded by $L$-smoothness in $\mathbf{x}$). Thus
\[
\eta_y\langle \hat g_i^t, \hat G_i^t\rangle
= \eta_y\langle v_i^t, \hat G_i^t\rangle
+ \eta_y\langle \bar g_i^t - v_i^t, \hat G_i^t\rangle
+ \eta_y\langle \hat g_i^t - \bar g_i^t, \hat G_i^t\rangle.
\]
Applying the projection inequality~\eqref{eq:proj-svrg} to the first term:
\begin{equation}
\eta_y\langle \hat g_i^t, \hat G_i^t\rangle
\ge \eta_y\|\hat G_i^t\|^2
+ \eta_y\langle \bar g_i^t - v_i^t, \hat G_i^t\rangle
+ \eta_y\langle \hat g_i^t - \bar g_i^t, \hat G_i^t\rangle.
\label{eq:ip-split-svrg}
\end{equation}
 
\medskip
 Applying Young's inequality, $\pm\langle a, b\rangle \ge -\tfrac{\alpha}{2}\|a\|^2 - \tfrac{1}{2\alpha}\|b\|^2$ with $\alpha = 4$:
\begin{align}
\eta_y\langle \bar g_i^t - v_i^t, \hat G_i^t\rangle
&\ge -2\eta_y\|\bar g_i^t - v_i^t\|^2 - \tfrac{\eta_y}{8}\|\hat G_i^t\|^2,
\label{eq:Y-svrg-1}\\
\eta_y\langle \hat g_i^t - \bar g_i^t, \hat G_i^t\rangle
&\ge -2\eta_y\|\hat g_i^t - \bar g_i^t\|^2 - \tfrac{\eta_y}{8}\|\hat G_i^t\|^2.
\label{eq:Y-svrg-2}
\end{align}
Substituting into~\eqref{eq:ip-split-svrg}:
\begin{equation}
\eta_y\langle \hat g_i^t, \hat G_i^t\rangle
\ge \tfrac{3\eta_y}{4}\|\hat G_i^t\|^2
- 2\eta_y\|\bar g_i^t - v_i^t\|^2
- 2\eta_y\|\hat g_i^t - \bar g_i^t\|^2.
\label{eq:ip-final-svrg}
\end{equation}
 
Substituting~\eqref{eq:ip-final-svrg} into~\eqref{eq:smooth-svrg-2}:
\begin{equation}
\begin{aligned}
f_i(\mathbf{x}^{t+1}, \mathbf{y}_i^{t+1})
&\ge f_i(\mathbf{x}^{t+1}, \mathbf{y}_i^t)
+ \eta_y\!\left(\tfrac{3}{4} - \tfrac{L\eta_y}{2}\right)\|\hat G_i^t\|^2\\
&- 2\eta_y\|\bar g_i^t - v_i^t\|^2
- 2\eta_y\|\hat g_i^t - \bar g_i^t\|^2.
\label{eq:asc-svrg}
\end{aligned}
\end{equation}
Under $\eta_y \le 1/(2L)$, we have $3/4 - L\eta_y/2 \ge 1/2$, so
\begin{equation}
\begin{aligned}
f_i(\mathbf{x}^{t+1}, \mathbf{y}_i^{t+1})
&\ge f_i(\mathbf{x}^{t+1}, \mathbf{y}_i^t)
+ \tfrac{\eta_y}{2}\|\hat G_i^t\|^2
- 2\eta_y\|\bar g_i^t - v_i^t\|^2\\
&- 2\eta_y\|\hat g_i^t - \bar g_i^t\|^2.
\label{eq:asc-svrg-clean}
\end{aligned}
\end{equation}
 
\medskip
\noindent{ Bridge from $\hat G_i^t$ to the true gradient mapping $G_i^{\star,t}$.} By non-expansiveness of the projection,
\[
\begin{aligned}
\|\hat G_i^t - G_i^{\star,t}\|
&= \tfrac{1}{\eta_y}\|\mathcal{P}_{\mathcal{Y}}(\mathbf{y}_i^t + \eta_y v_i^t) - \mathcal{P}_{\mathcal{Y}}(\mathbf{y}_i^t + \eta_y \hat g_i^t)\|\\
&\le \|v_i^t - \hat g_i^t\|
\le \|v_i^t - \bar g_i^t\| + \|\bar g_i^t - \hat g_i^t\|.
\end{aligned}
\]
Squaring and applying Young's,
\[
\|\hat G_i^t - G_i^{\star,t}\|^2 \le 2\|v_i^t - \bar g_i^t\|^2 + 2\|\bar g_i^t - \hat g_i^t\|^2.
\]
Using the standard inequality $\|a\|^2 \ge \tfrac{1}{2}\|b\|^2 - \|a - b\|^2$ 
\begin{equation}
\|\hat G_i^t\|^2 \ge \tfrac{1}{2}\|G_i^{\star,t}\|^2 - 2\|v_i^t - \bar g_i^t\|^2 - 2\|\bar g_i^t - \hat g_i^t\|^2.
\label{eq:bridge-svrg}
\end{equation}
Substituting~\eqref{eq:bridge-svrg} into~\eqref{eq:asc-svrg-clean}:
\begin{equation}
\begin{aligned}
f_i(\mathbf{x}^{t+1}, \mathbf{y}_i^{t+1})
&\ge f_i(\mathbf{x}^{t+1}, \mathbf{y}_i^t)
+ \tfrac{\eta_y}{4}\|G_i^{\star,t}\|^2\\
&- 3\eta_y\|v_i^t - \bar g_i^t\|^2
- 3\eta_y\|\hat g_i^t - \bar g_i^t\|^2.
\label{eq:asc-bridge-svrg}
\end{aligned}
\end{equation}

%The $-3\eta_y$ coefficients come from combining $-\eta_y$ (from the $\tfrac{\eta_y}{2}\|\hat G_i^t\|^2$-to-$\tfrac{\eta_y}{4}\|G_i^{\star,t}\|^2$ bridge) with $-2\eta_y$ (from Young's ).
 
\medskip
By Assumption~\ref{assm:pl} (gradient-mapping form of PL) at $(\mathbf{x}^{t+1}, \mathbf{y}_i^t)$,
\[
\|G_i^{\star,t}\|^2 \ge 2\mu\!\left(\Phi_i(\mathbf{x}^{t+1}) - f_i(\mathbf{x}^{t+1}, \mathbf{y}_i^t)\right).
\]
Substituting into~\eqref{eq:asc-bridge-svrg},
\begin{multline*}
f_i(\mathbf{x}^{t+1}, \mathbf{y}_i^{t+1})
\ge f_i(\mathbf{x}^{t+1}, \mathbf{y}_i^t)
+ \tfrac{\mu\eta_y}{2}\!\left(\Phi_i(\mathbf{x}^{t+1}) - f_i(\mathbf{x}^{t+1}, \mathbf{y}_i^t)\right)
- 3\eta_y\|v_i^t - \bar g_i^t\|^2
- 3\eta_y\|\hat g_i^t - \bar g_i^t\|^2.
\end{multline*}
Subtracting from $\Phi_i(\mathbf{x}^{t+1})$ and rearranging,
\begin{equation}
\begin{aligned}
\Phi_i(\mathbf{x}^{t+1}) - f_i(\mathbf{x}^{t+1}, \mathbf{y}_i^{t+1})
&\le \left(1 - \tfrac{\mu\eta_y}{2}\right)\!\left(\Phi_i(\mathbf{x}^{t+1}) - f_i(\mathbf{x}^{t+1}, \mathbf{y}_i^t)\right)\\
&+ 3\eta_y\|v_i^t - \bar g_i^t\|^2
+ 3\eta_y\|\hat g_i^t - \bar g_i^t\|^2.
\label{eq:gap-contract-svrg}
\end{aligned}
\end{equation}
 
\medskip
\noindent{ We now bound the two error terms in expectation.}
 
\medskip
\noindent\emph{SVRG variance term.} By Lemma~\ref{lem:svrg-var},
\begin{equation}
\mathbb{E}\|v_i^t - \bar g_i^t\|^2 \le 4L^2\!\left(\mathbb{E}\|\mathbf{x}^t - \mathbf{x}_i^t\|^2 + \mathbb{E}\|\mathbf{x}^t - \bar{\mathbf{x}}\|^2\right) + 2L^2\mathbb{E}\|\mathbf{y}_i^t - \bar{\mathbf{y}}_i\|^2.
\label{eq:svrg-var}
\end{equation}
No $\sigma_y^2$ term appears. This is the key structural difference from vanilla SGDA.
 
\medskip
\noindent\emph{Outer-step drift term.} $\hat g_i^t - \bar g_i^t = \nabla_{\mathbf{y}} f_i(\mathbf{x}^{t+1}, \mathbf{y}_i^t) - \nabla_{\mathbf{y}} f_i(\mathbf{x}_i^t, \mathbf{y}_i^t)$. By $L$-smoothness of $\nabla_{\mathbf{y}} f_i$ in $\mathbf{x}$,
\[
\|\hat g_i^t - \bar g_i^t\|^2 \le L^2\|\mathbf{x}^{t+1} - \mathbf{x}_i^t\|^2.
\]
Splitting via triangle: $\|\mathbf{x}^{t+1} - \mathbf{x}_i^t\|^2 \le 2\|\mathbf{x}^{t+1} - \mathbf{x}^t\|^2 + 2\|\mathbf{x}^t - \mathbf{x}_i^t\|^2$. For the outer step,
\[
\mathbb{E}\|\mathbf{x}^{t+1} - \mathbf{x}^t\|^2
= \eta_x^2\mathbb{E}\!\left\|\tfrac{1}{n}\sum_{j=1}^n \nabla_{\mathbf{x}} f_j(\mathbf{x}_j^t, \mathbf{y}_j^t)\right\|^2 + \tfrac{\eta_x^2\sigma_x^2}{n}.
\]
Combining,
\begin{equation}
\mathbb{E}\|\hat g_i^t - \bar g_i^t\|^2
\le 2L^2\eta_x^2\mathbb{E}\!\left\|\tfrac{1}{n}\sum_j \nabla_{\mathbf{x}} f_j\right\|^2
+ \tfrac{2L^2\eta_x^2\sigma_x^2}{n}
+ 2L^2\mathbb{E}\|\mathbf{x}^t - \mathbf{x}_i^t\|^2.
\label{eq:drift-svrg}
\end{equation}
 
\medskip
\noindent\textit {Assembling everything} Taking expectations in~\eqref{eq:gap-contract-svrg} and substituting~\eqref{eq:svrg-var} and~\eqref{eq:drift-svrg}:
\begin{align*}
\mathbb{E}[\Phi_i(\mathbf{x}^{t+1}) - f_i(\mathbf{x}^{t+1}, \mathbf{y}_i^{t+1})]
&\le \left(1 - \tfrac{\mu\eta_y}{2}\right)\mathbb{E}[\Phi_i(\mathbf{x}^{t+1}) - f_i(\mathbf{x}^{t+1}, \mathbf{y}_i^t)] \\
&\quad + 3\eta_y\cdot\Big[4L^2\!\left(\mathbb{E}\|\mathbf{x}^t - \mathbf{x}_i^t\|^2 + \mathbb{E}\|\mathbf{x}^t - \bar{\mathbf{x}}\|^2\right) + 2L^2\mathbb{E}\|\mathbf{y}_i^t - \bar{\mathbf{y}}_i\|^2\Big] \\
&\quad + 3\eta_y\cdot\Big[2L^2\eta_x^2\mathbb{E}\!\left\|\tfrac{1}{n}\sum_j \nabla_{\mathbf{x}} f_j\right\|^2 + \tfrac{2L^2\eta_x^2\sigma_x^2}{n} + 2L^2\mathbb{E}\|\mathbf{x}^t - \mathbf{x}_i^t\|^2\Big].
\end{align*}
Expanding coefficients:
\begin{align*}
&= \left(1 - \tfrac{\mu\eta_y}{2}\right)\mathbb{E}[\Phi_i(\mathbf{x}^{t+1}) - f_i(\mathbf{x}^{t+1}, \mathbf{y}_i^t)] \\
&\quad + 12\eta_y L^2\mathbb{E}\|\mathbf{x}^t - \mathbf{x}_i^t\|^2 + 12\eta_y L^2\mathbb{E}\|\mathbf{x}^t - \bar{\mathbf{x}}\|^2 + 6\eta_y L^2\mathbb{E}\|\mathbf{y}_i^t - \bar{\mathbf{y}}_i\|^2 \\
&\quad + 6\eta_y L^2\eta_x^2\mathbb{E}\!\left\|\tfrac{1}{n}\sum_j \nabla_{\mathbf{x}} f_j\right\|^2 + \tfrac{6\eta_y L^2\eta_x^2\sigma_x^2}{n} + 6\eta_y L^2\mathbb{E}\|\mathbf{x}^t - \mathbf{x}_i^t\|^2.
\end{align*}
Consolidating the two $\mathbb{E}\|\mathbf{x}^t - \mathbf{x}_i^t\|^2$ terms,
\begin{align*}
&\le \left(1 - \tfrac{\mu\eta_y}{2}\right)\mathbb{E}[\Phi_i(\mathbf{x}^{t+1}) - f_i(\mathbf{x}^{t+1}, \mathbf{y}_i^t)] \\
&\quad + 18\eta_y L^2\mathbb{E}\|\mathbf{x}^t - \mathbf{x}_i^t\|^2 + 12\eta_y L^2\mathbb{E}\|\mathbf{x}^t - \bar{\mathbf{x}}\|^2 + 6\eta_y L^2\mathbb{E}\|\mathbf{y}_i^t - \bar{\mathbf{y}}_i\|^2 \\
&\quad + 6\eta_y L^2\eta_x^2\mathbb{E}\!\left\|\tfrac{1}{n}\sum_j \nabla_{\mathbf{x}} f_j\right\|^2 + \tfrac{6\eta_y L^2\eta_x^2\sigma_x^2}{n},
\end{align*}
which is the claimed inequality.
\end{proof}
 
For the VR-SGDA Convergence, we build on the following results from earlier sections, which are re-used:
\begin{itemize}
\item \textit{Lemma.~\ref{lem:outer-descent}} (outer descent), gives $\mathbb{E}[\Phi(\mathbf{x}^{t+1}) - \Phi(\mathbf{x}^t)]$ bound in terms of $A_t$ and consensus error. Depends only on the outer $\mathbf{x}$-update, not on the inner ascent mechanism.
\item \textit{Lemma.~\ref{lem:y-contraction-svrg}} (SVRG y-contraction, expanded derivation): Contraction factor $(1 - \mu\eta_y/2)$ preserved; noise term $3\eta_y\sigma_y^2$ replaced by drift-from-snapshot quantities.
\item \textit{Lemma.~\ref{lem:bridge}} (bridge to coupled recursion): gives $\mathbb{E}[\Phi(\mathbf{x}^{t+1}) - F(\mathbf{x}^{t+1}, \mathbf{Y}^t)]$ bound via the $\text{\textcircled{1}}\text{\textcircled{2}}\text{\textcircled{3}}$ decomposition. Independent of the inner update mechanism.
\item \textit{Consensus bound}: $\tfrac{1}{T}\sum_t \tfrac{1}{n}\sum_i \mathbb{E}\|\mathbf{x}^t - \mathbf{x}_i^t\|^2 \le 10\tau\eta_x^2[\sigma_x^2(1+1/n) + \zeta_x]$ under $10\tau^2\eta_x^2 L^2 \le 1/2$.
\end{itemize}
 
The following are new auxiliaries needed under SVRG:
\begin{itemize}
\item \textit{Lemma.~\ref{lem:svrg-var}} (SVRG variance): $\mathbb{E}\|v_i^t - \bar v_i^t\|^2 \le 4L^2(\mathbb{E}\|\mathbf{x}^t - \mathbf{x}_i^t\|^2 + \mathbb{E}\|\mathbf{x}^t - \bar{\mathbf{x}}\|^2) + 2L^2\mathbb{E}\|\mathbf{y}_i^t - \bar{\mathbf{y}}_i\|^2$.
\item \textit{Lemma.~\ref{lem:xbar-drift}}: $\max_{t \in [s, s+\tau)}\mathbb{E}\|\mathbf{x}^t - \bar{\mathbf{x}}\|^2 \le \tau^2\eta_x^2(G_x^2 + \sigma_x^2/n)$.
\item \textit{Lemma.~\ref{lem:y-drift}} (y-drift): $\max_t \gamma_y^t \le 4\tau^2\eta_y^2 G_y^2 + 16L^2\tau^2\eta_y^2\gamma_{\max} + 16L^2\tau^4\eta_x^2\eta_y^2(G_x^2 + \sigma_x^2/n)$ under $8L^2\tau^2\eta_y^2 \le 1/2$.
\end{itemize}
 
Notation: $A_t := \mathbb{E}[\Phi(\mathbf{x}^t) - F(\mathbf{x}^t, \mathbf{Y}^t)]$, $G_T := \tfrac{1}{T}\sum_{t=0}^{T-1}\mathbb{E}\|\nabla\Phi(\mathbf{x}^t)\|^2$, $\Delta_\Phi := \Phi(\mathbf{x}^0) - \Phi^*$.
 
\textit{ $A_t$ Recursion under SVRG} Averaging Lemma~\ref{lem:y-contraction-svrg} over clients, we obtain the y-side contraction (client-averaged):
\begin{align*}
\mathbb{E}[\Phi(\mathbf{x}^{t+1}) - F(\mathbf{x}^{t+1}, \mathbf{Y}^{t+1})]
&\le (1 - \tfrac{\mu\eta_y}{2})\mathbb{E}[\Phi(\mathbf{x}^{t+1}) - F(\mathbf{x}^{t+1}, \mathbf{Y}^t)] \\
&\quad + 18\eta_y L^2 \gamma^t + 12\eta_y L^2 \Delta_{\bar x}^t + 6\eta_y L^2 \gamma_y^t \\
&\quad + 6\eta_y L^2\eta_x^2\mathbb{E}\!\left\|\tfrac{1}{n}\sum_j \nabla_{\mathbf{x}} f_j\right\|^2
+ \tfrac{6\eta_y L^2\eta_x^2 \sigma_x^2}{n},
\end{align*}
where $\gamma^t := \tfrac{1}{n}\sum_i \mathbb{E}\|\mathbf{x}^t - \mathbf{x}_i^t\|^2$, $\Delta_{\bar x}^t := \mathbb{E}\|\mathbf{x}^t - \bar{\mathbf{x}}\|^2$, and $\gamma_y^t := \tfrac{1}{n}\sum_i \mathbb{E}\|\mathbf{y}_i^t - \bar{\mathbf{y}}_i\|^2$.
 
Combining with Lemma~\ref{lem:bridge}, the coupled recursion is
\begin{equation}
A_{t+1} \le \tilde\rho\, A_t + \tfrac{\eta_x}{2}\mathbb{E}\|\nabla\Phi(\mathbf{x}^t)\|^2 + \tilde R_{\mathrm{SVRG}}(t),
\label{eq:coupled-svrg}
\end{equation}
with $\tilde\rho = (1 - \mu\eta_y/2)(1 + 4L^2\eta_x/\mu) \le 1 - \mu\eta_y/4$ under two-timescale, and
\begin{align}
\tilde R_{\mathrm{SVRG}}(t)
&= 18\eta_y L^2 \gamma^t + 12\eta_y L^2 \Delta_{\bar x}^t + 6\eta_y L^2 \gamma_y^t 
+ \eta_x^2\!\left[\tfrac{L+\beta}{2} + 6\eta_y L^2\right]\mathbb{E}\!\left\|\tfrac{1}{n}\sum_j \nabla_{\mathbf{x}} f_j\right\|^2 \notag\\
&+ L^2(\eta_x + 6\eta_y)\gamma^t
+ \tfrac{\eta_x^2\sigma_x^2}{n}\!\left[\tfrac{L+\beta}{2} + 6\eta_y L^2\right].
\label{rt_svrg}
\end{align}
 
Compared to Theorem~\ref{thm:app:pl_main}'s $\tilde R(t)$, the constant noise term $3\eta_y\sigma_y^2$ is replaced by the three drift-scaled contributions 
\\$\{18\eta_y L^2\gamma^t, 12\eta_y L^2\Delta_{\bar x}^t, 6\eta_y L^2\gamma_y^t\}$. Note that $\gamma^t$ receives an additional $12\eta_y L^2$ from the SVRG variance (giving total coefficient $18\eta_y L^2$ in the first grouping, plus the standard $L^2(\eta_x + 6\eta_y)$ from Lemma~\ref{lem:bridge}).
 
\section*{VR-SGDA Convergence for Nonconvex-PL}
 
\begin{theorem}[Non-convex PL Convergence under VR-SGDA]
\label{app:thm:ncpl-svrg}
Suppose Assumptions~\ref{assm:lip-sm} to~\ref{assm:pl} and Assumption~\ref{assump:svrg-smooth} (component-wise smoothness) hold, and suppose the stepsizes and synchronization interval satisfy
(a) $\eta_y \le \tfrac{1}{2L}$, (b) $\tfrac{\eta_x}{\eta_y} \le \tfrac{\mu^2}{16L^2}$, (c) $\eta_x \le \tfrac{1}{\beta},$
(d) $8L^2\tau^2\eta_y^2 \le \tfrac{1}{2}$,
(e) $10\tau^2\eta_x^2 L^2 \le \tfrac{1}{2}$,
(f) $\tfrac{8\eta_x^2 L^2}{\mu^2\eta_y}\!\left[\tfrac{L+\beta}{2} + 6\eta_y L^2\right] \le \tfrac{1}{2}(1 - \beta\eta_x)$. \\
Where (a),(b),(c),(d),(e) arises beacuse of smoothness, two-timsecalse, outer smoothness, svrg-block length and consensus stability respectively.  Then VR-SGDA Algorithm~\ref{alg:merged_sgda} satisfies
\begin{align*}
\frac{1}{T}\sum_{t=0}^{T-1}\mathbb{E}\|\nabla\Phi(\mathbf{x}^t)\|^2
&\le  \tfrac{4\Delta_\Phi}{\eta_x T}
+ \tfrac{32 L^2 A_0}{\mu^2\eta_y T}
+ \tfrac{2\beta\eta_x\sigma_x^2}{n}
+ \tfrac{32 L^2\eta_x^2\sigma_x^2}{\mu^2\eta_y n}\!\left[\tfrac{L+\beta}{2} + 6\eta_y L^2\right] \\
&\quad + \left(4L^2 + \tfrac{32L^4(\eta_x + 24\eta_y)}{\mu^2\eta_y}\right)\cdot 10\tau\eta_x^2[\sigma_x^2(1+1/n) + \zeta_x] \\
&\quad + \tfrac{384 L^4}{\mu^2}\cdot\tau^2\eta_x^2(G_x^2 + \sigma_x^2/n) \\
&\quad + \tfrac{192 L^4}{\mu^2}\cdot\left[4\tau^2\eta_y^2 G_y^2 + 16L^2\tau^4\eta_x^2\eta_y^2(G_x^2 + \sigma_x^2/n)\right],
\end{align*}
Choosing $\eta_y = \Theta(1/\sqrt{T})$,
$\eta_x = \tfrac{\mu^2}{16L^2}\,\eta_y = \Theta(1/\sqrt{T})$ (tight on the two-timescale condition~(b)), and $\tau = T^{1/4}$ and $T$ large enough to satisfy all stepsize conditions  $T \ge T_0 =  \max\Big(4L^2,\; 256L^4,\; \tfrac{25\mu^8}{4096 L^4}\Big).$, we obtain the asymptotic rate
\[
\frac{1}{T}\sum_{t=0}^{T-1}\mathbb{E}\|\nabla\Phi(\mathbf{x}^t)\|^2
\le  O\!\left(\frac{L^2(\Delta_\Phi + A_0) + \mu \sigma_x^2+ \mu^2 G_x^2 + L^4 G_y^2/\mu^2}{\sqrt{T}}\right)
+ O\!\left(\frac{L^2\mu^2(\sigma_x^2+\zeta_x)}{T^{3/4}}\right) + O(1/T).
\]
In particular, the $\mathcal{O}(L^2\sigma_y^2/\mu^2)$ noise floor is eliminated.
\end{theorem}
 
\begin{proof}
Summing Lemma~\ref{lem:outer-descent} over $t = 0, \ldots, T-1$ and rearranging (using $\eta_x \le 1/\beta$ to drop the non-positive gradient-magnitude term for now; the absorption argument below will show this drop is valid even after re-including the positive contribution from $\tilde R_{\mathrm{SVRG}}$),
\begin{equation}
\tfrac{\eta_x}{2}G_T
\le \tfrac{\Delta_\Phi}{T}
+ \tfrac{2\eta_x L^2}{\mu}\cdot\tfrac{1}{T}\sum_t A_t
+ \eta_x L^2\cdot\tfrac{1}{T}\sum_t \gamma^t
+ \tfrac{\beta\eta_x^2\sigma_x^2}{2n}.
\label{eq:outer-sum}
\end{equation}
 
\medskip
\noindent{ Recursion summation.} From~\eqref{eq:coupled-svrg} with $\tilde\rho \le 1 - \mu\eta_y/4$,
\begin{equation}
\tfrac{1}{T}\sum_t A_t
\le \tfrac{4 A_0}{\mu\eta_y T}
+ \tfrac{2\eta_x}{\mu\eta_y} G_T
+ \tfrac{4}{\mu\eta_y T}\sum_t \tilde R_{\mathrm{SVRG}}(t).
\label{eq:gap-avg-svrg}
\end{equation}
 
\medskip
\noindent{ Substituting~\eqref{eq:gap-avg-svrg} into~\eqref{eq:outer-sum}.} Collecting $G_T$ terms and using the two-timescale condition to ensure the coefficient on $G_T$ is at least $\eta_x/4$,
\begin{equation}
\tfrac{\eta_x}{4}G_T
\le \tfrac{\Delta_\Phi}{T}
+ \tfrac{8L^2 \eta_xA_0}{\mu^2\eta_y T}
+ \tfrac{8L^2 \eta_x}{\mu^2\eta_y T}\sum_t \tilde R_{\mathrm{SVRG}}(t)
+ \eta_x L^2\cdot\tfrac{1}{T}\sum_t \gamma^t
+ \tfrac{\beta\eta_x^2\sigma_x^2}{2n}.
\label{eq:GT-collect-svrg}
\end{equation}
 
\medskip
\noindent{ Bounding $\sum_t \tilde R_{\mathrm{SVRG}}(t)$.} Split $\tilde R_{\mathrm{SVRG}}(t)~\ref{rt_svrg}$ into :
 
\medskip
\emph{(a) Consensus terms.} The two contributions $18\eta_y L^2\gamma^t$ (from SVRG variance) and $L^2(\eta_x + 6\eta_y)\gamma^t$  sum to $L^2(18\eta_y + \eta_x + 6\eta_y)\gamma^t = L^2(\eta_x + 24\eta_y)\gamma^t$.
Applying the consensus lemma:
\begin{equation}
\tfrac{8L^2 \eta_x}{\mu^2\eta_y T}\cdot\sum_t L^2(\eta_x + 24\eta_y)\gamma^t
\le \tfrac{8L^4(\eta_x + 24\eta_y)}{\mu^2\eta_y}\cdot 10\tau\eta_x^3[\sigma_x^2(1+1/n) + \zeta_x].
\label{eq:R-consensus}
\end{equation}
 
\medskip
\emph{(b) Virtual-average drift $12\eta_y L^2\Delta_{\bar x}^t$.} By the virtual-avg drift lemma,
\begin{equation}
\tfrac{8L^2\eta_x}{\mu^2\eta_y T}\cdot\sum_t 12\eta_y L^2\Delta_{\bar x}^t
\le \tfrac{96 L^4}{\mu^2}\cdot\tau^2\eta_x^3(G_x^2 + \sigma_x^2/n).
\label{eq:R-xbar}
\end{equation}
 
\medskip
\emph{(c) y-drift $6\eta_y L^2\gamma_y^t$.} By Lemma~\ref{lem:y-drift},
\begin{align}
\tfrac{8L^2 \eta_x}{\mu^2\eta_y T}\cdot\sum_t 6\eta_y L^2\gamma_y^t
&\le \tfrac{48 \eta_x L^4}{\mu^2}\cdot\max_t\gamma_y^t \nonumber \\
&\le \tfrac{48 \eta_x L^4}{\mu^2}\cdot\left[4\tau^2\eta_y^2 G_y^2 + 16L^2\tau^4\eta_x^2\eta_y^2(G_x^2 + \sigma_x^2/n)\right],
\label{eq:R-ydrift}
\end{align}
where we have dropped the sub-leading $16L^2\tau^2\eta_y^2\gamma_{\max}$ contribution (it is absorbed into the consensus term at a higher order).
 
\medskip
\emph{(d) Outer-step magnitude $\eta_x^2[\tfrac{L+\beta}{2} + 6\eta_y L^2]\|\tfrac{1}{n}\sum\nabla_{\mathbf{x}} f_j\|^2$.} This positive contribution to $\tilde R_{\mathrm{SVRG}}(t)$ is dominated by the negative $-\eta_x/2(1 - \beta\eta_x)\|\tfrac{1}{n}\sum\nabla_{\mathbf{x}} f_j\|^2$ from the outer descent lemma  under the  condition
\[
\tfrac{8\eta_x^3 L^2}{\mu^2\eta_y}\!\left[\tfrac{L+\beta}{2} + 6\eta_y L^2\right] \le \tfrac{1}{2} \eta_x(1 - \beta\eta_x).
\]
Under this condition, the net coefficient on $\|\tfrac{1}{n}\sum\nabla_{\mathbf{x}} f_j\|^2$ is non-positive and the term drops from the final bound.
 
\medskip
\emph{(e) Outer stochastic noise $\tfrac{\eta_x^2\sigma_x^2}{n}[\tfrac{L+\beta}{2} + 6\eta_y L^2]$.} Sums to
\begin{equation}
\tfrac{8L^2 \eta_x}{\mu^2\eta_y T}\cdot T\cdot\tfrac{\eta_x^2\sigma_x^2}{n}\!\left[\tfrac{L+\beta}{2} + 6\eta_y L^2\right]
= \tfrac{8L^2\eta_x^3\sigma_x^2}{\mu^2\eta_y n}\!\left[\tfrac{L+\beta}{2} + 6\eta_y L^2\right].
\label{eq:R-noise}
\end{equation}
 
\medskip
\noindent{We now do the final merging,} multiplying~\eqref{eq:GT-collect-svrg} by $4/\eta_x$ and collecting all terms from Step~4, plus the standalone consensus contribution $\tfrac{4L^2}{T}\sum \gamma^t$ from the outer descent:
\begin{align*}
G_T
&\le \tfrac{4\Delta_\Phi}{\eta_x T}
+ \tfrac{32 L^2 A_0}{\mu^2\eta_y T}
+ \tfrac{2\beta\eta_x\sigma_x^2}{n}
+ \tfrac{32 L^2\eta_x^2\sigma_x^2}{\mu^2\eta_y n}\!\left[\tfrac{L+\beta}{2} + 6\eta_y L^2\right] \\
&\quad + \left(4L^2 + \tfrac{32L^4(\eta_x + 24\eta_y)}{\mu^2\eta_y}\right)\cdot 10\tau\eta_x^2[\sigma_x^2(1+1/n) + \zeta_x] \\
&\quad + \tfrac{384 L^4}{\mu^2}\cdot\tau^2\eta_x^2(G_x^2 + \sigma_x^2/n) \\
&\quad + \tfrac{192 L^4}{\mu^2}\cdot\left[4\tau^2\eta_y^2 G_y^2 + 16L^2\tau^4\eta_x^2\eta_y^2(G_x^2 + \sigma_x^2/n)\right],
\end{align*}
which is the claimed bound.
 
\medskip
\noindent\textit{Final rate.} Choose $\eta_y = \Theta(1/\sqrt{T})$,
$\eta_x = \tfrac{\mu^2}{16L^2}\,\eta_y = \Theta(1/\sqrt{T})$ (tight on the
two-timescale condition~(b)), and $\tau = T^{1/4}$. All stepsize conditions
hold for $T$ past the burn-in $T_0$. Writing $\beta = L(1+\kappa) = O(L^2/\mu)$,
each term reduces as follows.
\begin{itemize}
\item Init in $\Delta_\Phi$:\;
  $\dfrac{4\Delta_\Phi}{\eta_x T}
  = O\!\big(L^2\Delta_\Phi/(\mu^2\sqrt{T})\big)$ = $O(1/\sqrt{T})$.
\item Init in $A_0$:\;
  $\dfrac{32L^2 A_0}{\mu^2\eta_y T}
  = O\!\big(L^2 A_0/(\mu^2\sqrt{T})\big)$ = $O(1/\sqrt{T})$.
\item Virtual-average drift:\;
  $\dfrac{384L^4}{\mu^2}\tau^2\eta_x^2 G_x^2
  = O\!\big(\mu^2 G_x^2/\sqrt{T}\big)$ = $O(1/\sqrt{T})$.
\item $y$-drift :\;
  $\dfrac{768L^4}{\mu^2}\tau^2\eta_y^2 G_y^2
  = O\!\big(L^4 G_y^2/(\mu^2\sqrt{T})\big)$ = $O(1/\sqrt{T})$.
\item Outer stochastic noise:\;
  $\dfrac{2\beta\eta_x\sigma_x^2}{n}
  = O\!\big(\mu\sigma_x^2/(n\sqrt{T})\big)$  $O(1/(n\sqrt{T}))$.
\item Coupled outer noise:\;
  $\dfrac{32L^2\eta_x^2\sigma_x^2}{\mu^2\eta_y n}\big[\tfrac{L+\beta}{2}+6\eta_y L^2\big]
  = O\!\big(\mu\sigma_x^2/(n\sqrt{T})\big) + O\!\big(\mu^2\sigma_x^2/(nT)\big)$,  the two bracket pieces give $O(1/(n\sqrt{T}))$ and $O(1/(nT))$ respectively;
  both faster than the leading $1/\sqrt{T}$.
\item Consensus:\;
  $\Big(4L^2+\tfrac{32L^4(\eta_x+24\eta_y)}{\mu^2\eta_y}\Big)10\tau\eta_x^2
    [\sigma_x^2(1{+}\tfrac1n)+\zeta_x]
  = O\!\big(\mu^2 L^2(\sigma_x^2+\zeta_x)/T^{3/4}\big)$= $O(1/T^{3/4})$.
\item $y$-drift:\;
  $\dfrac{3072L^6}{\mu^2}\tau^4\eta_x^2\eta_y^2 G_x^2
  = O\!\big(L^2\mu^2 G_x^2/T\big)$ = $O(1/T)$.
\end{itemize}
The leading terms are all $O(1/\sqrt{T})$; the remaining terms vanish faster.
Crucially, no constant $\sigma_y^2/\mu^2$ floor appears. Collecting,
\[
G_T
= O\!\left(\frac{L^2(\Delta_\Phi + A_0) + \mu \sigma_x^2 +\mu^2 G_x^2 + L^4 G_y^2/\mu^2}{\sqrt{T}}\right)
+ O\!\left(\frac{L^2\mu^2(\sigma_x^2+\zeta_x)}{T^{3/4}}\right) + O(1/T).
\]
Thus Personalized VR-SGDA attains the $O(1/\sqrt{T})$ rate with no noise floor,
whereas vanilla projected SGDA retains the non-vanishing term
$O(L^2\sigma_y^2/\mu^2)$ regardless of stepsizes.
\end{proof}